\PassOptionsToPackage{table,dvipsnames}{xcolor}

\documentclass[10pt,twocolumn,letterpaper]{article}

\usepackage{iccv}              

\usepackage{caption}
\usepackage{makecell}
\usepackage{nicematrix}
\usepackage{array}
\usepackage{tikz}
\usepackage{soul} 
\usepackage{xcolor} 
\usepackage{algorithm}
\usepackage{algorithmic}
\usepackage{multirow}
\usepackage{graphicx}
\usepackage{float}
\usepackage{subcaption}
\usepackage{rotating}   
\usepackage{setspace}
\usepackage{booktabs}   
\usepackage{amssymb}    
\usepackage{pifont}  
\usepackage{amsmath}
\usepackage{arydshln}  
\usepackage{booktabs}  
\usepackage{colortbl}  
\usepackage{booktabs}
\usetikzlibrary{spy, calc}
\definecolor{iccvblue}{rgb}{0.21,0.49,0.74}
\usepackage[pagebackref,breaklinks,colorlinks,allcolors=iccvblue]{hyperref}
\usepackage{threeparttable}

\def\paperID{****} 
\def\confName{QIFK} 
\def\confYear{2025}

\title{Dual Recursive Feedback on Generation and Appearance Latents for Pose-Robust Text-to-Image Diffusion}

\newcommand*\samethanks[1][\value{footnote}]{\footnotemark[#1]}
\author{
Jiwon Kim$^1$ \quad Pureum Kim$^1$ \quad SeonHwa Kim$^1$ \quad Soobin Park$^2$  \quad Eunju Cha$^2$\thanks{Corresponding author.} \quad Kyong Hwan Jin$^1$\samethanks \\
$^1$Korea University \quad\qquad $^2$Sookmyung Women's University \\
{\tt\small \{jwonkim, pureum\_kim, sunkim0062, kyong\_jin\}@korea.ac.kr, \{psb1219j, eunju.cha\}@sookmyung.ac.kr}
}

\begin{document}

\maketitle
\begin{abstract}
Recent advancements in controllable text-to-image (T2I) diffusion models, such as Ctrl-X and FreeControl, have demonstrated robust spatial and appearance control without requiring auxiliary module training. However, these models often struggle to accurately preserve spatial structures and fail to capture fine-grained conditions related to object poses and scene layouts. To address these challenges, we propose a training-free \textit{Dual Recursive Feedback (DRF)} system that properly reflects control conditions in controllable T2I models. 
The proposed DRF consists of appearance feedback and generation feedback that recursively refines the intermediate latents to better reflect the given appearance information and the user's intent.
This dual-update mechanism guides latent representations toward reliable manifolds, effectively integrating structural and appearance attributes. Our approach enables fine-grained generation even between class-invariant structure-appearance fusion, such as transferring human motion onto a tiger's form. Extensive experiments demonstrate the efficacy of our method in producing high-quality, semantically coherent, and structurally consistent image generations. Our source code is available at \url{https://github.com/jwonkm/DRF}.

\end{abstract}
\vspace{-12pt}    
\section{Introduction}
\label{sec:intro}
Recent advancements in diffusion models~\cite{dhariwal2021diffusion,ho2020denoising,podell2023sdxl,rombach2022high,song2020score} have led to rapid progress in the field of image generation. Among these developments, the need to generate images that accurately reflect a user’s intended expression has emerged, resulting in various approaches for incorporating user intent into Text-to-Image (T2I) generation and Image-to-Image (I2I) translation tasks.

Text-to-Image (T2I) techniques have been developed to synthesize realistic and creative images solely from textual descriptions. However, T2I methods often face limitations in controlling the spatial composition of the generated images, making the representation of complex structures challenging. To address this, T2I models have progressively incorporated conditional guidance techniques, enabling more detailed control over the desired image attributes. For instance, approaches such as ControlNet~\cite{zhang2023adding} leverage specific structures (\textit{e.g.,} pose, edge, or depth) to provide users with finer control, thus allowing the generated images to align more precisely with user intent.
\begin{figure}[!t]
        \centering
        \includegraphics[width=0.48\textwidth]{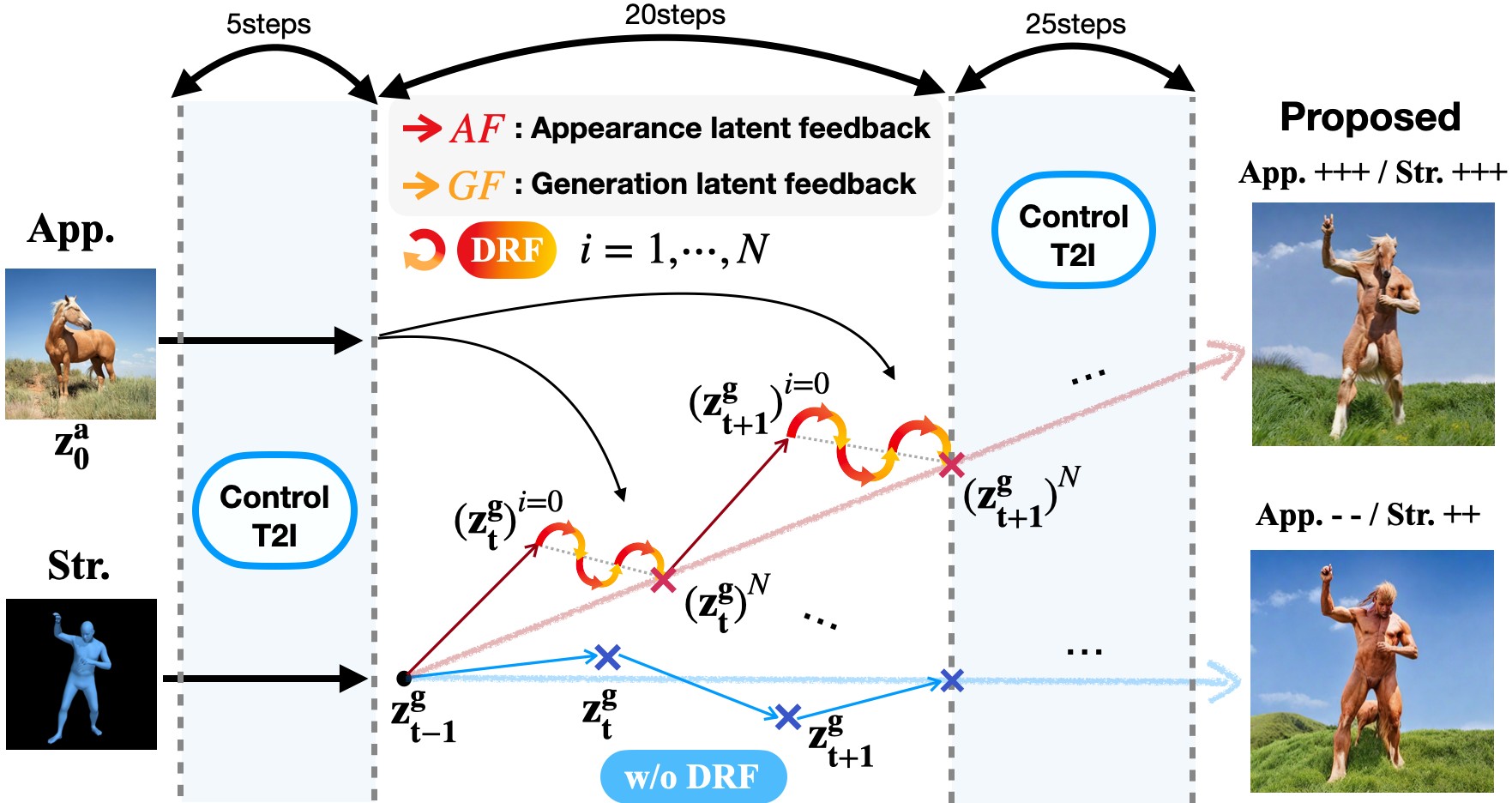}
    \vspace{-18pt}
    \caption{\textbf{Flowchart of Dual Recursive Feedback (DRF).} Illustration of a diffusion-based generative model with latent feedback mechanisms (DRF) for controlling both appearance 
    and generation latent 
    in class-invariant text-to-image synthesis. The proposed method refines latent updates to achieve fine-grained control, improving results with desired structural and appearance attributes.}
    \label{fig:teasor}
    \vspace{-14pt}
\end{figure}
Beyond T2I generations, a line of research known as Image-to-Image (I2I) translation~\cite{hertz2022prompt,tumanyan2023plug,hertz2023delta,kim2025identity} extends generating capabilities by taking an existing source image as input, and then transforming it according to a text query. Since this approach leverages the structural or contextual elements of the original image, I2I offers more precise control than T2I. Nonetheless, a technical trade-off between reliability and diversity still remained.

\begin{figure*}[t!]
    \centering
   \begin{subfigure}[b]{0.12\textwidth}  
    \includegraphics[width=\textwidth]{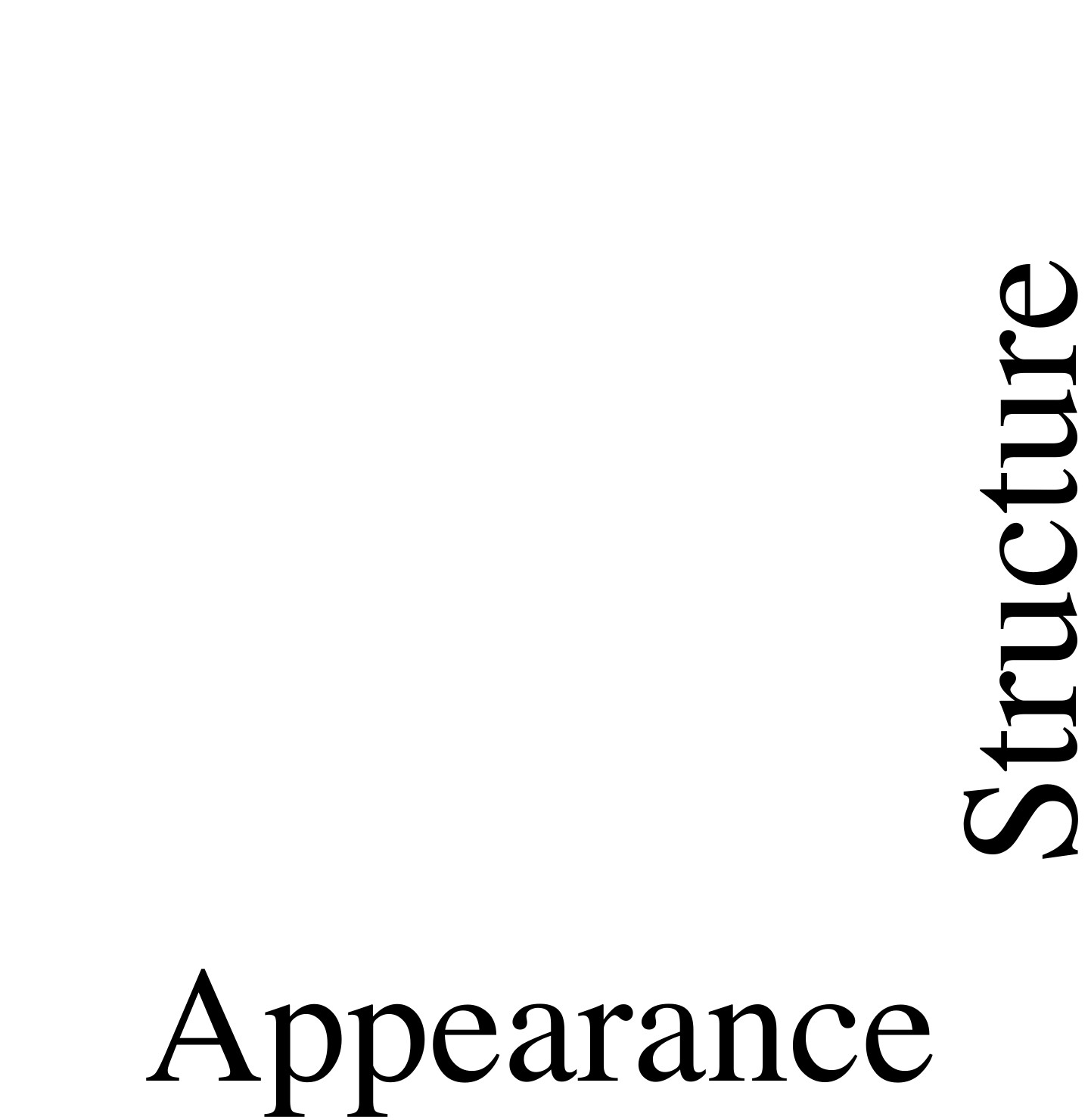}
    \caption*{}
\end{subfigure}
\hfill
\begin{subfigure}[b]{0.12\textwidth}
    \includegraphics[width=\textwidth]{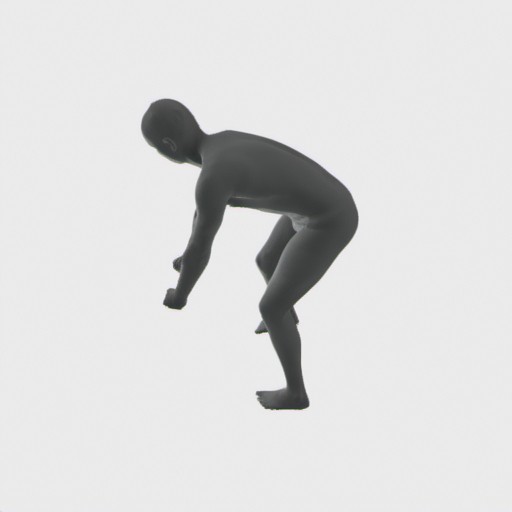}
    \caption*{}
\end{subfigure}
\hfill
\begin{subfigure}[b]{0.12\textwidth}
    \includegraphics[width=\textwidth]{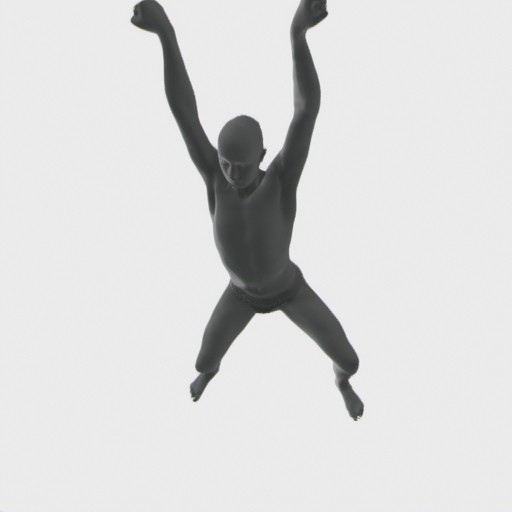}
    \caption*{}
\end{subfigure}
\hfill
\begin{subfigure}[b]{0.12\textwidth}
    \includegraphics[width=\textwidth]{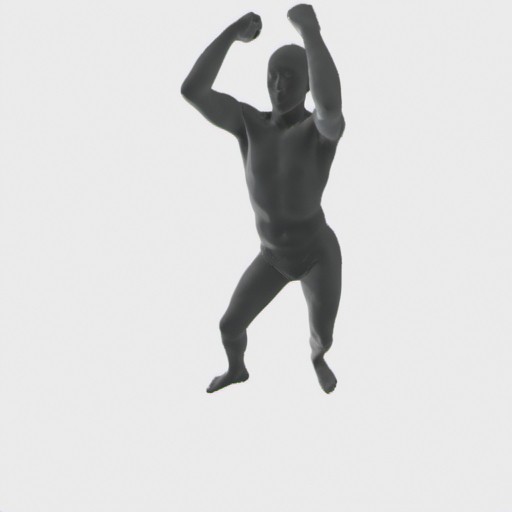}
    \caption*{}
\end{subfigure}
\hfill
\begin{subfigure}[b]{0.12\textwidth}
    \includegraphics[width=\textwidth]{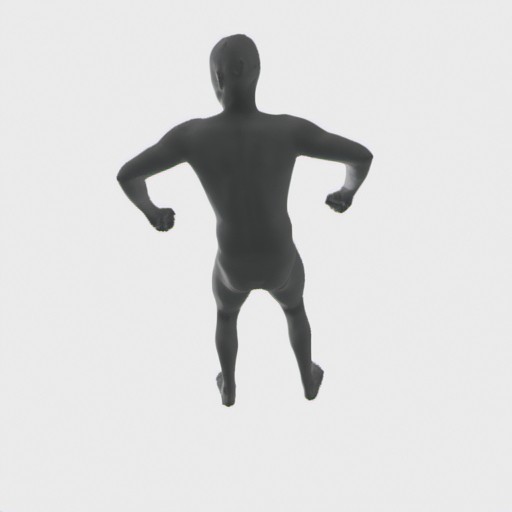}
    \caption*{}
\end{subfigure}
\hfill
\begin{subfigure}[b]{0.12\textwidth}
    \includegraphics[width=\textwidth]{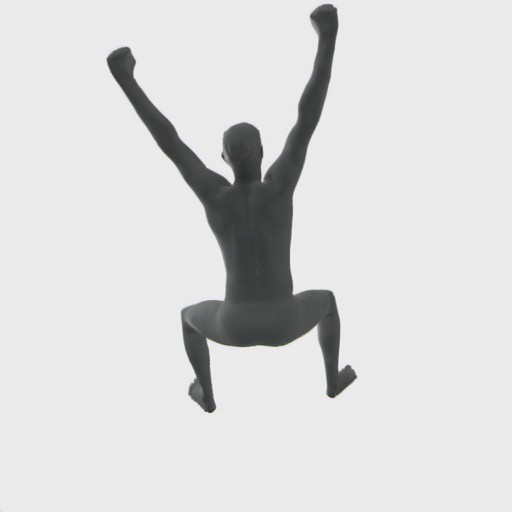}
    \caption*{}
\end{subfigure}
\hfill
\begin{subfigure}[b]{0.12\textwidth}
    \includegraphics[width=\textwidth]{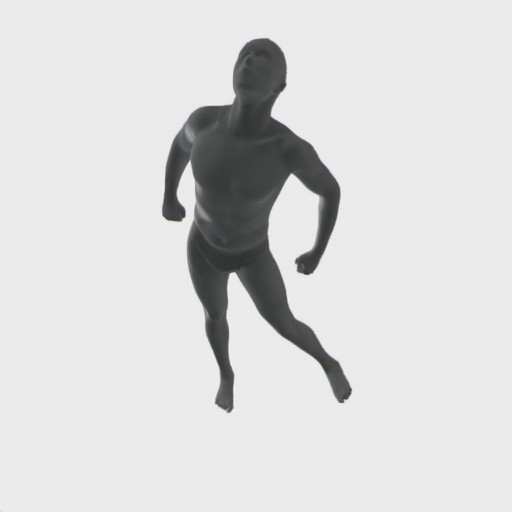}
    \caption*{}
\end{subfigure}
\hfill
\begin{subfigure}[b]{0.12\textwidth}
    \includegraphics[width=\textwidth]{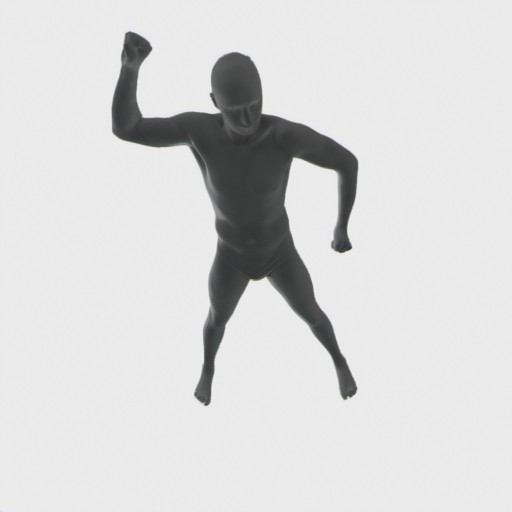}
    \caption*{}
\end{subfigure}
\\[-1.0em]  

\begin{subfigure}[b]{0.12\textwidth}
    \includegraphics[width=\textwidth]{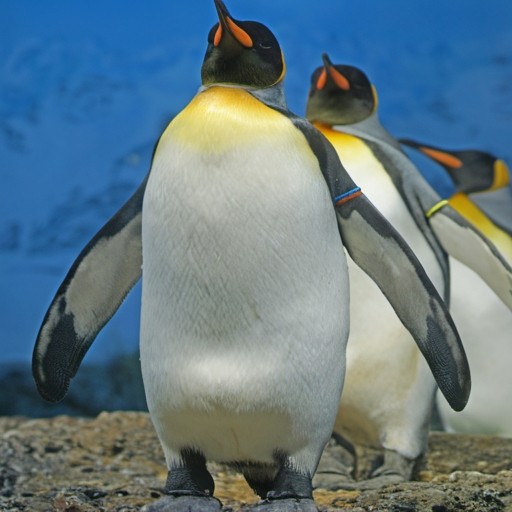}
    \caption*{}
\end{subfigure}
\hfill
\begin{subfigure}[b]{0.12\textwidth}
    \includegraphics[width=\textwidth]{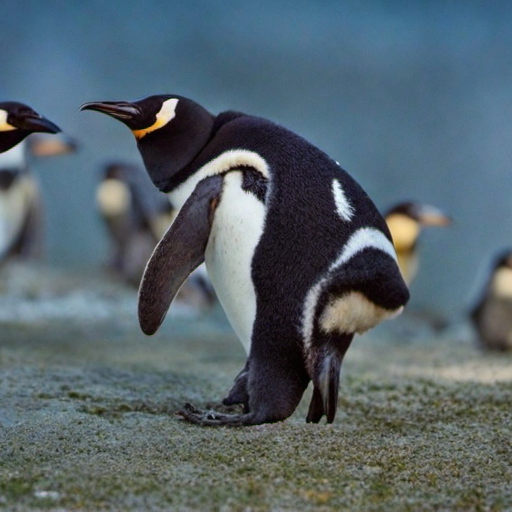}
    \caption*{}
\end{subfigure}
\hfill
\begin{subfigure}[b]{0.12\textwidth}
    \includegraphics[width=\textwidth]{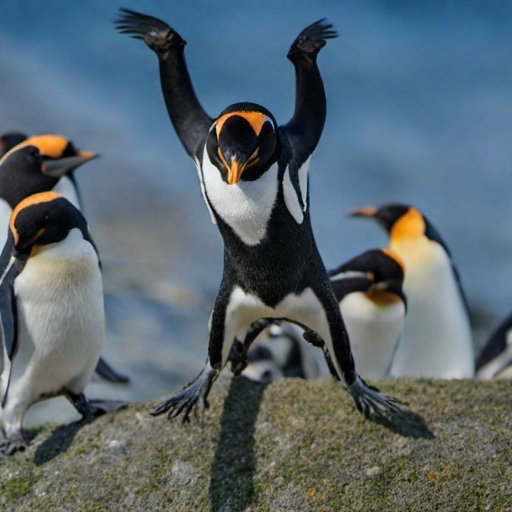}
    \caption*{}
\end{subfigure}
\hfill
\begin{subfigure}[b]{0.12\textwidth}
    \includegraphics[width=\textwidth]{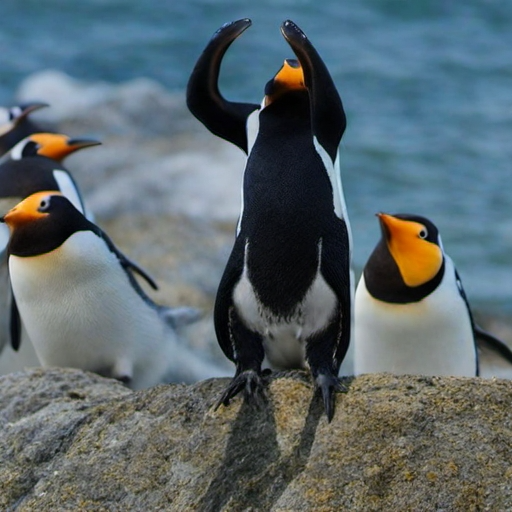}
    \caption*{}
\end{subfigure}
\hfill
\begin{subfigure}[b]{0.12\textwidth}
    \includegraphics[width=\textwidth]{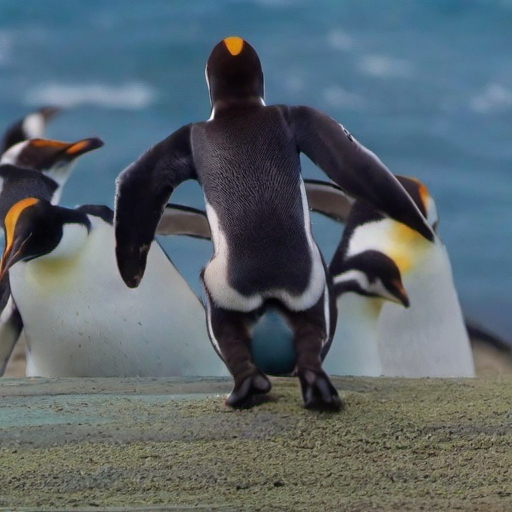}
    \caption*{}
\end{subfigure}
\hfill
\begin{subfigure}[b]{0.12\textwidth}
    \includegraphics[width=\textwidth]{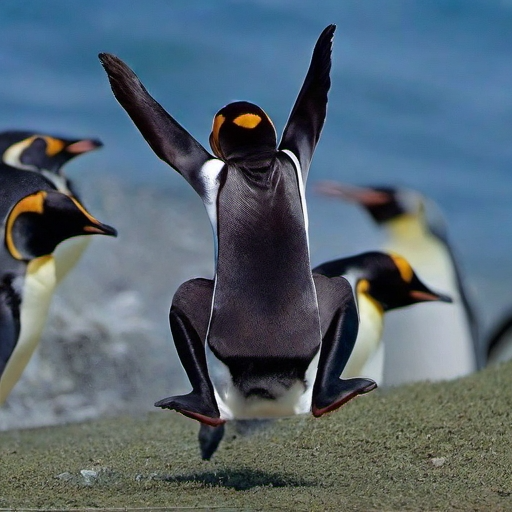}
    \caption*{}
\end{subfigure}
\hfill
\begin{subfigure}[b]{0.12\textwidth}
    \includegraphics[width=\textwidth]{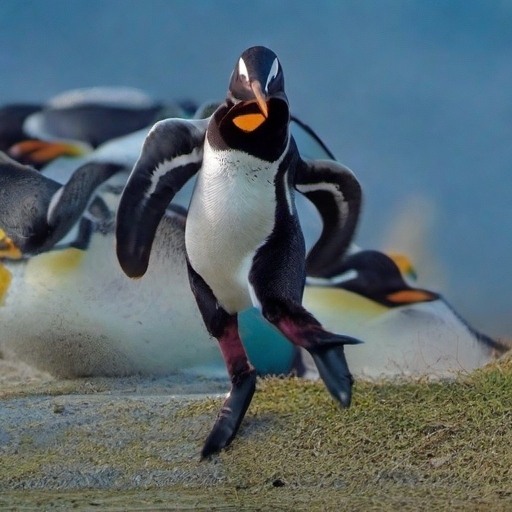}
    \caption*{}
\end{subfigure}
\hfill
\begin{subfigure}[b]{0.12\textwidth}
    \includegraphics[width=\textwidth]{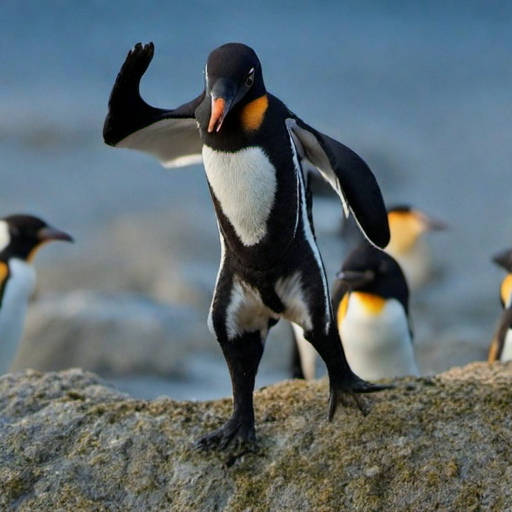}
    \caption*{}
\end{subfigure}
\\[-1.0em]
\begin{subfigure}[b]{0.12\textwidth}
    \includegraphics[width=\textwidth]{img/main_qual/1x1/1x1_2.jpg}
    \caption*{}
\end{subfigure}
\hfill
\begin{subfigure}[b]{0.12\textwidth}
    \includegraphics[width=\textwidth]{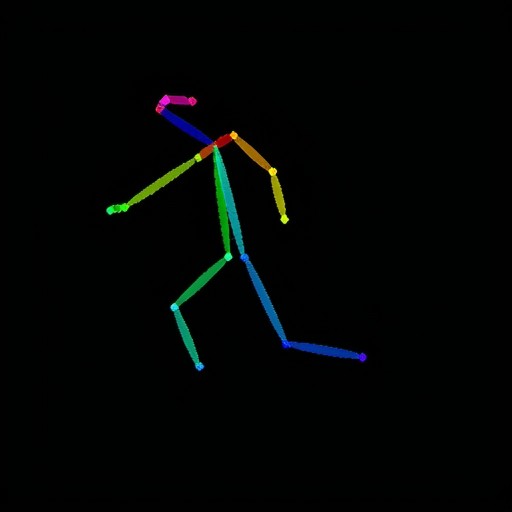}
    \caption*{}
\end{subfigure}
\hfill
\begin{subfigure}[b]{0.12\textwidth}
    \includegraphics[width=\textwidth]{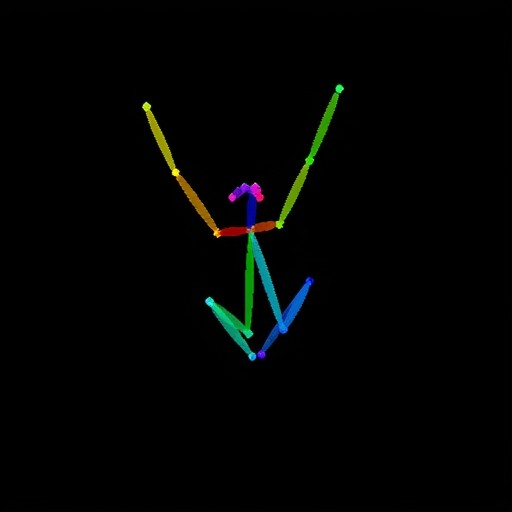}
    \caption*{}
\end{subfigure}
\hfill
\begin{subfigure}[b]{0.12\textwidth}
    \includegraphics[width=\textwidth]{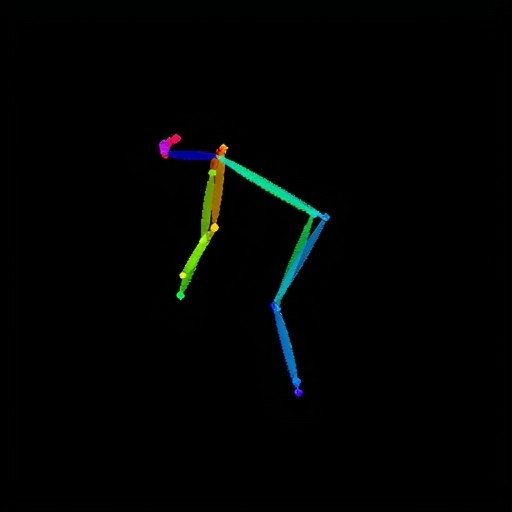}
    \caption*{}
\end{subfigure}
\hfill
\begin{subfigure}[b]{0.12\textwidth}
    \includegraphics[width=\textwidth]{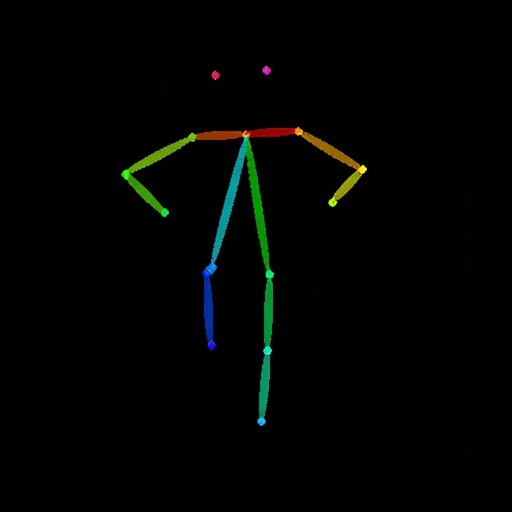}
    \caption*{}
\end{subfigure}
\hfill
\begin{subfigure}[b]{0.12\textwidth}
    \includegraphics[width=\textwidth]{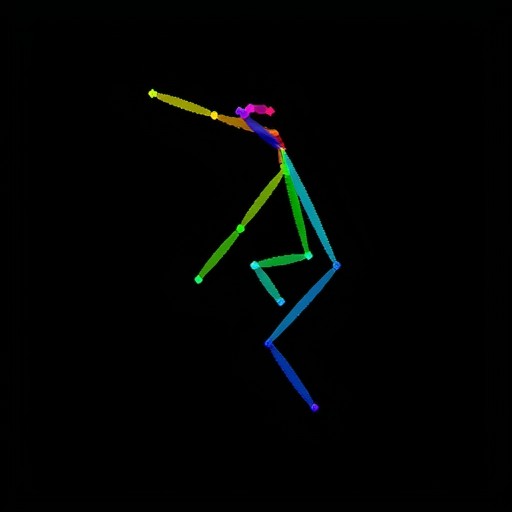}
    \caption*{}
\end{subfigure}
\hfill
\begin{subfigure}[b]{0.12\textwidth}
    \includegraphics[width=\textwidth]{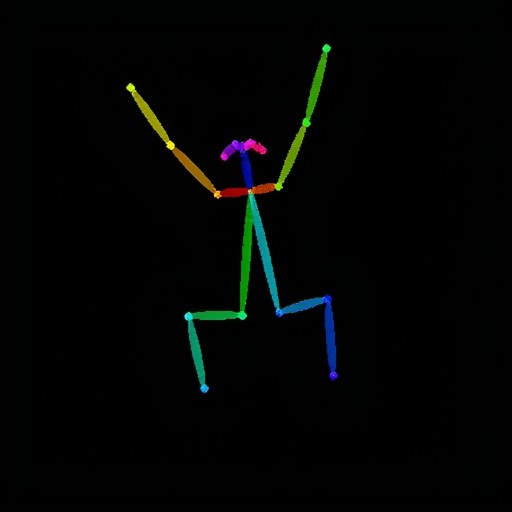}
    \caption*{}
\end{subfigure}
\hfill
\begin{subfigure}[b]{0.12\textwidth}
    \includegraphics[width=\textwidth]{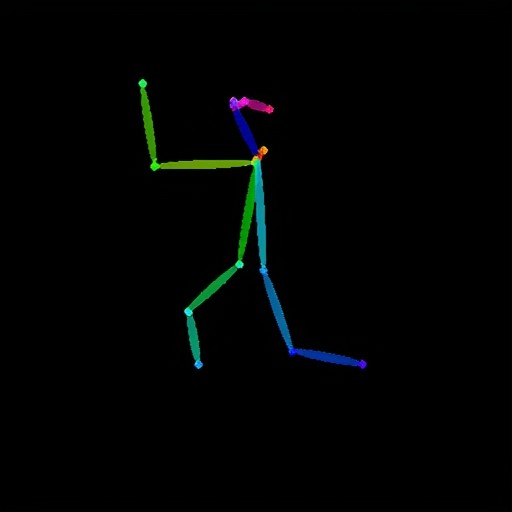}
    \caption*{}
\end{subfigure}
\\[-1.0em]
\begin{subfigure}[b]{0.12\textwidth}
    \includegraphics[width=\textwidth]{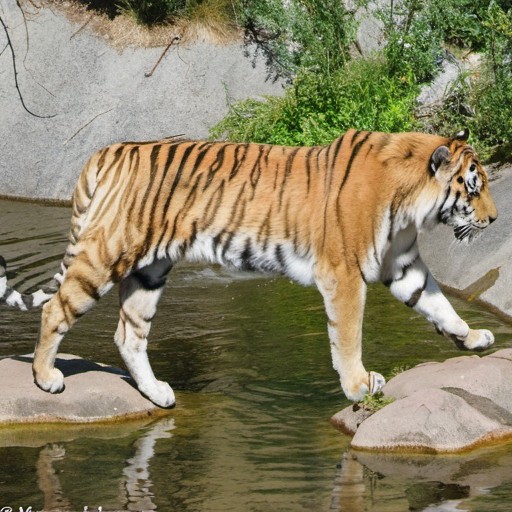}
    \caption*{}
\end{subfigure}
\hfill
\begin{subfigure}[b]{0.12\textwidth}
    \includegraphics[width=\textwidth]{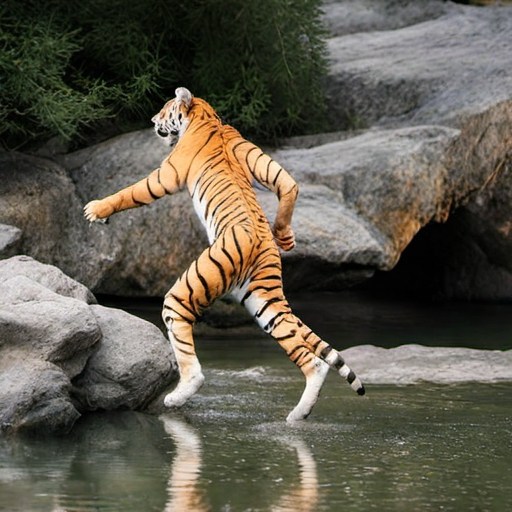}
    \caption*{}
\end{subfigure}
\hfill
\begin{subfigure}[b]{0.12\textwidth}
    \includegraphics[width=\textwidth]{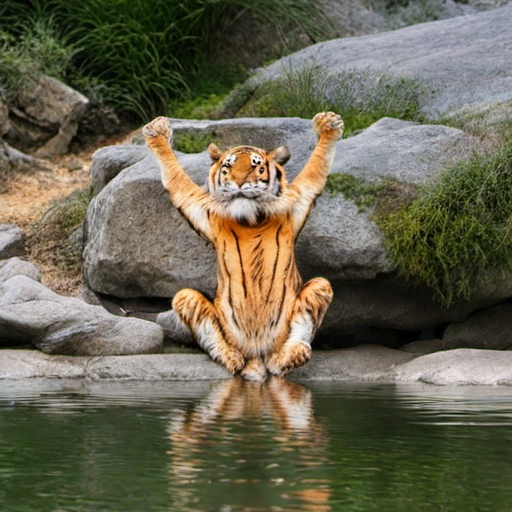}
    \caption*{}
\end{subfigure}
\hfill
\begin{subfigure}[b]{0.12\textwidth}
    \includegraphics[width=\textwidth]{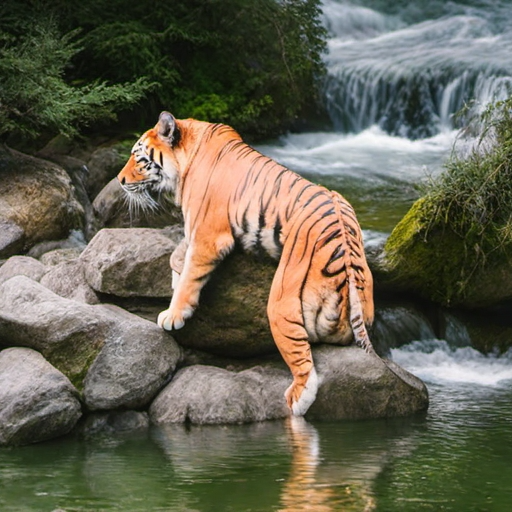}
    \caption*{}
\end{subfigure}
\hfill
\begin{subfigure}[b]{0.12\textwidth}
    \includegraphics[width=\textwidth]{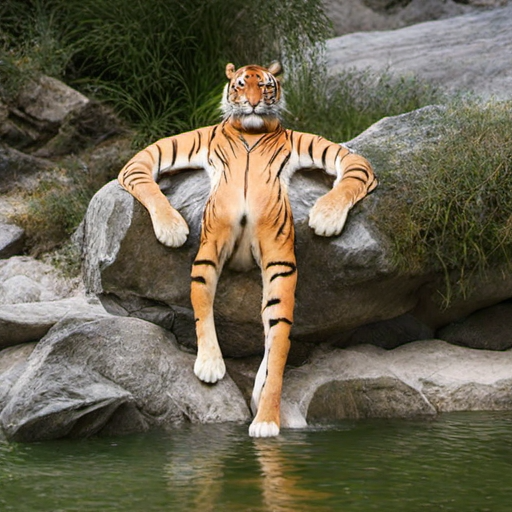}
    \caption*{}
\end{subfigure}
\hfill
\begin{subfigure}[b]{0.12\textwidth}
    \includegraphics[width=\textwidth]{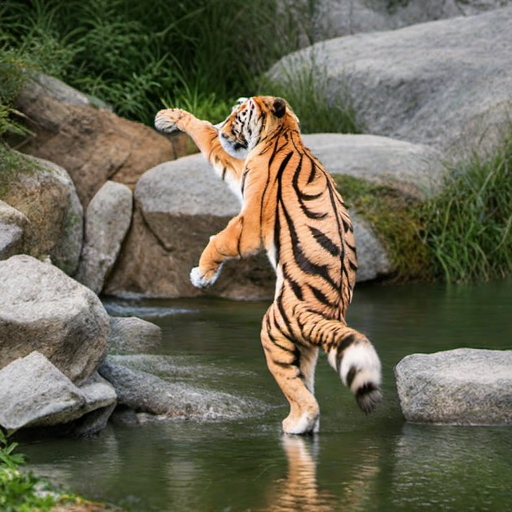}
    \caption*{}
\end{subfigure}
\hfill
\begin{subfigure}[b]{0.12\textwidth}
    \includegraphics[width=\textwidth]{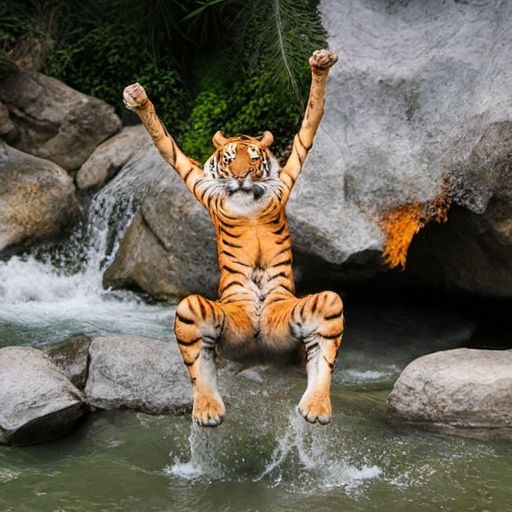}
    \caption*{}
\end{subfigure}
\hfill
\begin{subfigure}[b]{0.12\textwidth}
    \includegraphics[width=\textwidth]{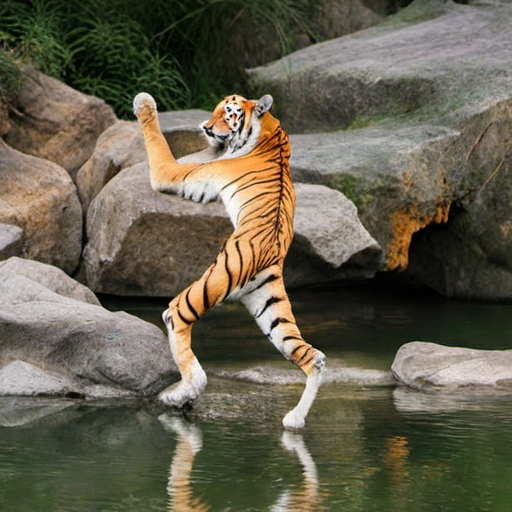}
    \caption*{}
\end{subfigure}
\vspace{-18pt}
\caption{\textbf{Cross-domain Pose transfer via Dual Recursive Feedback(DRF).} Our DRF demonstrates robust pose generations based on structures (mesh, skeleton). The first row illustrates human pose structures and their application to penguin (appearance) images, generating semantically consistent with a given penguin appearance. Similarly, the second example presents human skeleton-based structures applied to a tiger (appearance), demonstrating the adaptability of motion synthesis across different poses. }
\vspace{-14pt}
\label{fig:mesh_penguin}
\end{figure*}

Another strategy for reflecting both the structure to be incorporated into the generated image and the appearance to be preserved is to use dedicated images as prompts for each aspect, such as IP-Adapter~\cite{ye2023ip} and Ctrl-X~\cite{lin2025ctrl}. By directly specifying what structures and appearance the user wants, these methods can better integrate the user's intent into the generated results. However, such techniques still suffer from the need to retrain models every time or the dependence on categorical similarity between the two images. As a result, they may fail to generate appropriate images in class-invariant setups where the categories of structure and appearance are very different or where there is a significant structural mismatch between structure and appearance. This indicates that the conventional approaches do not offer a universally reliable solution yet.\\
\indent To address these issues, we propose a model called \textit{Dual Recursive Feedback (DRF)} that successfully exploits the flexibility of score guidance to synthesize the image with the given structure and appearance, even under extreme conditions. 
A structure image and an appearance image are provided as inputs, thereby eliminating the need for additional training or guidance mechanisms.
Then, we recursively refine the resulting latent space by adjusting a text-conditioned score describing the appearance to accurately reflect the identity, as shown in \cref{fig:teasor}. 
Owing to appearance feedback, the appearance leakage can be resolved even if the structure and appearance differ significantly.
Furthermore, latent values derived from injecting both images are also updated to align with the user's intended prompt, which is referred to as generation feedback. By recursively applying appearance and generation feedbacks to the intermediate latents, the final output with the proposed method enables to reflect the user's intent. Moreover, our framework is compatible with various T2I models, where it consistently outperforms the baseline models in terms of quality.

In summary, our main contributions are as follows:
\begin{itemize}

     \item We present a novel method that employs dual recursive feedback to preserve appearance fidelity and structural integrity while excelling at pose transfer, enabling high-quality image generation that accurately reflects user intent—even in class-invariant setups. 

    \item Our method not only generates images where the appearance image is adapted to follow the structure outlined by the structure image, but also allows for diverse scenarios by manipulating the prompt to align the structure's object: guiding the structure image to adopt the textures and colors of the appearance image.

\end{itemize}
\vspace{-2pt}
\begin{figure*}[!t]
        \centering
        \includegraphics[width=0.98\textwidth]{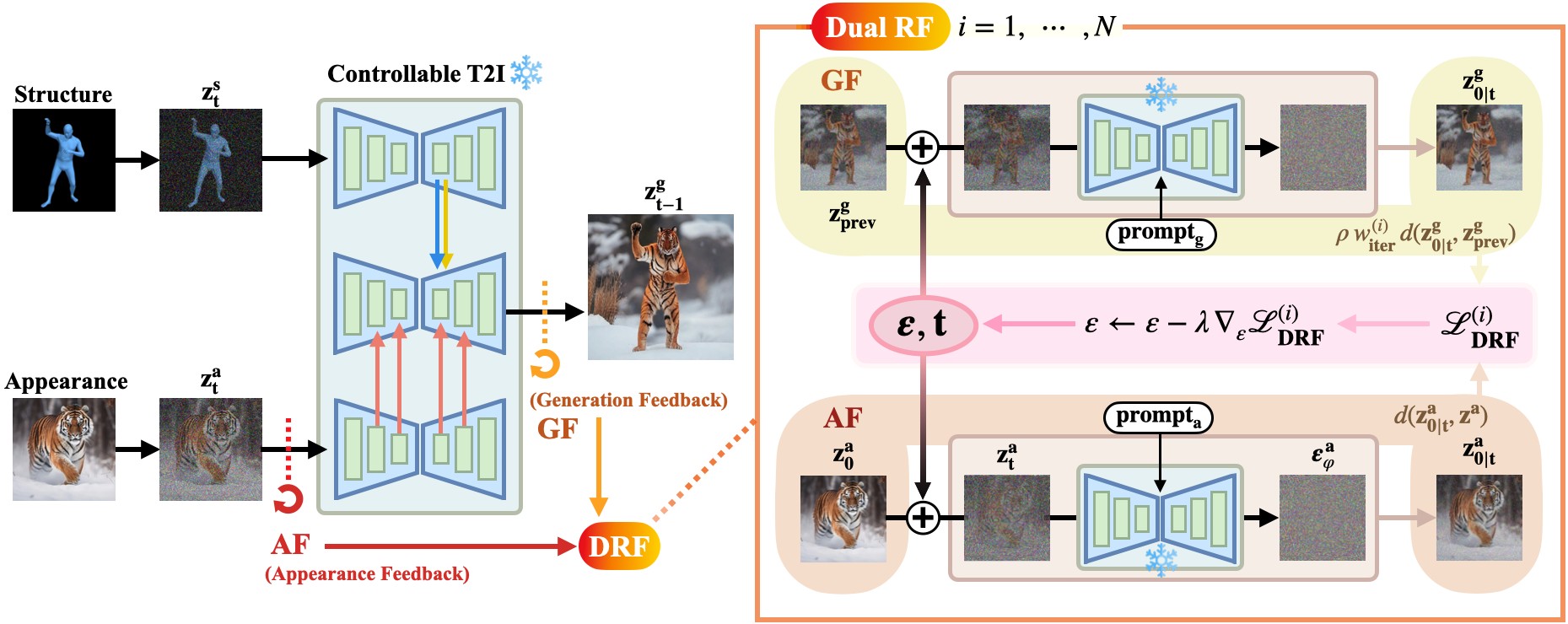}
    \vspace{-5pt}
    \caption{\textbf{Flowchart of DRF.} The backbone of our algorithm starts from the generation latent obtained from the Ctrl-X~\cite{lin2025ctrl} framework. DRF iteratively updates by obtaining guided noise $\epsilon_{\theta}^a$ and $\epsilon_{\theta}^g$ through appearance and generation feedback. The distillation function derived from these two noises is combined to update the generation latent.}
    \vspace{-14pt}
    \label{fig:pipeline}
\end{figure*}

\section{Related works}
\label{sec:related}

\subsection{Controllable Text-to-Image diffusion}
Text-to-Image (T2I) models have continued to evolve by incorporating conditional guidance methods for more precise image generation. ControlNet~\cite{zhang2023adding}, which augments a pretrained T2I diffusion model with spatial conditional control, allowing users to specify additional constraints—such as layout, pose, and shape—to reflect a desired spatial composition. However, this approach requires fine-tuning multiple models when controlling numerous spatial factors, thereby increasing both model count and training overhead. By contrast, FreeControl~\cite{mo2024freecontrol} imposes no additional training demands on the pretrained T2I model and generally adheres well to textual descriptions; however, it depends solely on gradient-based manipulation in the latent space to minimize loss, resulting in significant computational costs and GPU usage. To facilitate faster and more lightweight T2I operations, Ctrl-X~\cite{lin2025ctrl} has been introduced, injecting separate images for structure and appearance into a pretrained diffusion model, using feed-forward feature replacement to preserve structure and self-attention layers to align generated features with appearance image statistics. While this method increased efficiency, it remains less effective at consistently maintaining the target appearance features. For instance, though Ctrl-X demonstrates robust performance when both source images belong to the same category (\textit{e.g.,} appearance: ``horse", structure: ``a mesh of horse"), its performance degrades when dealing with class-invariant setups, resulting in subpar generation results.
\vspace{-5pt}
\subsection{Score distillation sampling}
Score Distillation Sampling (SDS)~\cite{poole2022dreamfusion} is a direct, gradient-based optimization approach to diffusion models. The pretrained text-image diffusion model’s implicit image distribution and denoising-direction score are inverted to adjust images toward a desired text condition iteratively. Due to the noisy gradients in SDS, which frequently cause oversaturated colors and overly smoothed regions. Delta Denoising Score (DDS)~\cite{hertz2023delta} and Identity-preserving Distillation Sampling (IDS)~\cite{kim2025identity} were subsequently introduced to mitigate these effects. Notably, IDS employs a fixed-point regularization (FPR) method, which preserves the original image’s identity throughout the editing process, resulting in high-quality image generation. Nevertheless, because FPR strongly constrains the source image to maintain its identity, the degree of editing becomes limited, yielding lower prompt alignment compared to prior approaches~\cite{poole2022dreamfusion,hertz2023delta,nam2024contrastive}.

\section{Methodology}
\vspace{-2pt}
\label{sec:method}
\subsection{Preliminaries}
\vspace{-2pt}
\noindent\textbf{Diffusion models and sampling guidance.}\quad
The forward diffusion process is to gradually add Gaussian noise, $\epsilon\sim\mathcal{N}(0, \mathbf{I})$, to a clean image \(\mathbf{z}_0\) across \(T\) timesteps:
\begingroup
\setlength{\abovedisplayskip}{4pt}   
\setlength{\belowdisplayskip}{4pt}   
\begin{equation} \label{eq:forward}
    \mathbf{z}_t = \sqrt{\alpha_t}\,\mathbf{z}_0 + \sqrt{1-\alpha_t}\,\epsilon, \quad t\in\{1,\dots,T\},
\end{equation}
\endgroup
where $\alpha_t$ is the cumulative product of noise variances at step \(t\), and \(\mathbf{z}_t\) denotes the resulting noisy image at that step. Then, the diffusion model learns to denoise the data step by step
to reconstruct the output image.
Sampling guidance modifies the standard denoising procedure to incorporate user constraints. In classifier-free guidance (CFG)~\cite{ho2022classifier}, the model obtains both conditional ${\epsilon}(x_t|{c})$ and unconditional ${\epsilon}(x_t|{\varnothing})$ noise predictions, leveraging their difference to strongly enforce the desired condition:
\begingroup
\setlength{\abovedisplayskip}{5pt}   
\setlength{\belowdisplayskip}{5pt}   
\begin{equation} \label{eq:cfg}
    {\epsilon}_\theta^\omega (\mathbf{z}_t,y,t) 
    = (1+\omega)\epsilon_\theta(\mathbf{z}_t,y,t) 
    -\omega\epsilon_\theta(\mathbf{z}_t,\varnothing,t).
\end{equation}
\endgroup
\noindent\textbf{Score guidance models.}\quad
Approaches like IDS~\cite{kim2025identity} tackle directional ambiguity by preserving the identity of the original image. 
In IDS, FPR is proposed to modify the current latent for the adjustment of the text-conditioned score to the original latent as follows:
\begingroup
\setlength{\abovedisplayskip}{5pt}   
\setlength{\belowdisplayskip}{4pt}   
\begin{equation}\label{eq:update}
    \mathbf{z}^{\text{org}}_t \leftarrow \mathbf{z}_{t}^{\text{org}} - \lambda \nabla_{\mathbf{z}^{\text{org}}_t}\mathcal{L}_{\text{FPR}},
\end{equation}
\endgroup
where
\begingroup
\setlength{\abovedisplayskip}{4pt}   
\setlength{\belowdisplayskip}{4pt}   
\begin{equation}
\mathcal{L}_{\text{FPR}} = d ( \mathbf{z}^{\text{org}}, \mathbf{z}_{0|t}^\text{org}).
\label{eq:ids}
\end{equation}
\endgroup
$d(\mathbf{x}, \mathbf{y})$ denotes any metric to compare $\mathbf{x}$ and $\mathbf{y}$. $\mathbf{z}_{0|t}^\text{org}$ is the posterior mean associated with the original image given by:
\begingroup
\setlength{\abovedisplayskip}{4pt}   
\setlength{\belowdisplayskip}{4pt}   
{\small
\begin{equation}
\mathbf{z}_{0\mid t}^{\text{org}} =  \mathbb{E}[\mathbf{z}_0^{\text{org}}\mid\mathbf{z}_t^{\text{org}}]=\frac{1}{\sqrt{\alpha_{t}}}\left(
    \mathbf{z_t^{\text{org}}} -\sqrt{1-\alpha_{t}} {\epsilon}_{\theta}^{\omega}
    \right).
\label{eq:posterior}
\end{equation}
}
\endgroup
However, this strong correction overly retains the original content, limiting the generated image’s ability to align with the target prompt.

\vspace{3pt}

\noindent \textbf{Notations.}\quad
For the controllable Text-to-Image (T2I) generation using the pretrained diffusion models, the appearance image $\mathbf{I}^a$ and the structure image $\mathbf{I}^s$ can be used.
For the sake of simplicity, the notations for images, latents and other components required during the image synthesis are summarized in \cref{tab:notation}.
\begin{table}[H]
\centering
\renewcommand{\arraystretch}{1.3}
\scalebox{0.52}{%
\begin{tabular}{l l l l}
\toprule
\textbf{Notation} & \textbf{Description} & \textbf{Notation} & \textbf{Description}\\
\midrule
$\mathbf{I}^g$            & Generation image                  
                 & $\mathbf{z}_{t-1}^g$    & Generation latent               \\
$\mathbf{I}^a$             & Appearance image                    
                 & $\mathbf{z}_0^a$            & Clean latent of $\mathbf{I}^a$                \\
$\mathbf{I}^s$     & Structure image               
                 & $\mathbf{z}_t^a$ & Noisy latent of $\mathbf{I}^a$ at time $t$ \\
$f_t^s$           & Convolutional feature from $\mathbf{I}^s$  
                 & $\mathbf{z}_{\text{prev}}^g$           & Previous generation latent in recursion  \\
$h_t^a$           & Self-attention-based appearance feature
                 & $A_t^s$ & Self-attention map derived from $\mathbf{I}^s$\\
\bottomrule
\end{tabular}
}
\vspace{-10pt}
\caption{Notation Summary for DRF}
\vspace{-10pt}
\label{tab:notation}
\end{table}

\subsection{Dual Recursive Feedback (DRF)}
\begin{figure}[!t] 
\footnotesize
\centering 

\newcommand{\imgwidth}{0.18\linewidth} 

\begin{tikzpicture}[x=1cm, y=1cm]
    \node[anchor=south] (FigA1) at (0,0) {
        \includegraphics[width=\imgwidth]{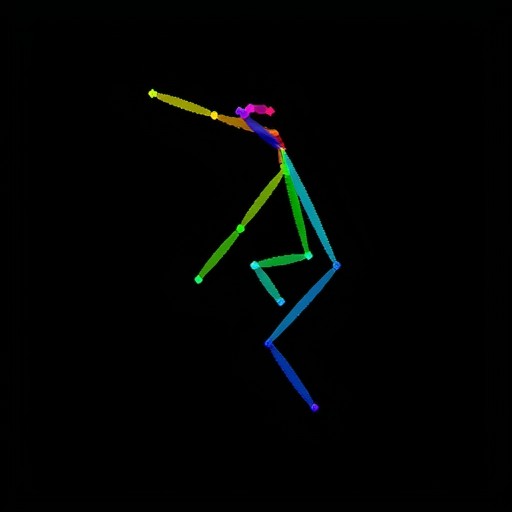}
    };
    \node[anchor=south, yshift=-1mm] at (FigA1.north) {\scriptsize Structure};
\end{tikzpicture}\hspace{-1mm}%
\begin{tikzpicture}[x=1cm, y=1cm]
    \node[anchor=south] (FigD1) at (0,0) {
        \includegraphics[width=\imgwidth]{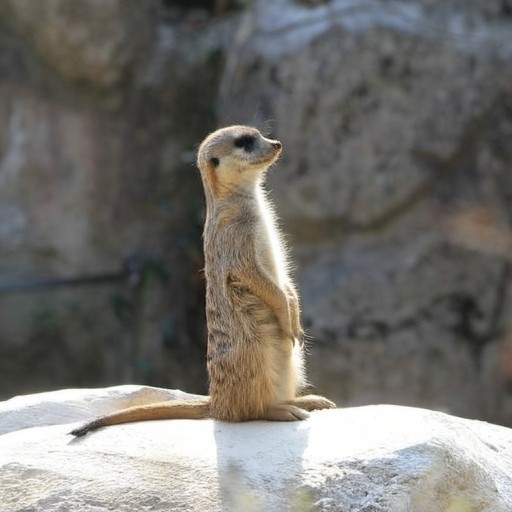}
    };
    \node[anchor=south, yshift=-1mm] at (FigD1.north) {\scriptsize Appearance};
\end{tikzpicture}\hspace{-1mm}%
\begin{tikzpicture}[x=1cm, y=1cm]
    \node[anchor=south] (FigE1) at (0,0) { 
        \includegraphics[width=\imgwidth]{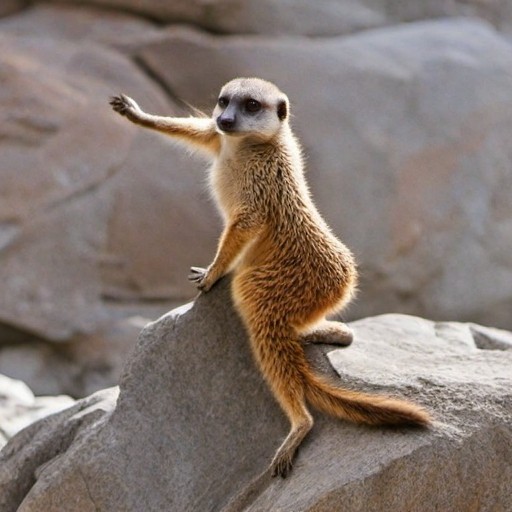}
    };
    \node[anchor=south, yshift=-1mm] at (FigE1.north) {\scriptsize AF+GF(\textbf{DRF})}; 
\end{tikzpicture}\hspace{-1mm}%
\begin{tikzpicture}[x=1cm, y=1cm]
    \node[anchor=south] (FigB1) at (0,0) {
        \includegraphics[width=\imgwidth]{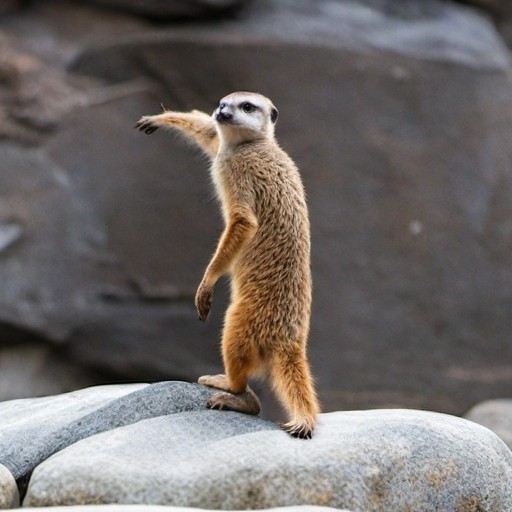}
    };
    \node[anchor=south, yshift=-1mm] at (FigB1.north) {\scriptsize AF};
\end{tikzpicture}\hspace{-1mm}%
\begin{tikzpicture}[x=1cm, y=1cm]
    \node[anchor=south] (FigC1) at (0,0) {
        \includegraphics[width=\imgwidth]{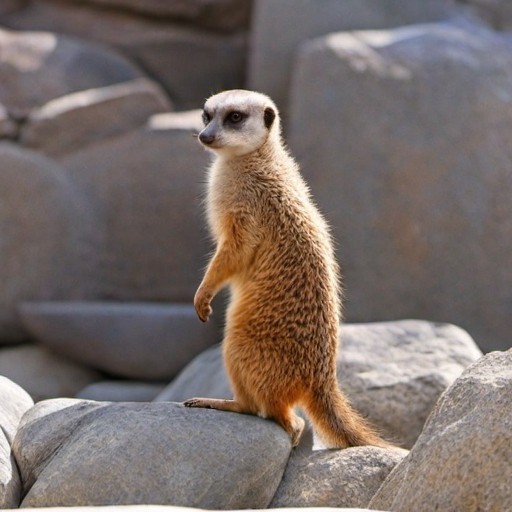}
    };
    \node[anchor=south, yshift=-1mm] at (FigC1.north) {\scriptsize w/o DRF};
\end{tikzpicture}

\vspace{-8pt}
\caption{\textbf{Comparison of Appearance feedback only and DRF.} When only Appearance feedback is executed without Generation feedback, the excessive emphasis on appearance image occurs.}
\vspace{-12pt}
\label{fig:app_feedback}
\end{figure}
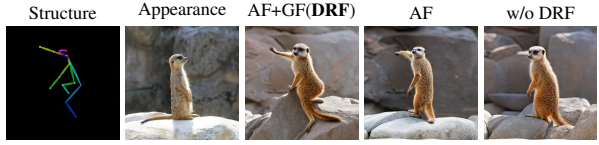

The flowchart of the proposed method is illustrated in \cref{fig:pipeline}. In T2I diffusion models where $\mathbf{I}^a$ and $\mathbf{I}^s$ serve as controllable elements, our method robustly transforms the generated image $\mathbf{I}^g$ on a class-invariant dataset while retaining the features of $\mathbf{I}^a$ and adhering to the structure of $\mathbf{I}^s$.  \\
\indent We begin by injecting $\mathbf{I}^a$ and $\mathbf{I}^s$ into the T2I model without guidance and then focus on the resulting output latent $\mathbf{z}^g$ as in \cite{lin2025ctrl}. Specifically, $\mathbf{z}^g$ incorporates $f_t^s$ and $A_t^s$ reflecting structural information from $\mathbf{I}^s$, as well as an appearance latent $h_t^a$ transferred from $\mathbf{I}^a$. Hence, $\mathbf{z}^g$ can be expressed as:
\begingroup
\setlength{\abovedisplayskip}{4pt}   
\setlength{\belowdisplayskip}{4pt}   
\begin{equation}
\mathbf{z}_{t-1}^g = \mbox{Ctrl-X}(\mathbf{z}_t^g\mid t, y, f_t^s, A_t^s, h_t^a ).
\end{equation}
\endgroup
We observed that when $\mathbf{I}^a$ and $\mathbf{I}^s$ belong to class-invariant setup--different categories or exhibiting significant structural discrepancies--, simply separating appearance and structural information within the attention map is insufficient to yield $\mathbf{I}^g$ that correctly reflects both features of $\mathbf{I}^a$ and $\mathbf{I}^s$, as shown in the third column of \cref{fig:app_feedback}. To address this limitation, we propose leveraging a score guidance method on $\mathbf{z}^g$ during the generation. 
We refer to this technique as \textbf{Dual Recursive Feedback (DRF)}.
\vspace{-5pt}
\vspace{8pt}

\noindent\textbf{Appearance Feedback.}\quad
To address “appearance leakage” issue reported in FreeControl~\cite{mo2024freecontrol} and misalignment of appearance features noted in Ctrl-X~\cite{lin2025ctrl}, gradient based on the latents associated with the appearance, such as $\mathbf{z}_t^a$ and $\mathbf{z}^a$, needs to be investigated.
As analyzed in IDS~\cite{kim2025identity}, the text-conditioned score $\epsilon_\theta^a$ computed from $\mathbf{z}_t^a$ cannot always be guaranteed as a gradient to $\mathbf{z} ^a$, leading to a loss of appearance identity.
Thus, our method employs the concept of a \textit{fixed point}---inspired by numerical analysis---to guide the gradient updates toward $\mathbf{z}^a$. 
\vspace{4pt}
\begin{figure}[!t] 
\scriptsize
\centering 

\newcommand{\imgwidth}{0.23\linewidth} 

\begin{tikzpicture}[x=1cm, y=1cm]
    \node[anchor=south] (FigA1) at (0,0) {
        \includegraphics[width=\imgwidth]{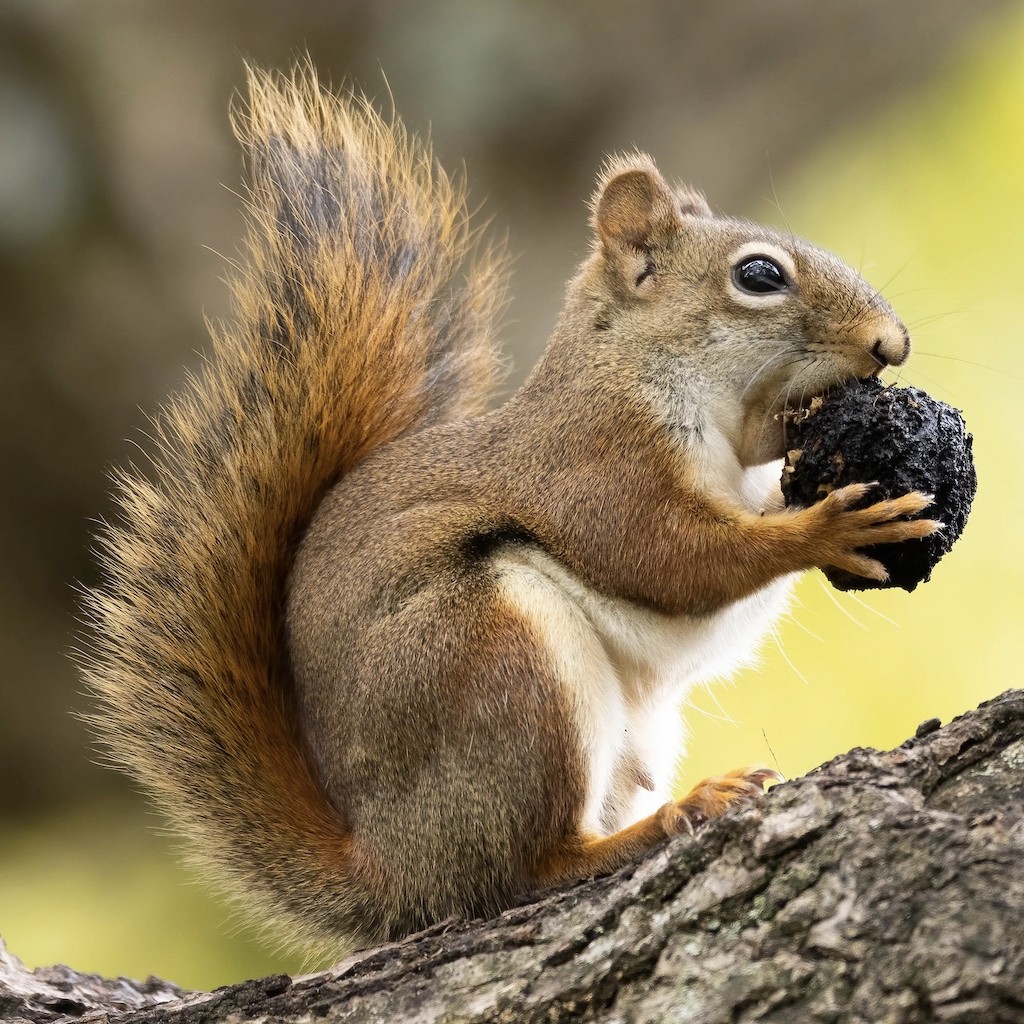}
    };
    \node[anchor=south, yshift=-1mm] at (FigA1.north) {\scriptsize Structure};
\end{tikzpicture}\hspace{-1mm}%
\begin{tikzpicture}[x=1cm, y=1cm]
    \node[anchor=south] (FigD1) at (0,0) {
        \includegraphics[width=\imgwidth]{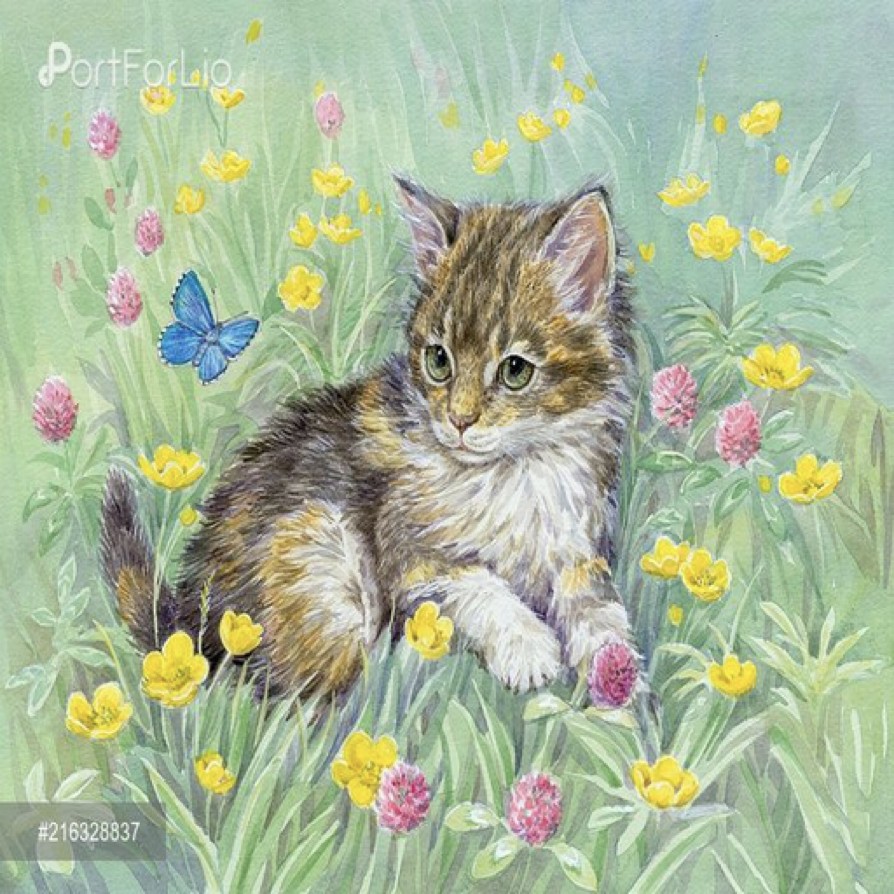}
    };
    \node[anchor=south, yshift=-1mm] at (FigD1.north) {\scriptsize Appearance};
\end{tikzpicture}\hspace{-1mm}%
\begin{tikzpicture}[x=1cm, y=1cm]
    \node[anchor=south] (FigC1) at (0,0) {
        \includegraphics[width=\imgwidth]{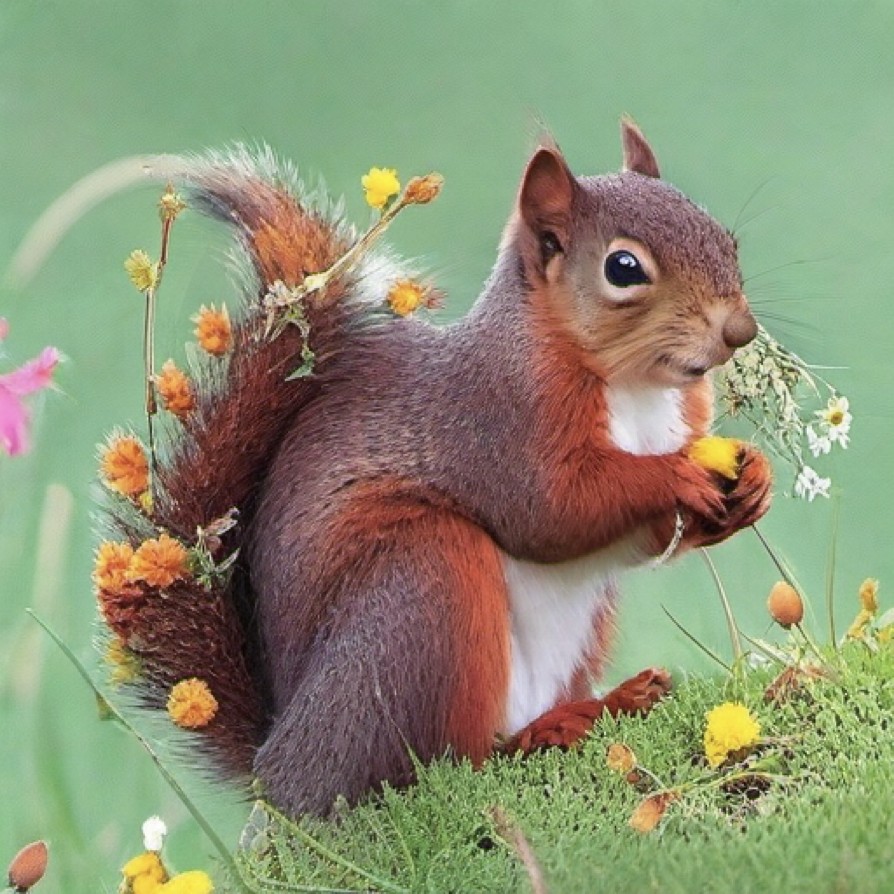}
    };
    \node[anchor=south, yshift=-1mm] at (FigD1.north) {\scriptsize \textbf{DRF \cref{eq:sim_app_latent}}};
\end{tikzpicture}\hspace{-1mm}%
\begin{tikzpicture}[x=1cm, y=1cm]
    \node[anchor=south] (FigB1) at (0,0) {
        \includegraphics[width=\imgwidth]{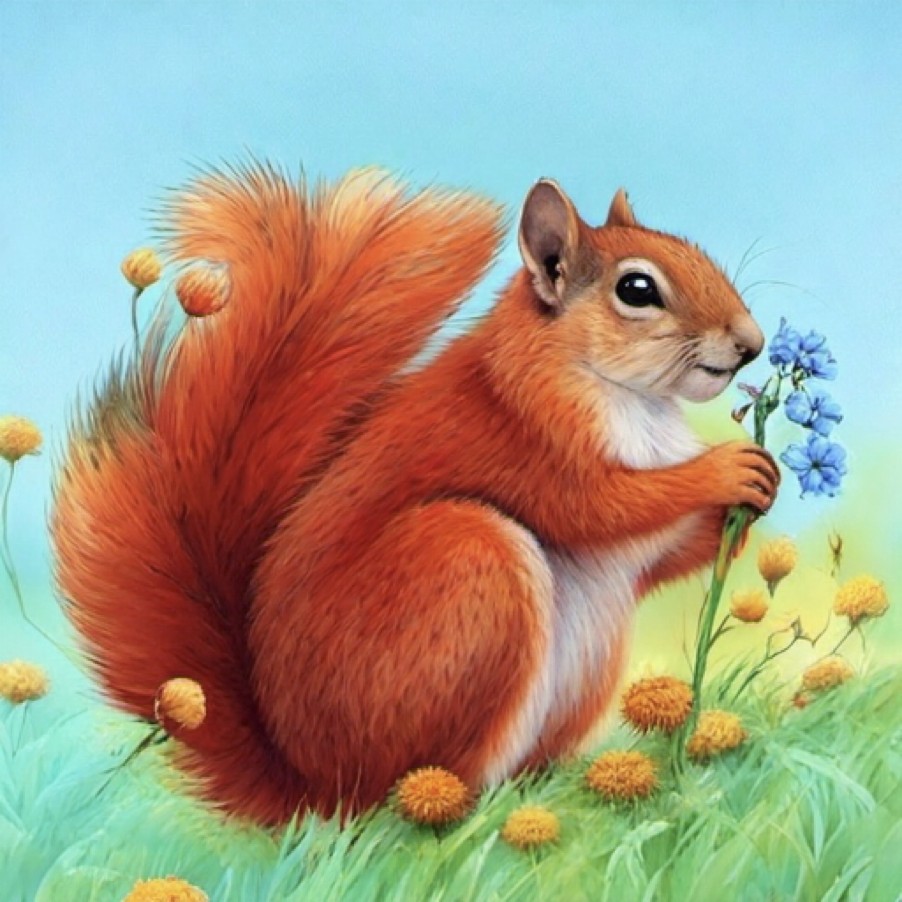}
    };
    \node[anchor=south, yshift=-1mm] at (FigD1.north) {\scriptsize Forward \cref{eq:forward}};
\end{tikzpicture}
\vspace{-3pt}

\begin{tikzpicture}[x=1cm, y=1cm]
    \node[anchor=south] (FigA1) at (0,0) {
        \includegraphics[width=\imgwidth]{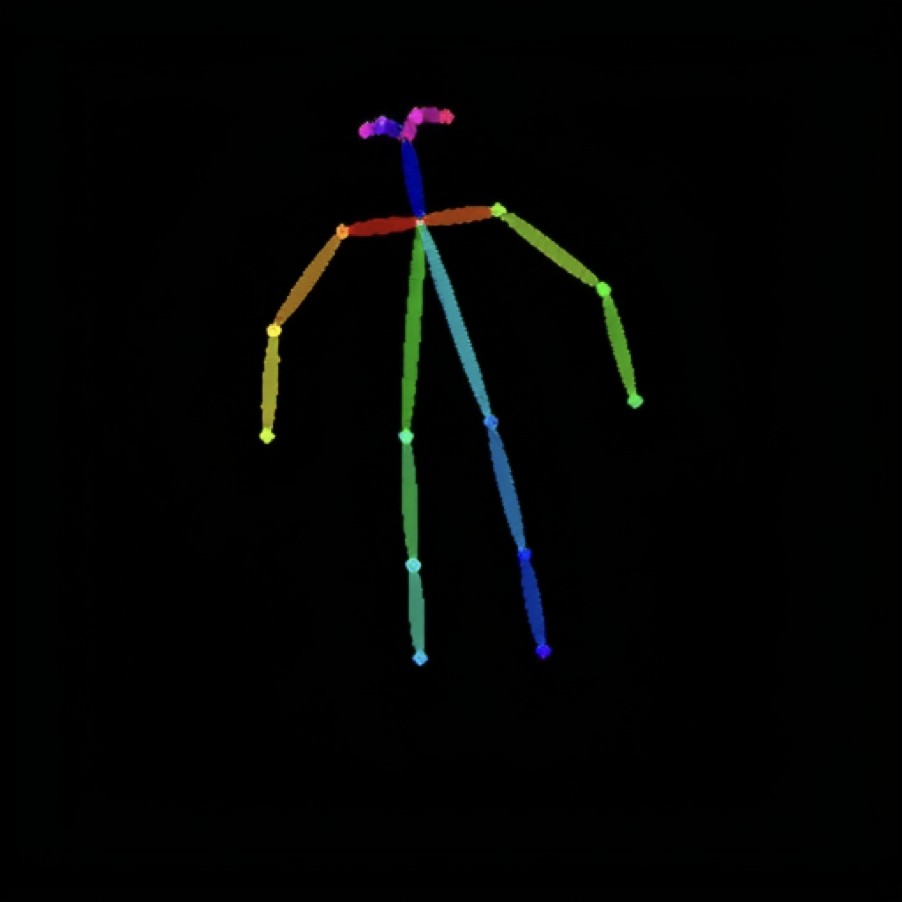}
    };
\end{tikzpicture}\hspace{-1mm}%
\begin{tikzpicture}[x=1cm, y=1cm]
    \node[anchor=south] (FigD1) at (0,0) {
        \includegraphics[width=\imgwidth]{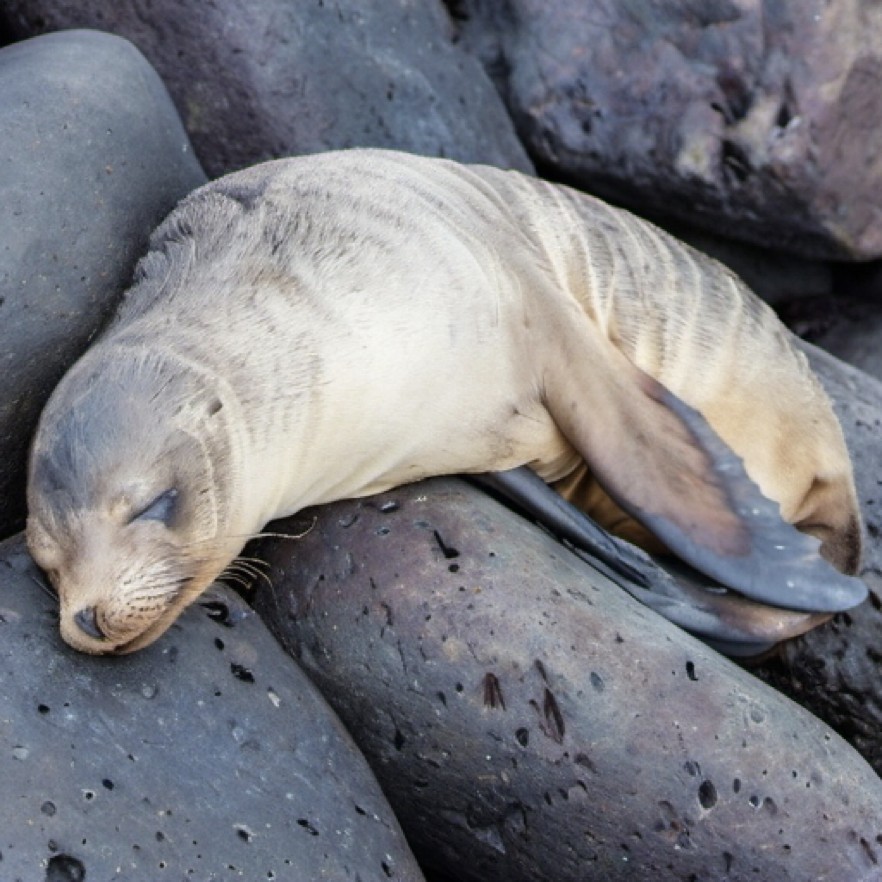}
    };
\end{tikzpicture}\hspace{-1mm}%
\begin{tikzpicture}[x=1cm, y=1cm]
    \node[anchor=south] (FigC1) at (0,0) {
        \includegraphics[width=\imgwidth]{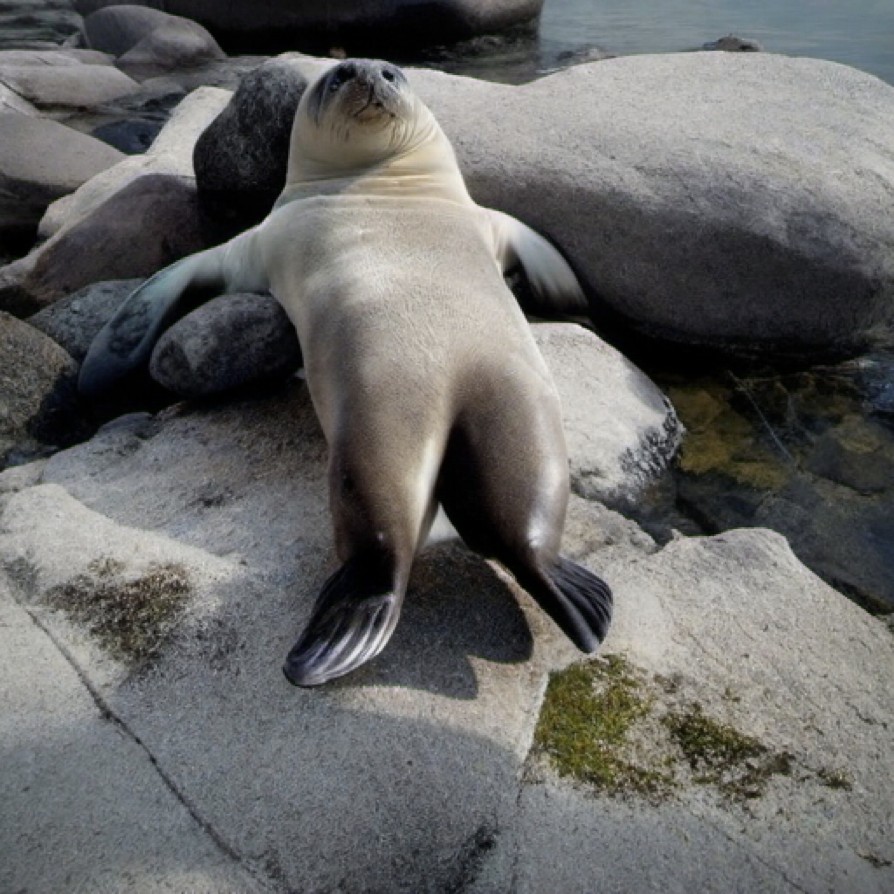}
    };
\end{tikzpicture}\hspace{-1mm}%
\begin{tikzpicture}[x=1cm, y=1cm]
    \node[anchor=south] (FigB1) at (0,0) {
        \includegraphics[width=\imgwidth]{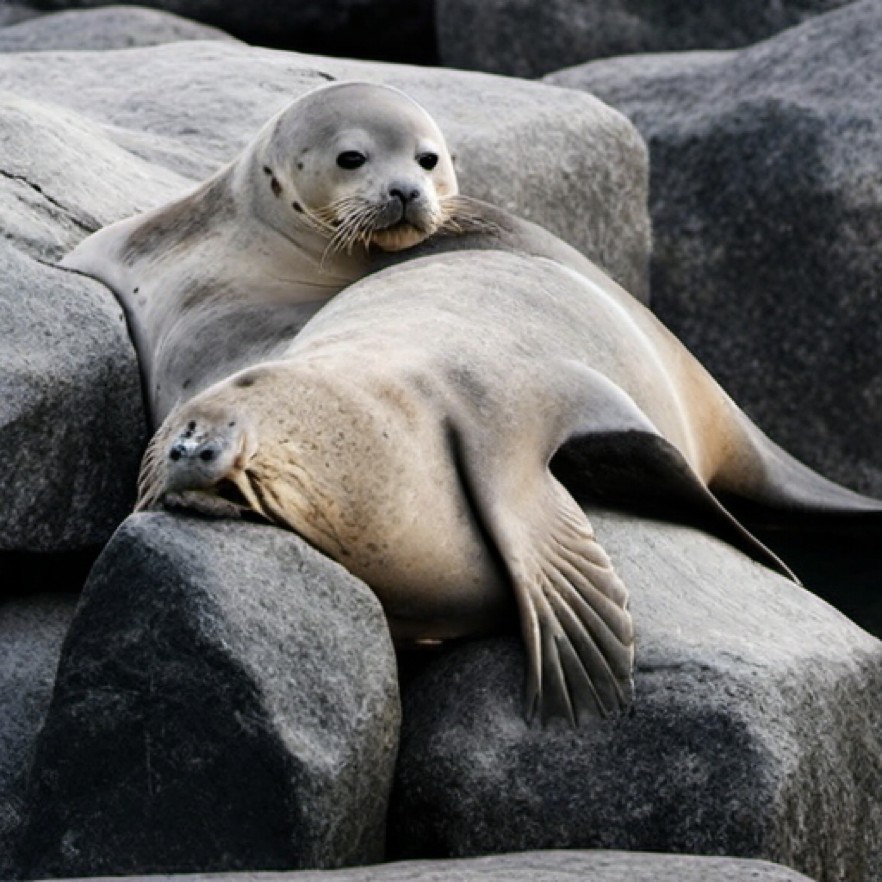}
    };
\end{tikzpicture}

\vspace{-8pt}
\caption{\textbf{Outputs according to the forward process of appearance latent}. When structure and appearance images are given, \textit{third} and \textit{fourth column} show the outputs of DRF with $\tilde{\mathbf{z}}_t^a$ and $\mathbf{z}_t^a$ for appearance feedback.}
\vspace{-15pt}
\label{fig:diffusion_process_qual}
\end{figure}
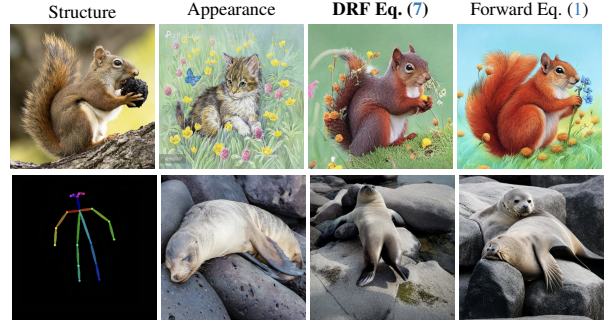

\vspace{7pt}
\begin{figure*}[t]
\centering
\resizebox{\textwidth}{!}{%
\setlength{\tabcolsep}{1.2pt}
\arrayrulecolor{gray}        
\setlength{\arrayrulewidth}{1pt}  
\setlength{\dashlinedash}{3pt}    
\setlength{\dashlinegap}{2pt}     
\begin{tabular}{>{\centering\arraybackslash}m{0.125\textwidth}
    >{\centering\arraybackslash}m{0.125\textwidth}
    >{\centering\arraybackslash}m{0.125\textwidth}
    >{\centering\arraybackslash}m{0.125\textwidth}
    >{\centering\arraybackslash}m{0.125\textwidth}
    >{\centering\arraybackslash}m{0.125\textwidth}
    >{\centering\arraybackslash}m{0.125\textwidth}
    >{\centering\arraybackslash}m{0.125\textwidth}}

\multicolumn{2}{c}{} 
& \multicolumn{3}{c}{\textbf{Training - free methods}} 
& \multicolumn{3}{c}{\textbf{Training - based methods}} 
\\
  \arrayrulecolor{SpringGreen}
  \cline{3-5}
  \arrayrulecolor{gray}
  \cline{6-8}
  \arrayrulecolor{gray}

    \footnotesize{Structure} & \footnotesize{Appearance} & \footnotesize \textbf{DRF (Ours)} & \footnotesize Ctrl-X\cite{lin2025ctrl} & \footnotesize FreeControl\cite{mo2024freecontrol}\ & \footnotesize Uni-Control\cite{qin2023unicontrol} & \footnotesize ControlNet\cite{zhang2023adding} + IP-Adapter\cite{ye2023ip} & \footnotesize T2I-Adapter\cite{mou2024t2i} + IP-Adapter\cite{ye2023ip} \\
    \includegraphics[width=\linewidth]{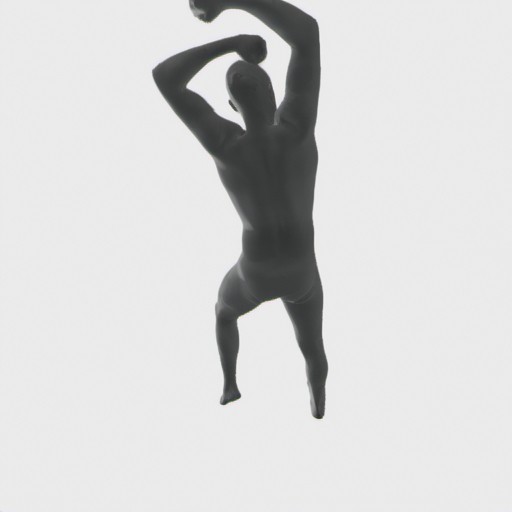} &
    \includegraphics[width=\linewidth]{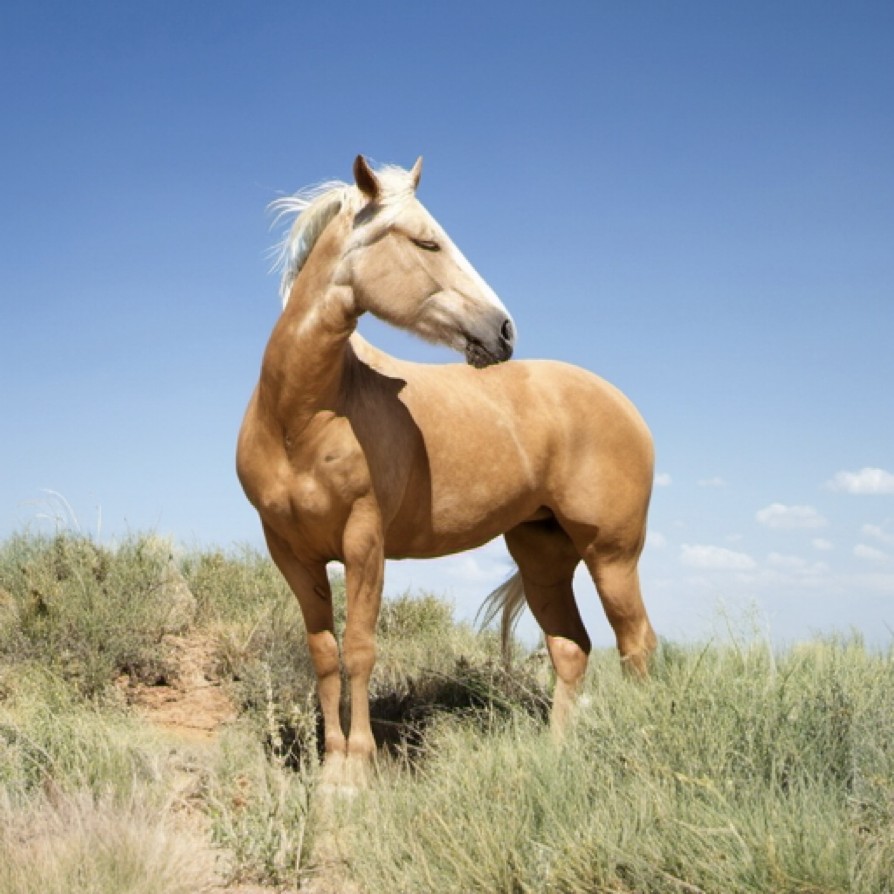} &
    \includegraphics[width=\linewidth]{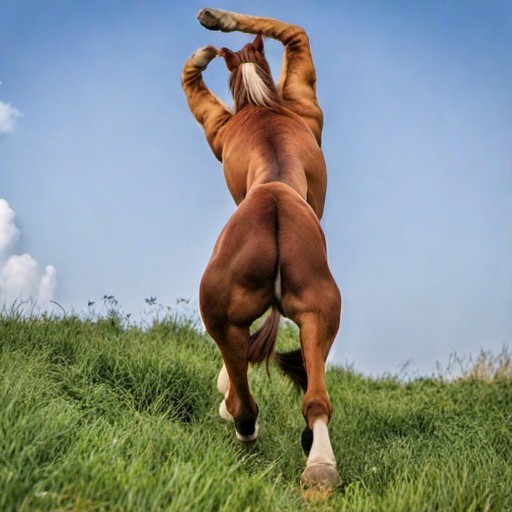} &
    \includegraphics[width=\linewidth]{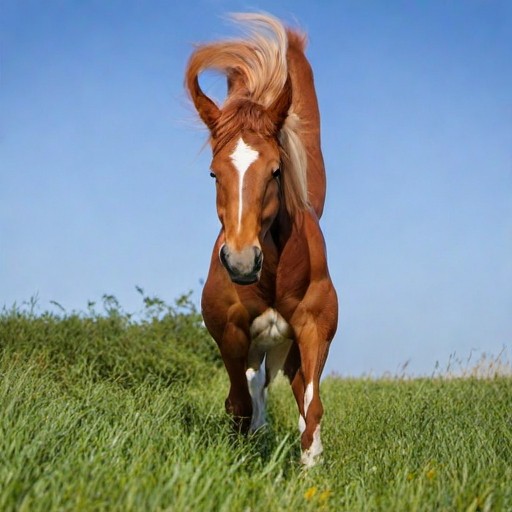} &
    \includegraphics[width=\linewidth]{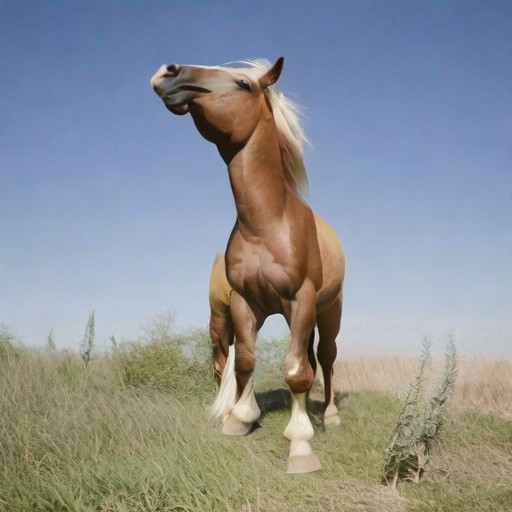} &
    \includegraphics[width=\linewidth]{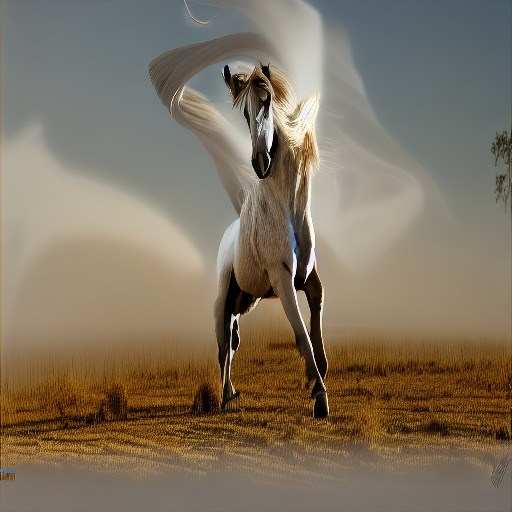} &
    \includegraphics[width=\linewidth]{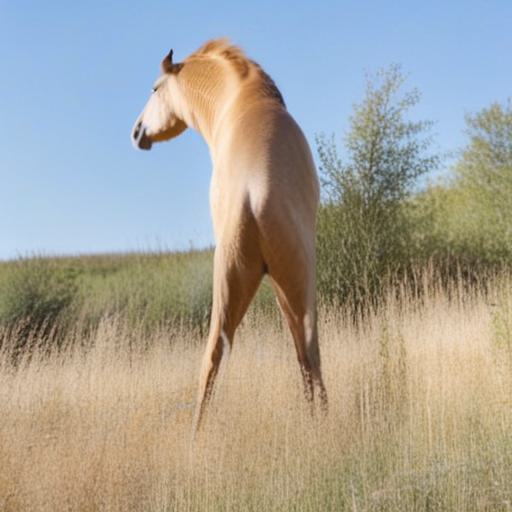} &
    \includegraphics[width=\linewidth]{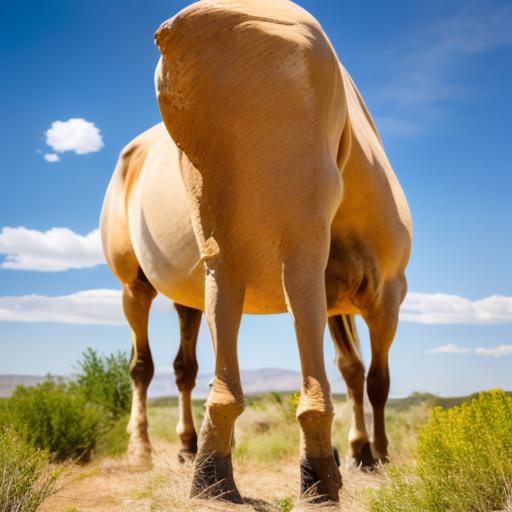}
    \\
    \includegraphics[width=\linewidth]{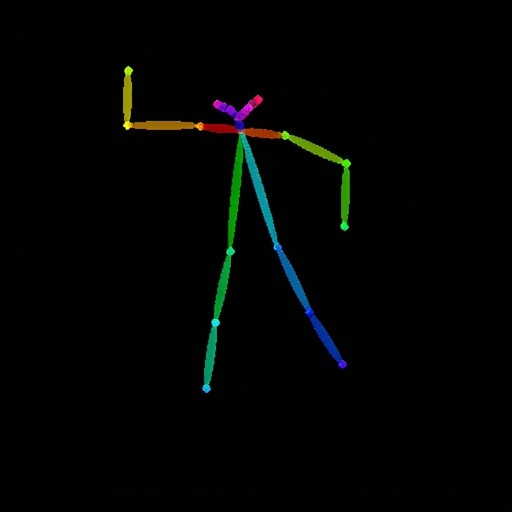} &
    \includegraphics[width=\linewidth]{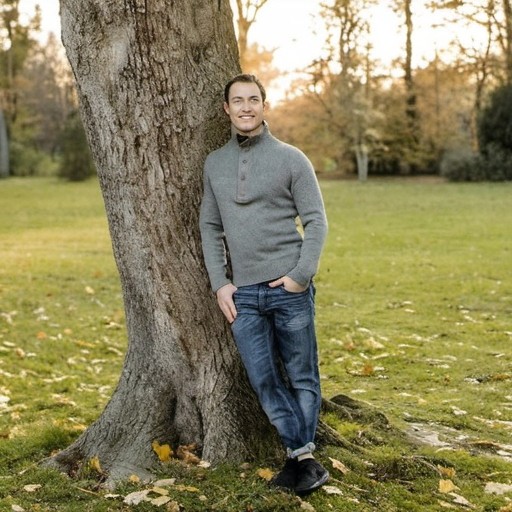} &
    \includegraphics[width=\linewidth]{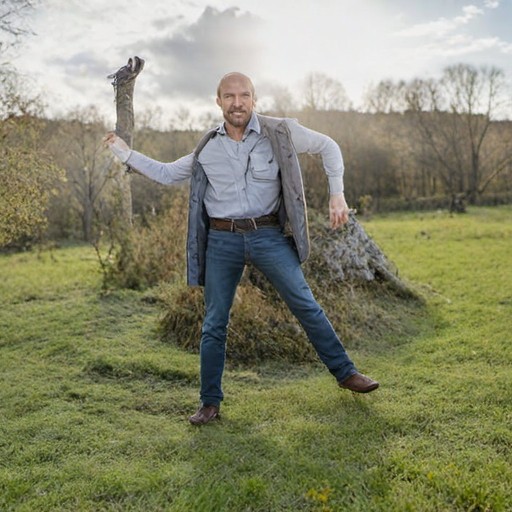} &
    \includegraphics[width=\linewidth]{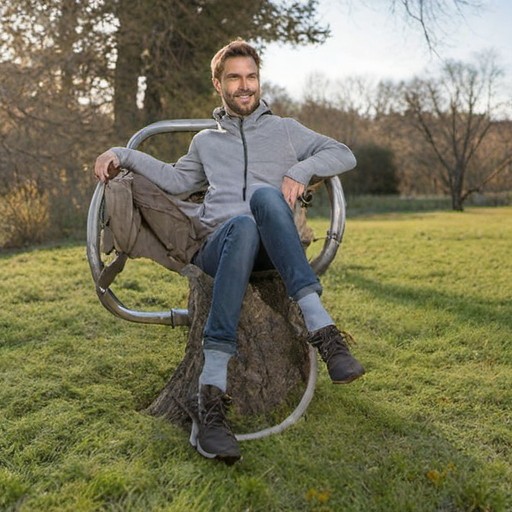} &
    \includegraphics[width=\linewidth]{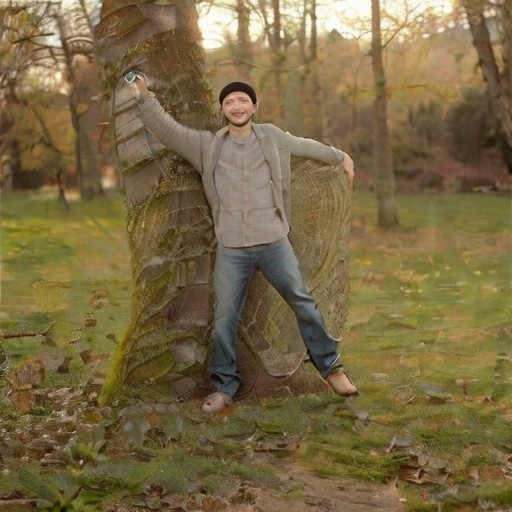} &
    \includegraphics[width=\linewidth]{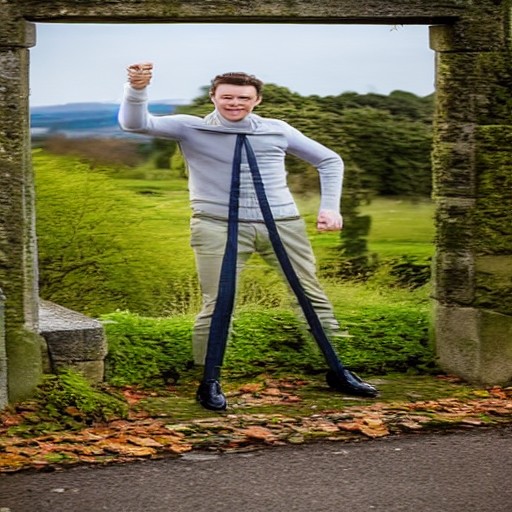} &
    \includegraphics[width=\linewidth]{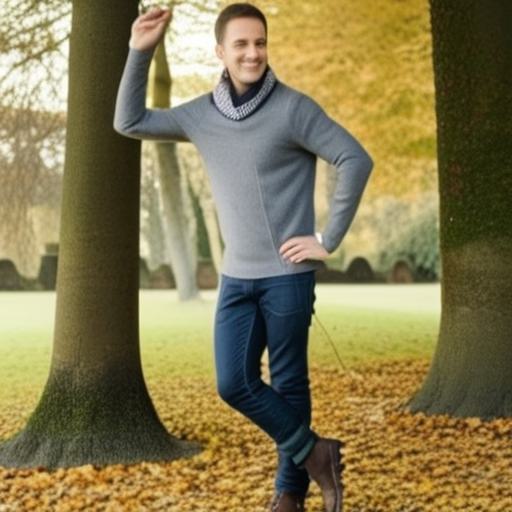} &
    \includegraphics[width=\linewidth]{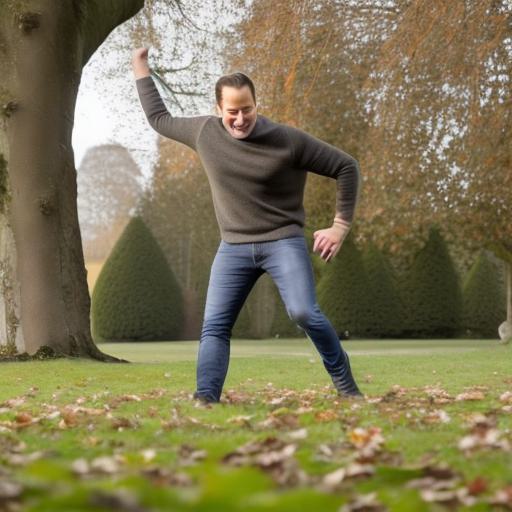}
    \\
    \includegraphics[width=\linewidth]{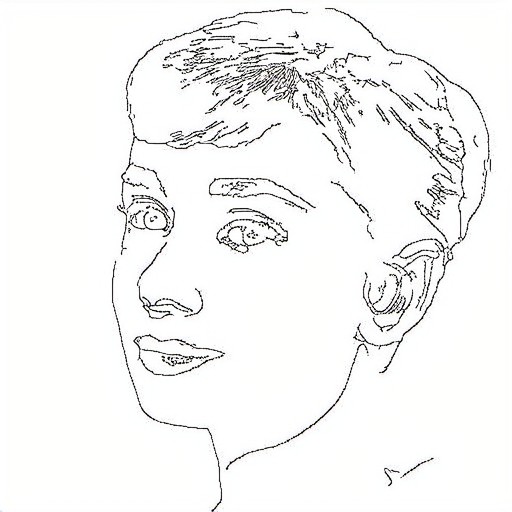} &
    \includegraphics[width=\linewidth]{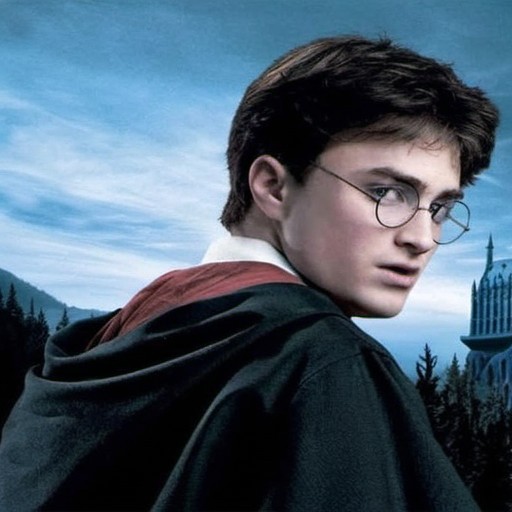} &
    \includegraphics[width=\linewidth]{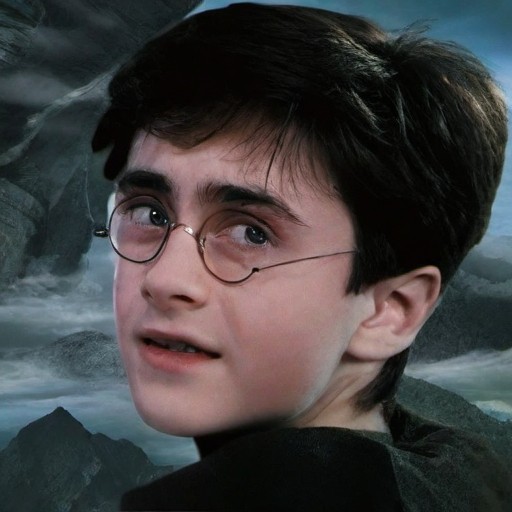} &
    \includegraphics[width=\linewidth]{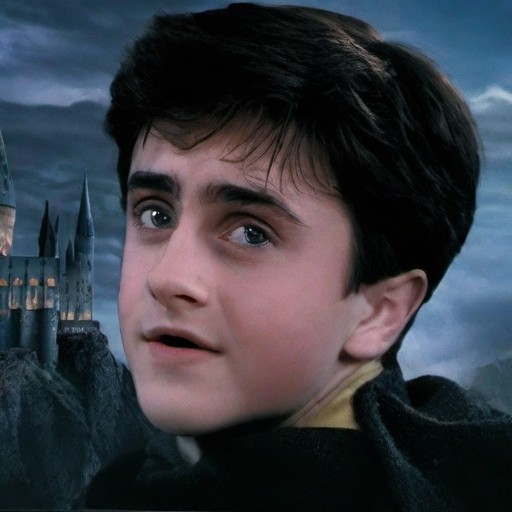} &
    \includegraphics[width=\linewidth]{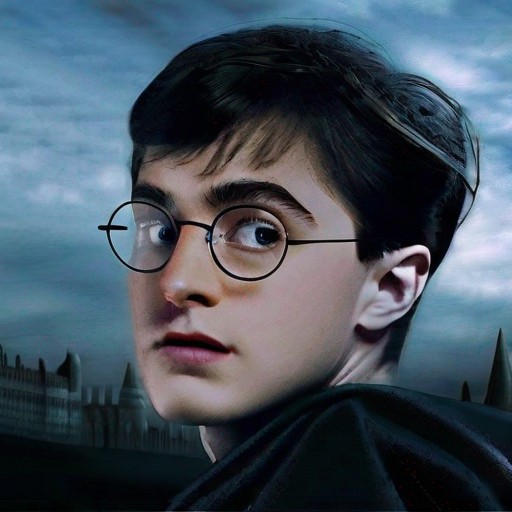} &
    \includegraphics[width=\linewidth]{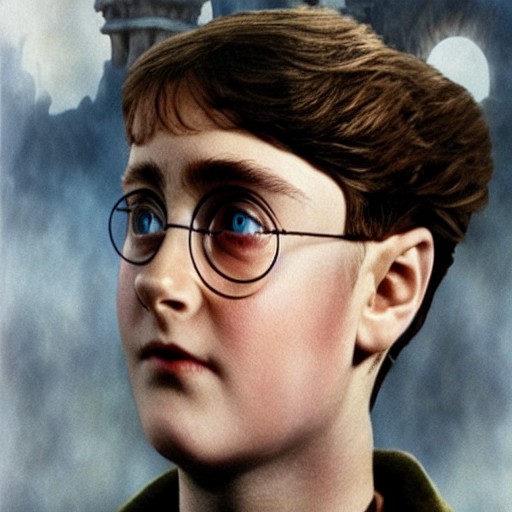} &
    \includegraphics[width=\linewidth]{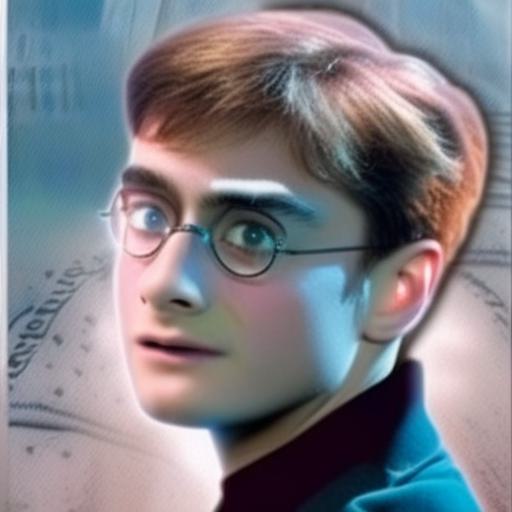} &
    \includegraphics[width=\linewidth]{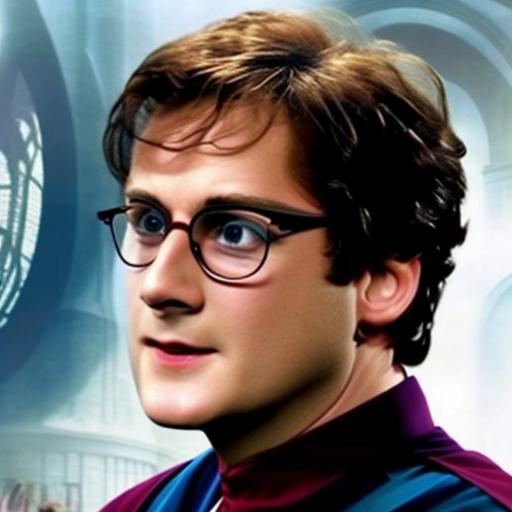} 
    \\
    \includegraphics[width=\linewidth]{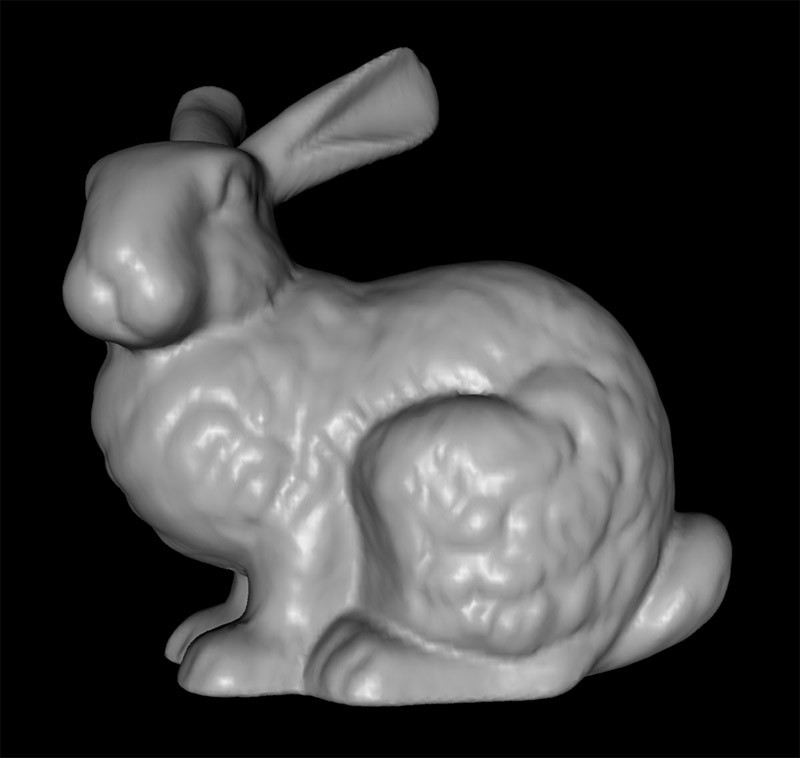} &
    \includegraphics[width=\linewidth]{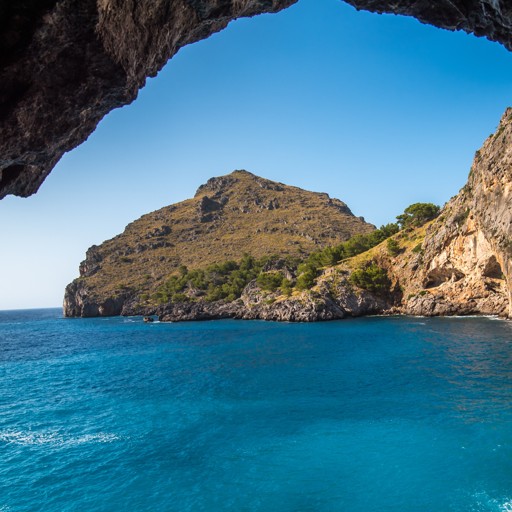} &
    \includegraphics[width=\linewidth]{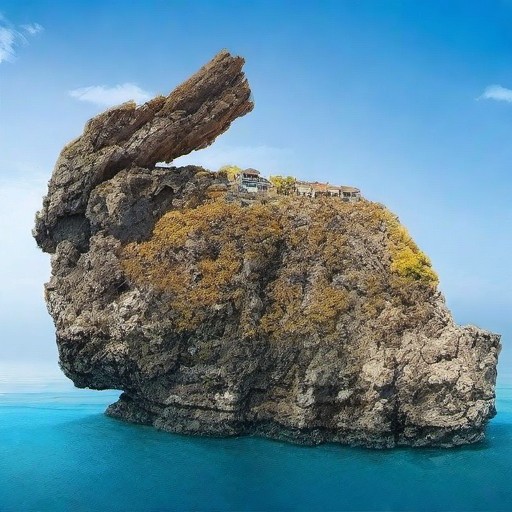} &
    \includegraphics[width=\linewidth]{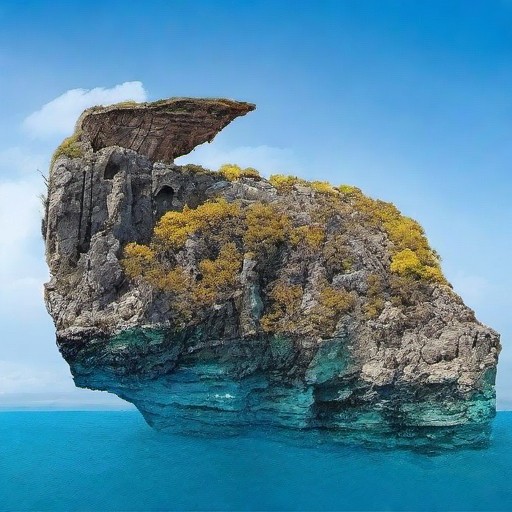} &
    \includegraphics[width=\linewidth]{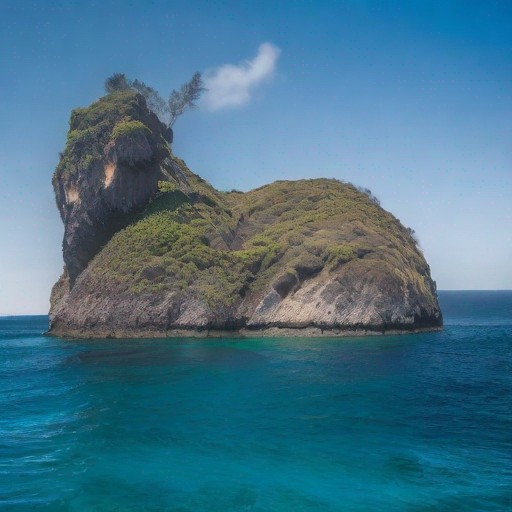} &
    \includegraphics[width=\linewidth]{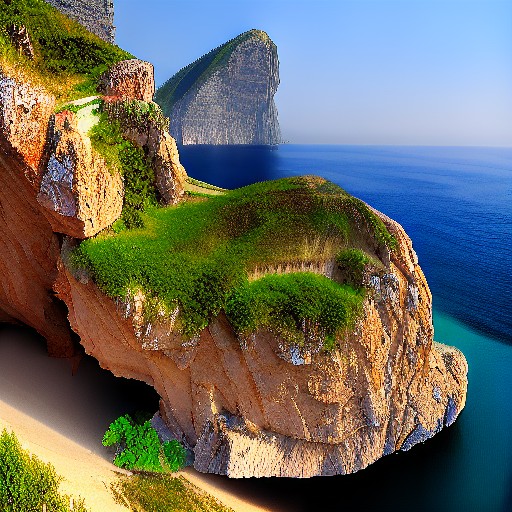} &
    \includegraphics[width=\linewidth]{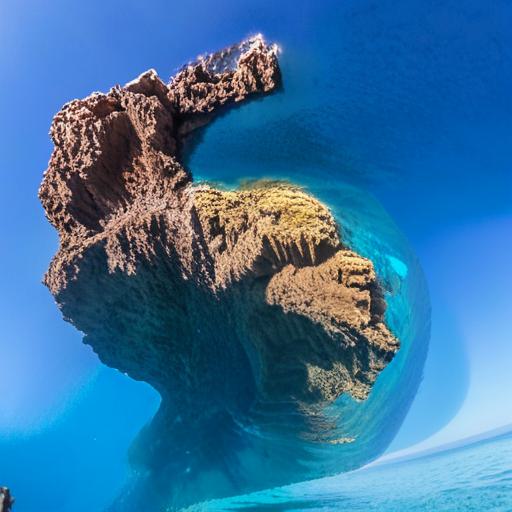} &
    \includegraphics[width=\linewidth]{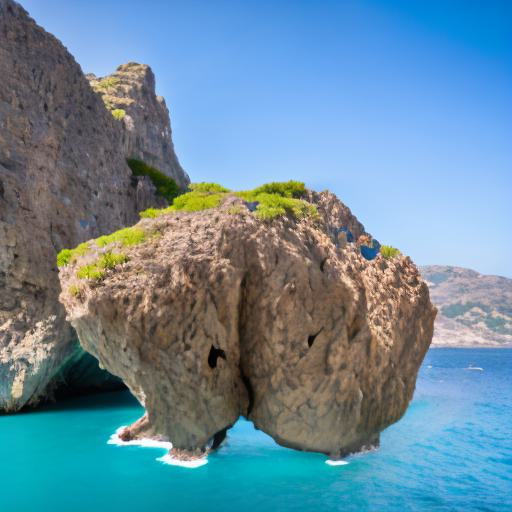}
    \\
    \hdashline
    \noalign{\vspace{4pt}}
    \arrayrulecolor{black}
    \includegraphics[width=\linewidth]{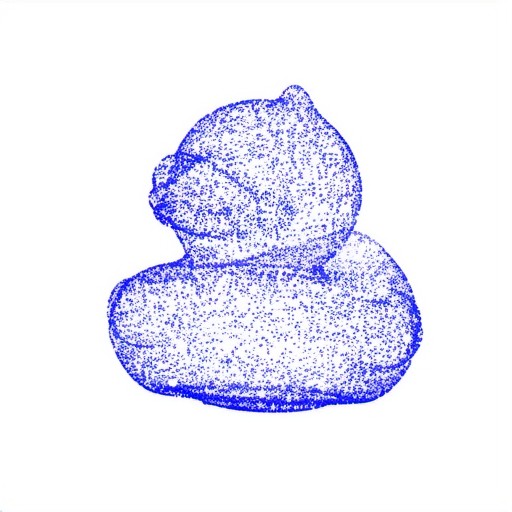} &
    \includegraphics[width=\linewidth]{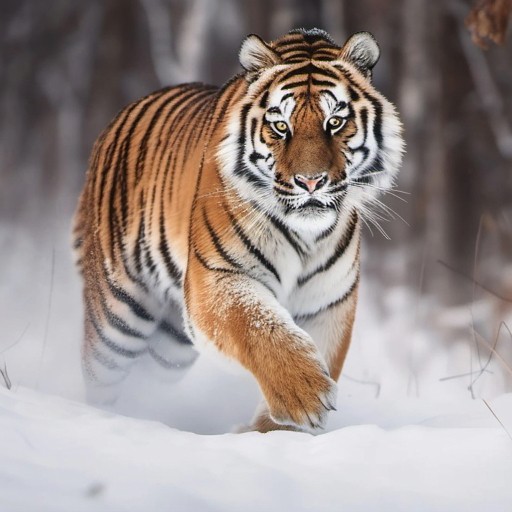} &
    \includegraphics[width=\linewidth]{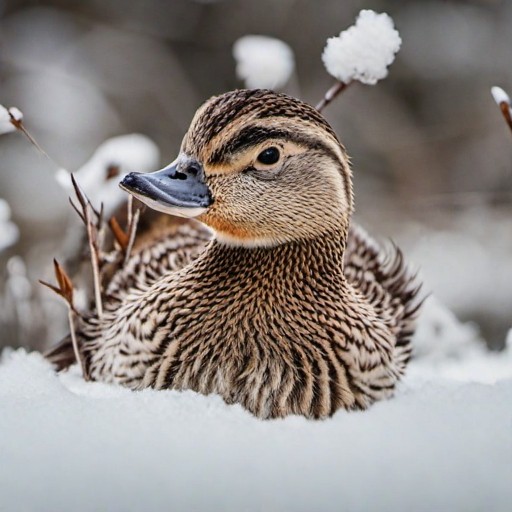} &
    \includegraphics[width=\linewidth]{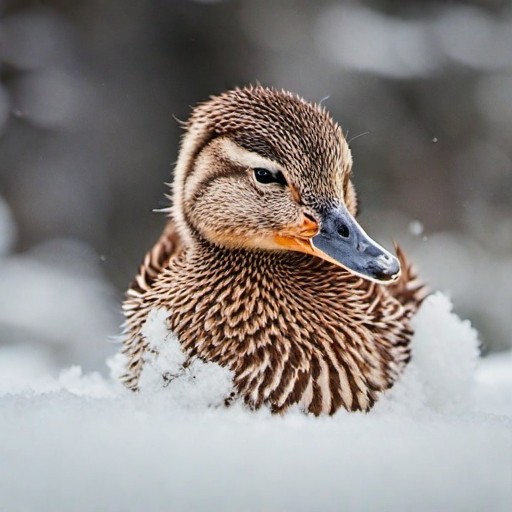} &
    \includegraphics[width=\linewidth]{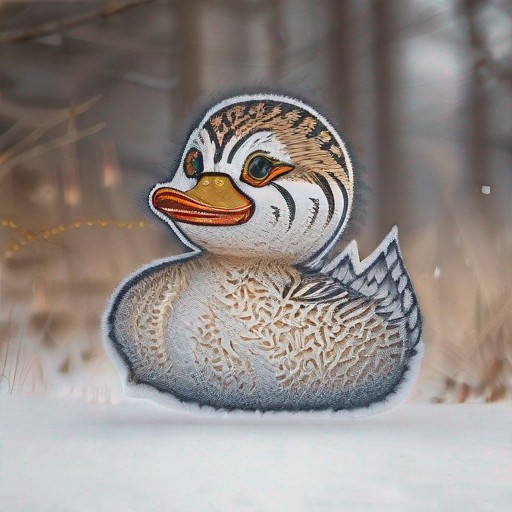} &
    \includegraphics[width=\linewidth]{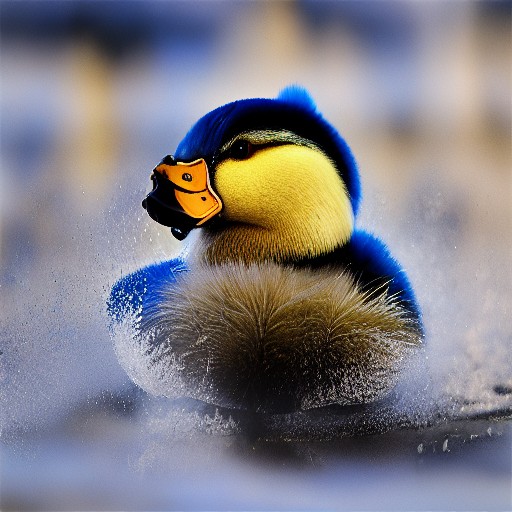} &
    \includegraphics[width=\linewidth]{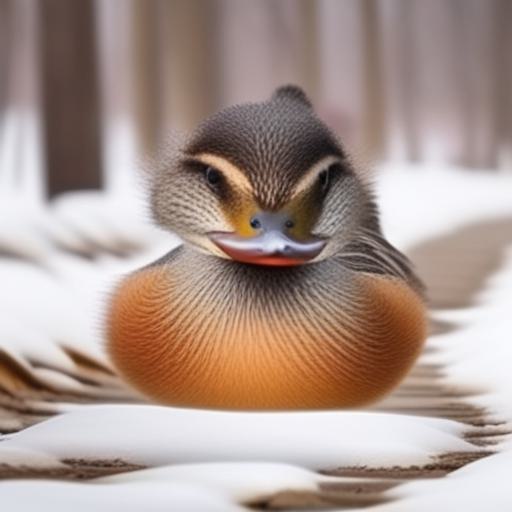} &
    \includegraphics[width=\linewidth]{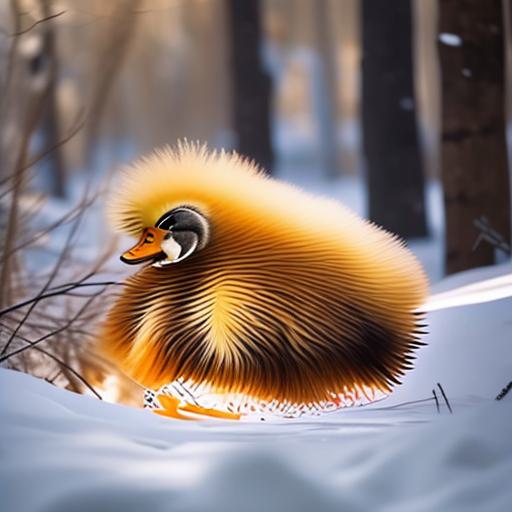}
    \\[-4pt]
    \multicolumn{8}{c}{%
      \small
      \vspace{0pt}
      \setulcolor{Dandelion}\setul{0.3pt}{1.5pt}%
      \textit{``A photo of \ul{a duck} on the snow field"}%
      \vspace{0pt}
    }\\[2pt]
    \includegraphics[width=\linewidth]{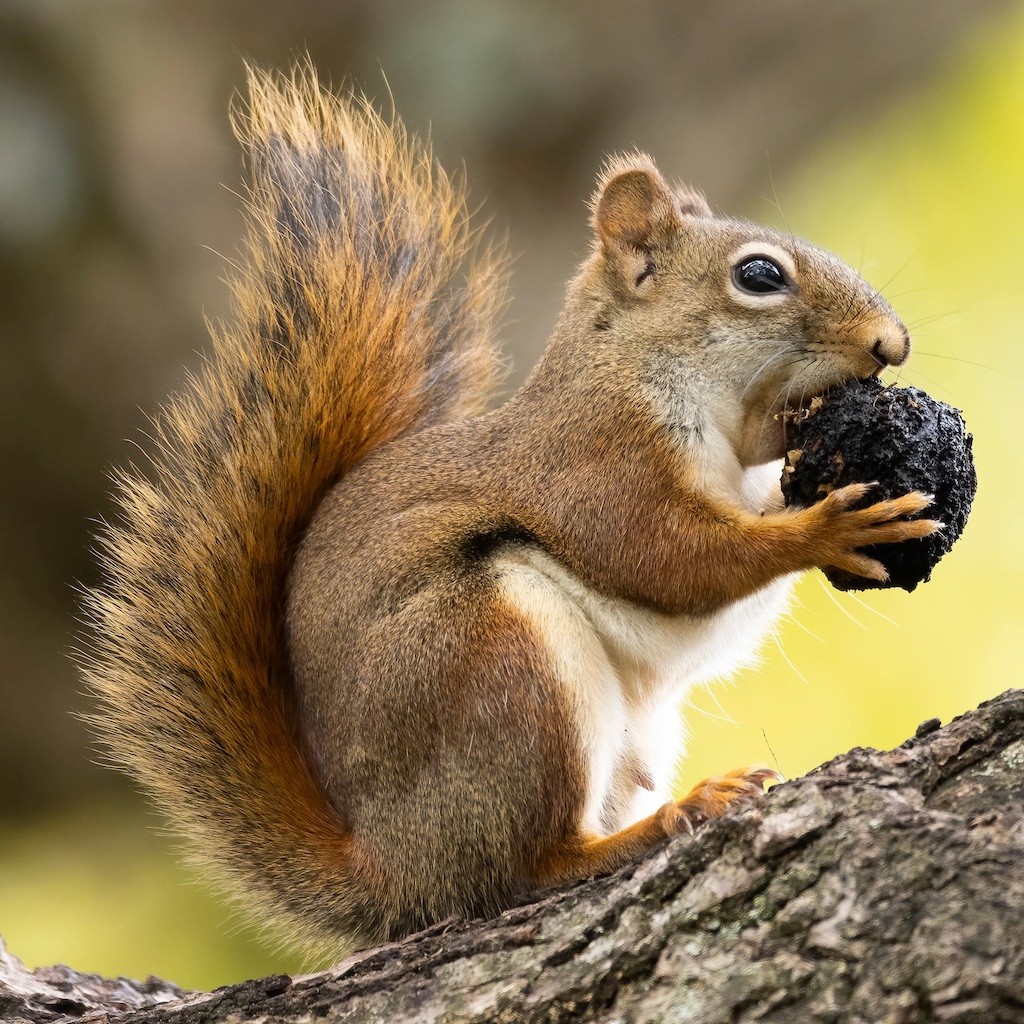} &
    \includegraphics[width=\linewidth]{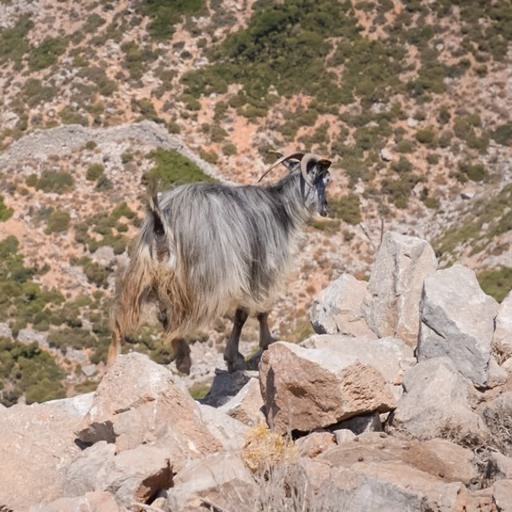} &
    \includegraphics[width=\linewidth]{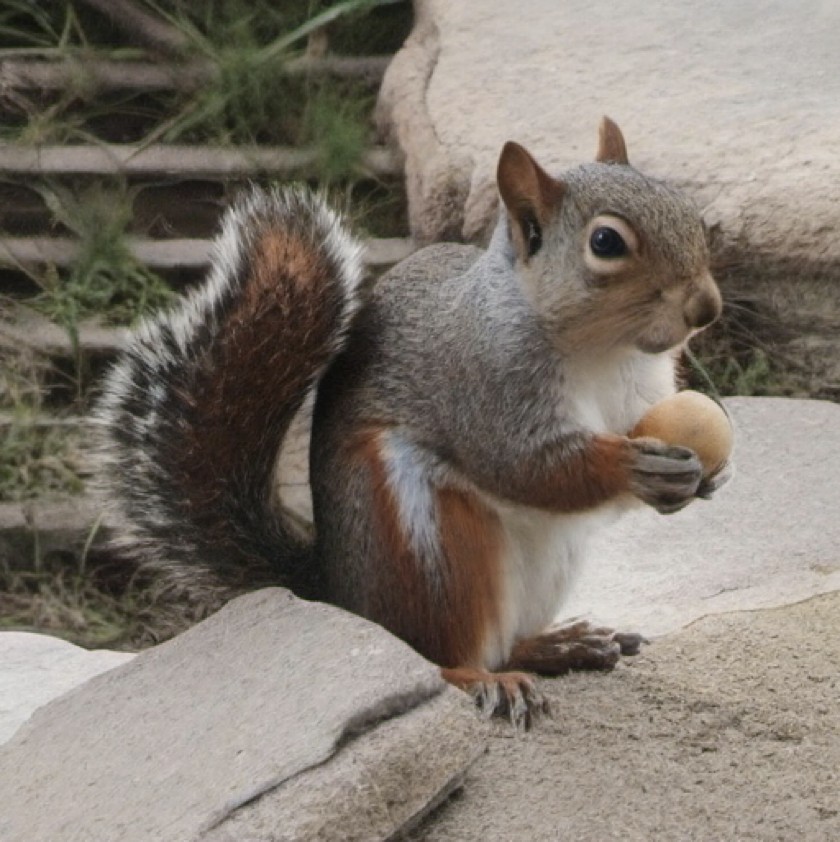} &
    \includegraphics[width=\linewidth]{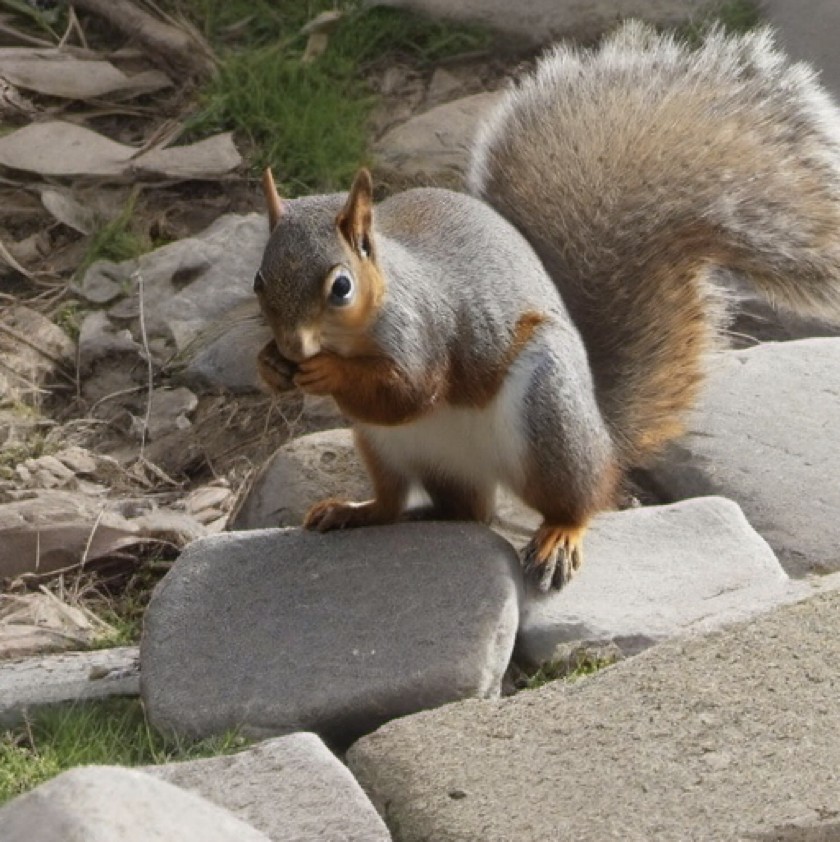} &
    \includegraphics[width=\linewidth]{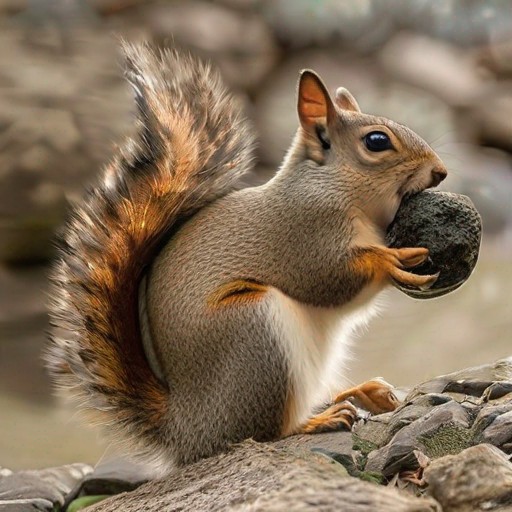} &
    \includegraphics[width=\linewidth]{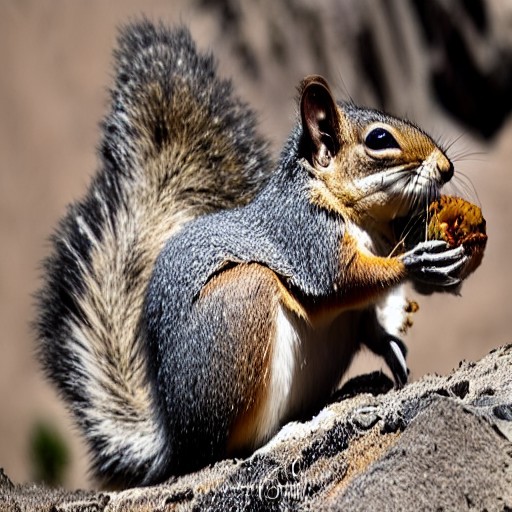} &
    \includegraphics[width=\linewidth]{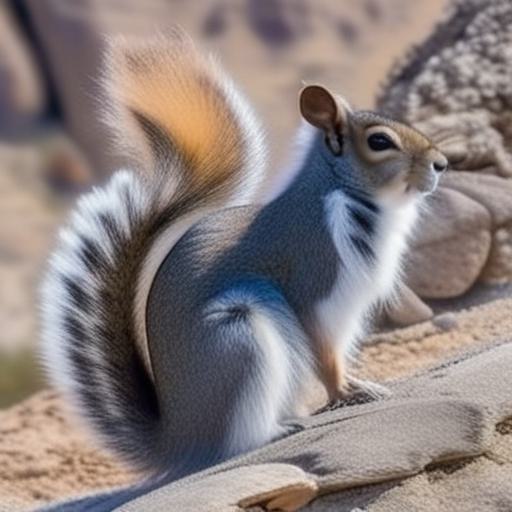} &
    \includegraphics[width=\linewidth]{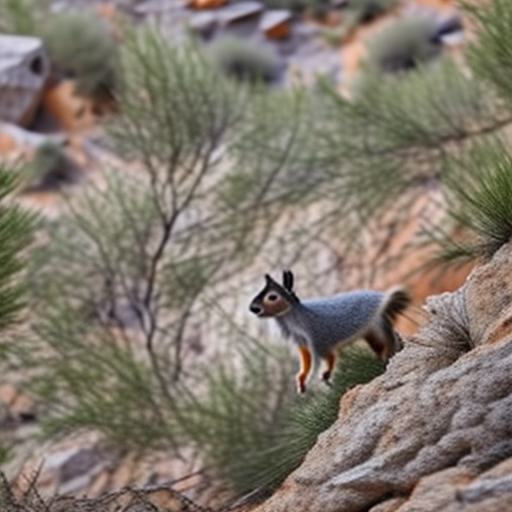}
    \\[-4pt]
    \multicolumn{8}{c}{%
      \small
      \vspace{0pt}
      \setulcolor{Dandelion}\setul{0.3pt}{1.5pt}%
      \textit{``A photo of \ul{a squirrel} walking on stone"}%
      \vspace{0pt}
    }\\[1pt]  
\end{tabular}
} 

\vspace{-10pt}
\caption{\textbf{Qualitative results of appearance and structure control.} DRF successfully generates fused images that preserve the characteristics of the appearance and structural images.}
\vspace{-14pt}
\label{fig:comparison_qual}
\end{figure*}
\vspace{-8pt}

Unlike IDS, which translates the given source to the target by optimizing the latent with a randomly sampled time step $t\sim\mathcal{U}(0, 1)$, our method generates the output from the initial noise $\mathbf{z}_1\sim\mathcal{N}(0, \mathbf{I})$ for timestep $t=1, \dots, 0$. Since the posterior mean $\mathbf{z}_{0|t}^a$ in the large timestep has insufficient information on $\mathbf{z}_0^a$, the difference between $\mathbf{z}_{0|t}^a$ and $\mathbf{z}_0^a$ increases, causing excessive guidance to $\mathbf{z}_0^a$, shown in last column of \cref{fig:diffusion_process_qual}. We avoid the excessive difference by modifying \cref{eq:forward} to compute modified stochastic latent $\tilde{\mathbf{z}}_t^a$ as following:
\begingroup
\setlength{\abovedisplayskip}{5pt}   
\setlength{\belowdisplayskip}{5pt}   
\begin{equation}
\scalebox{0.98}{$
\tilde{\mathbf{z}}_t^a = 
\sqrt{\frac{\alpha_t}{\alpha_{t-1}}}\mathbf{z}_0^a + 
\sqrt{1-\frac{\alpha_t}{\alpha_{t-1}}}\epsilon.
$}
\label{eq:sim_app_latent}
\end{equation}
\endgroup
By designing a loss that minimizes the difference between $\mathbf{z}^a$ and $\mathbf{z}_{0|t}^a=\mathbb{E}[\mathbf{z}_0|\tilde{\mathbf{z}}_t^a]$ calculated from the score $\epsilon_\theta^a=\epsilon_\theta^\omega(\tilde{\mathbf{z}}_t^a, y^a, t)$, we ensure that the appearance information remains intact:
\begingroup
\setlength{\abovedisplayskip}{4pt}   
\setlength{\belowdisplayskip}{4pt}   
\begin{equation}
\mathcal{L}_{\text{app}} = d ( \mathbf{z}_{0|t}^a, \mathbf{z}_{0}^a),
\label{eq:loss_app}
\end{equation}
\endgroup
Even though the overemphasis on the appearance image is mitigated by the modified stochastic latent, considering only appearance feedback may hinder its seamless integration in $\mathbf{I}^g$, as demonstrated in the fourth column of \cref{fig:app_feedback}.
\\
\noindent\textbf{Generation Feedback.}\quad
To further address appearance overfitting, we incorporate generation feedback to $\mathbf{z}^g$. This ensures a more balanced fusion of appearance signals and ultimately improves the generation quality, as shown in the third column of \cref{fig:app_feedback}.
Specifically, as the feedback is iteratively applied in a recursive manner, the score $\epsilon_\theta^{g}$ updated by the previous gradient modifies $\mathbf{z}_{\text{prev}}^g$. 
We treat $\mathbf{z}_{\text{prev}}^g$ as another fixed point so that the current update direction can be aligned with $\mathbf{z}_{\text{prev}}^g$, allowing for more efficient image synthesis. A loss for the output feedback is designed to minimize the difference between $\mathbf{z}_{\text{prev}}^g$ and $\mathbf{z}_{0|t}^g$ as follows:
\begingroup
\setlength{\abovedisplayskip}{4pt}   
\setlength{\belowdisplayskip}{4pt}   
\begin{equation}
\mathcal{L}_{\text{gen}} = d ( \mathbf{z}_{0|t}^g, \mathbf{z}_\text{prev}^g).
\label{eq:loss_out}
\end{equation}
\endgroup

\noindent\textbf{Dual Recursive Feedback.}\quad
We combine the appearance feedback loss \cref{eq:loss_app} and the generation feedback loss \cref{eq:loss_out} to formulate our final \textbf{Dual Recursive Feedback (DRF)}. 
Using the structural information from the initial steps contains sufficient information \cite{lin2025ctrl}; however, due to the nature of initializing the initial latent $\mathbf{z}_T^g$ with random Gaussian noise, the structure of the image generated in the first inference step may not align with $\mathbf{I}^s$ \cite{tumanyan2023plug}.  
Therefore, we applied the DRF to the intermediate 20 steps after the first five steps, considering both the efficiency and generation quality. The discussion and experimental evaluation of the DRF application steps are detailed in the supplementary.
As we progress through multiple iterations of DRF at each step, 
$\mathbf{z}_t^g$ is successively refined, incorporating more corrective feedback and thus becoming increasingly aligned with both the appearance and structure constraints. 

To balance appearance and generation feedback, we adopt an exponential weighting scheme that progressively amplifies the generation-feedback weight as the recursive iterations ($i$) advance: 
\begingroup
\setlength{\abovedisplayskip}{4pt}   
\setlength{\belowdisplayskip}{5pt}   
\begin{equation}
\resizebox{0.4\hsize}{!}{$
w_{\text{iter}}^{(i)} 
         \;=\; 
         \sqrt{
           \frac{\exp\!\Bigl(k\,\tfrac{i}{N-1}\Bigr) \;-\;1}{
                 \exp(k)\;-\;1}
         }.
$}
\label{eq:weight}
\end{equation}
\endgroup

\noindent An empirical comparison of alternative weighting strategies is presented in the supplementary.

\noindent Namely, appearance feedback is more crucial at the beginning so that the identity of the appearance can be reflected properly in the output, even in complex cases. As the number of iterations increases, the generation feedback is taken more into account to achieve the output that better matches the user's intent. To this end, the \textit{DRF loss} is defined as follows:

\begingroup
\setlength{\abovedisplayskip}{4pt}   
\setlength{\belowdisplayskip}{4pt}   
\begin{equation}
\mathcal{L}_{\text{DRF}}^{(i)} 
         \;\leftarrow\;
         d\bigl(\mathbf{z}_{0\mid t}^{a},\,\mathbf{z}_0^{a}\bigr)
         \;+\;
         \rho\,w_{\text{iter}}^{(i)}\,
         d\bigl(\mathbf{z}_{0\mid t}^{g},\,\mathbf{z}_{\text{prev}}^{g}\bigr),
\label{eq:loss_DRF}
\end{equation}
\endgroup
where $\rho$ is a hyper-parameter. 

Note that the injection noise $\epsilon$ is added to $\mathbf{z}_{t-1}^g$ and $\mathbf{z}^a$ to obtain the stochastic latent $\mathbf{z}_t^g$ and $\tilde{\mathbf{z}}_t^a$, which can be considered as the connection between $\mathbf{z}_t^g$ and $\tilde{\mathbf{z}}_t^a$. Thus, we updated the injection noise 
$\epsilon$ in the direction that minimizes the \textit{DRF loss} over multiple recursive steps:
\begingroup
\setlength{\abovedisplayskip}{4pt}   
\setlength{\belowdisplayskip}{4pt}   
\begin{equation}
\epsilon \;\leftarrow\; 
        \epsilon 
        \;-\;        \lambda\,\nabla_{\epsilon}\,\mathcal{L}_{\text{DRF}}^{(i)}.
        \label{eq:update}
\end{equation}
\endgroup
With the modified noise $\epsilon$ in \cref{eq:update}, the fully refined $\mathbf{z}^{g*}$ can be provided. Extensive sensitivity analyses for the hyper-parameters are reported in the supplementary.
Consequently, this final $\mathbf{z}^{g*}$ retains the essential features of the appearance image while being appropriately corrected at each step to preserve the structure as well, thereby generating a high-quality image that honors both sets of constraints. 

\newcommand{\bfz}{\mathbf{z}}
\newcommand{\hbfz}{\hat{\mathbf{z}}}

\begin{algorithm}[!t]
\setstretch{1.2}
\footnotesize               
\caption{Dual Recursive Feedback}
\label{alg:DRF}
\begin{algorithmic}[1]

\REQUIRE $\mathbf{z}_0^{a}$, $\mathbf{z}_{t-1}^{g}$, $y^{a}$: Prompts of appearance, $y^{g}$: Prompts of generation, $\omega$, $\lambda$, $\rho$, $k$, $N$ : balancing parameters \& constants

    \STATE $\epsilon \sim \mathcal{N}(0, \mathbf{I})$
    \FOR{$i = 1, \dots, N$}
        \STATE $\mathbf{z}_{t}^{g} 
                \leftarrow 
                \sqrt{\frac{\alpha_t}{\alpha_{t-1}}}
                \,\mathbf{z}_{t-1}^{g}
                \;+\;
                \sqrt{1-\frac{\alpha_t}{\alpha_{t-1}}}\,
                \epsilon$
        \STATE $\tilde{\mathbf{z}}_{t}^{a} 
                \leftarrow 
                \sqrt{\frac{\alpha_t}{\alpha_{t-1}}}
                \,\mathbf{z}_0^{a}
                \;+\;
                \sqrt{1-\frac{\alpha_t}{\alpha_{t-1}}}\,
                \epsilon$
        \STATE 
        $\epsilon_\theta^a\;,\; \epsilon_\theta^g
          \leftarrow \epsilon_\theta^\omega\bigl(\tilde{\mathbf{z}}_{t}^a, 
         \,y^{a}, t\bigr)
         \;,\;  \,\epsilon_\theta^\omega\bigl(\mathbf{z}_{t}^g, 
         \,y^{g}, t\bigr)$


        \STATE
        $\mathbf{z}_{0\mid t}^{a}\;,\;\mathbf{z}_{0\mid t}^{g}
         \leftarrow
         \mathbb{E}[\mathbf{z}_0\mid\tilde{\mathbf{z}}_t^a]\;,\;\mathbb{E}[\mathbf{z}_0\mid\mathbf{z}_t^g]$

        \STATE
        $w_{\text{iter}}^{(i)} 
         \;=\; 
         \sqrt{
           \frac{\exp\!\Bigl(k\,\tfrac{i}{N-1}\Bigr) \;-\;1}{
                 \exp(k)\;-\;1}
         }$

        \STATE
        $\mathcal{L}_{\text{DRF}}^{(i)} 
         \;\leftarrow\;
         d\bigl(\mathbf{z}_{0\mid t}^{a},\,\mathbf{z}_0^{a}\bigr)
         \;+\;
         \rho\,w_{\text{iter}}^{(i)}\,
         d\bigl(\mathbf{z}_{0\mid t}^{g},\,\mathbf{z}_{\text{prev}}^{g}\bigr)$

        \STATE 
        $\epsilon \;\leftarrow\; 
        \epsilon 
        \;-\;
        \lambda\,\nabla_{\epsilon}\,\mathcal{L}_{\text{DRF}}^{(i)}$

        \STATE 
        $\mathbf{z}_{\text{prev}}^{g} 
         \;\leftarrow\; 
         \mathbf{z}_{0\mid t}^{g}$
    \ENDFOR
\RETURN $\mathbf{z}_{t}^{g*}$
\end{algorithmic}
\end{algorithm}
\section{Experiments}
\vspace{-2pt}
\label{sec:experiments}
\begin{figure*}[t]
\centering
\resizebox{\textwidth}{!}{%
\setlength{\tabcolsep}{1.2pt}
\begin{tabular}{>{\centering\arraybackslash}m{0.125\textwidth}
                >{\centering\arraybackslash}m{0.125\textwidth}
                >{\centering\arraybackslash}m{0.125\textwidth}
                >{\centering\arraybackslash}m{0.125\textwidth}
                >{\centering\arraybackslash}m{0.125\textwidth}
                >{\centering\arraybackslash}m{0.125\textwidth}
                >{\centering\arraybackslash}m{0.125\textwidth}
                >{\centering\arraybackslash}m{0.125\textwidth}
                >{\centering\arraybackslash}m{0.125\textwidth}}
    \footnotesize{Structure} & \footnotesize{Appearance} & \footnotesize \textbf{DRF (Ours)} & \footnotesize{Ctrl-X\cite{lin2025ctrl}} & \footnotesize{FreeControl\cite{mo2024freecontrol}} & \footnotesize{Uni-ControlNet\cite{qin2023unicontrol}} & \footnotesize{ControlNet\cite{zhang2023adding} + IP-Adapter\cite{ye2023ip}} & \footnotesize{T2I-Adapter\cite{mou2024t2i} + IP-Adapter\cite{ye2023ip}} \\
    \includegraphics[width=\linewidth]{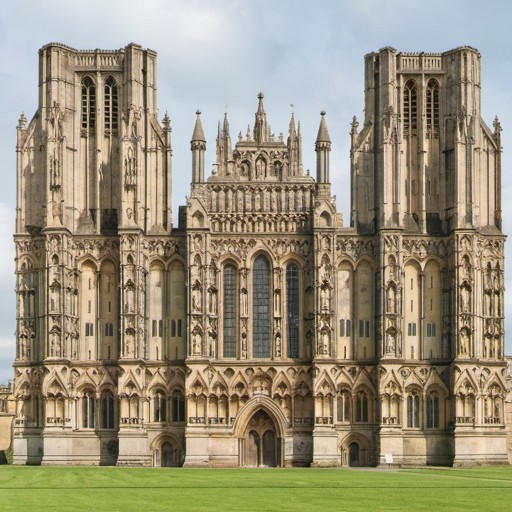} &
    \includegraphics[width=\linewidth]{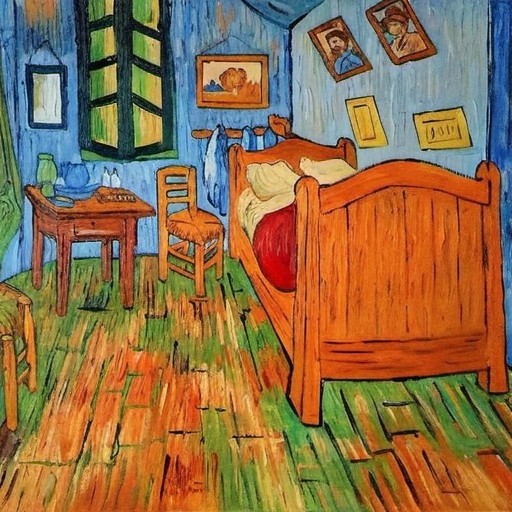} &
    \includegraphics[width=\linewidth]{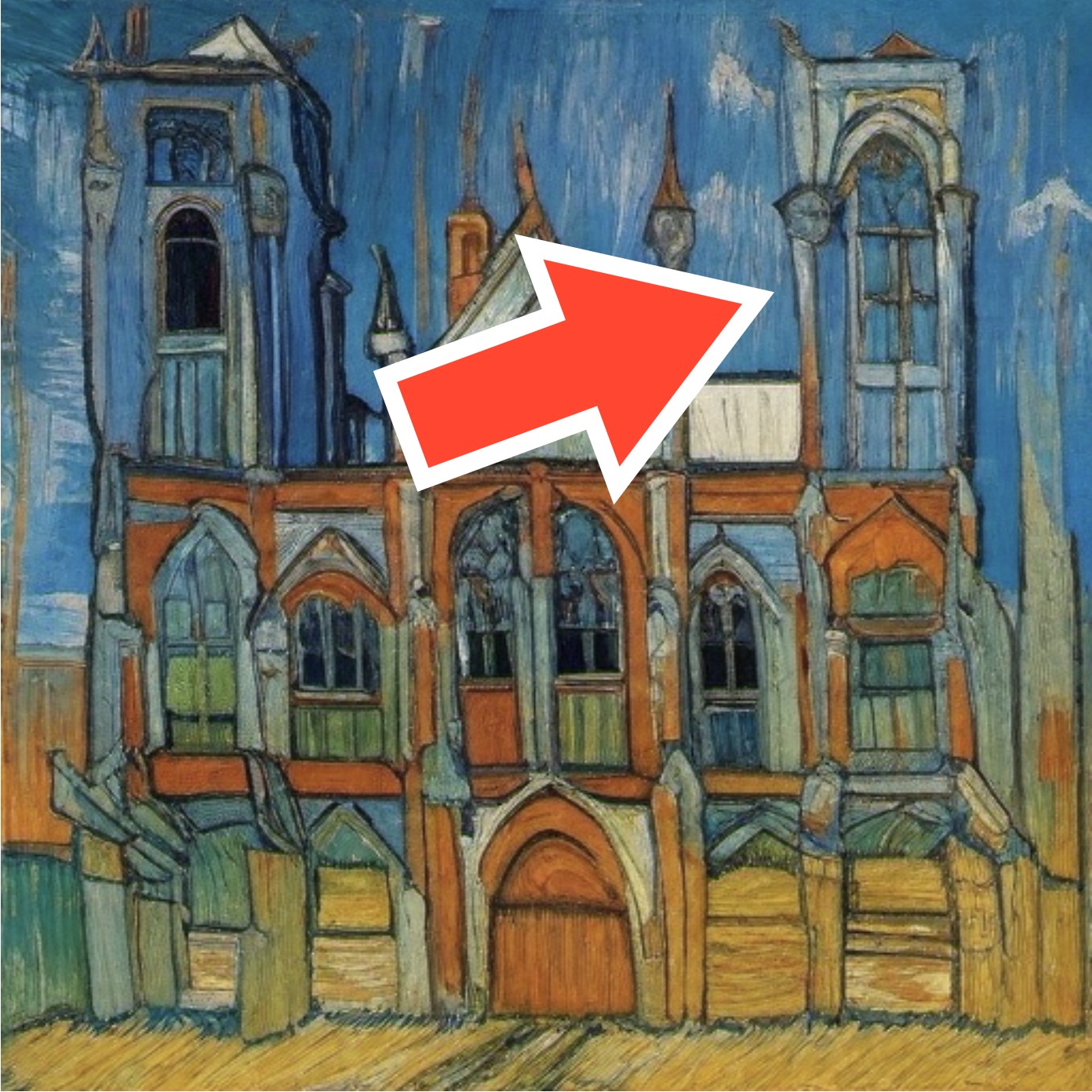} &
    \includegraphics[width=\linewidth]{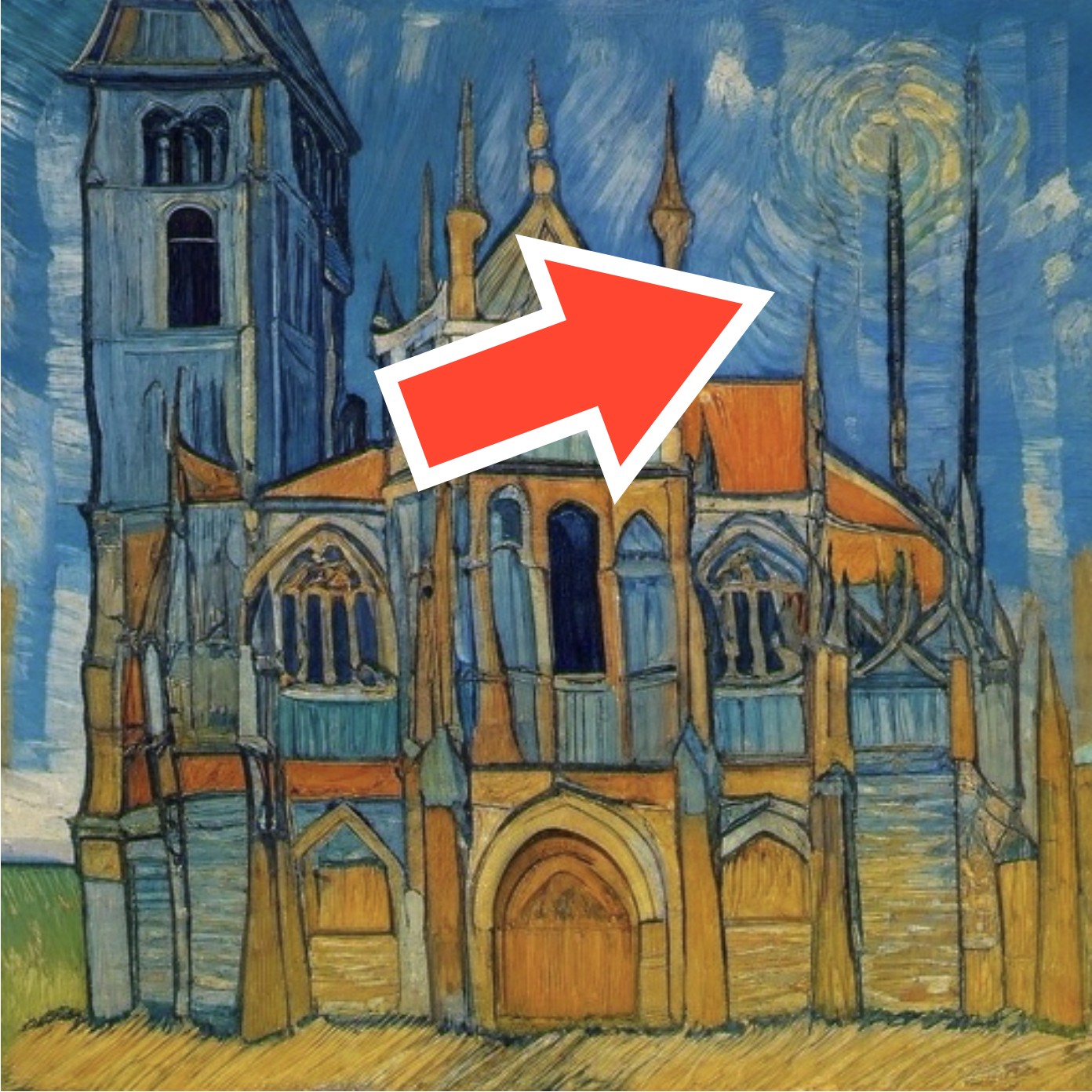} &
    \includegraphics[width=\linewidth]{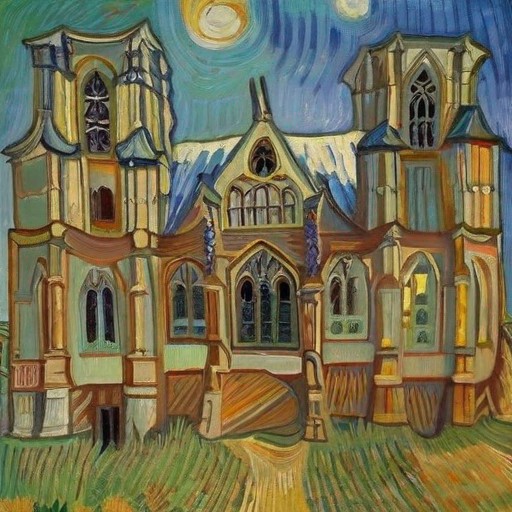} &
    \includegraphics[width=\linewidth]{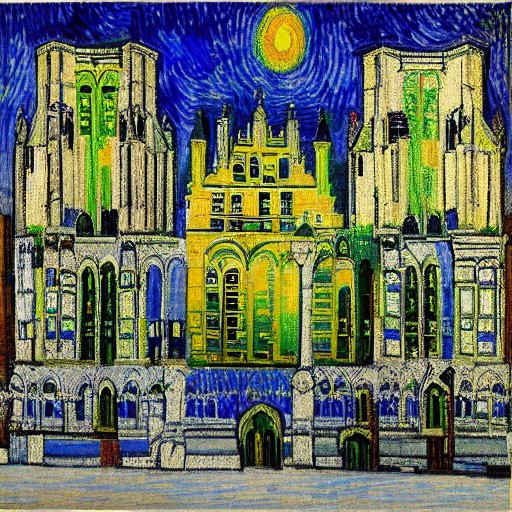} &
    \includegraphics[width=\linewidth]{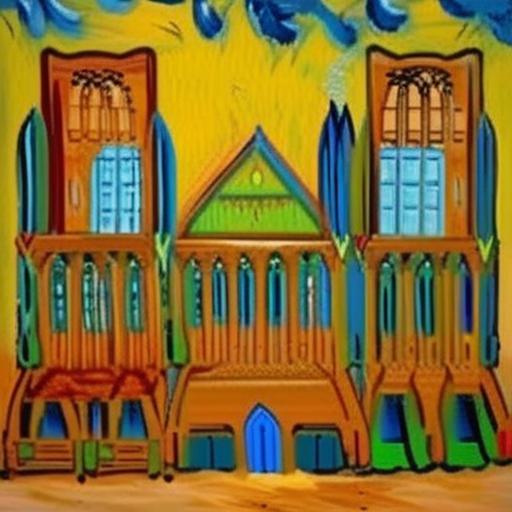} &
    \includegraphics[width=\linewidth]{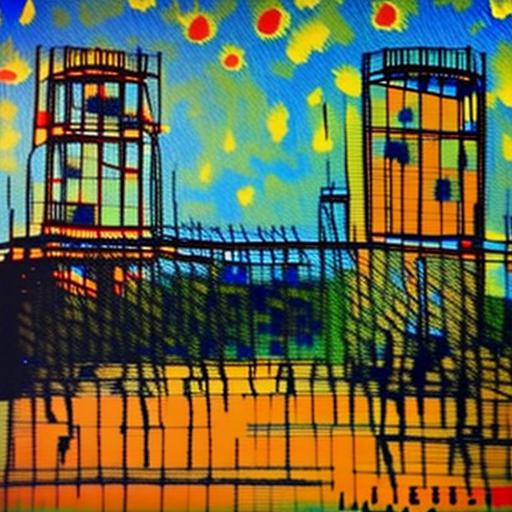}\\
    \includegraphics[width=\linewidth]{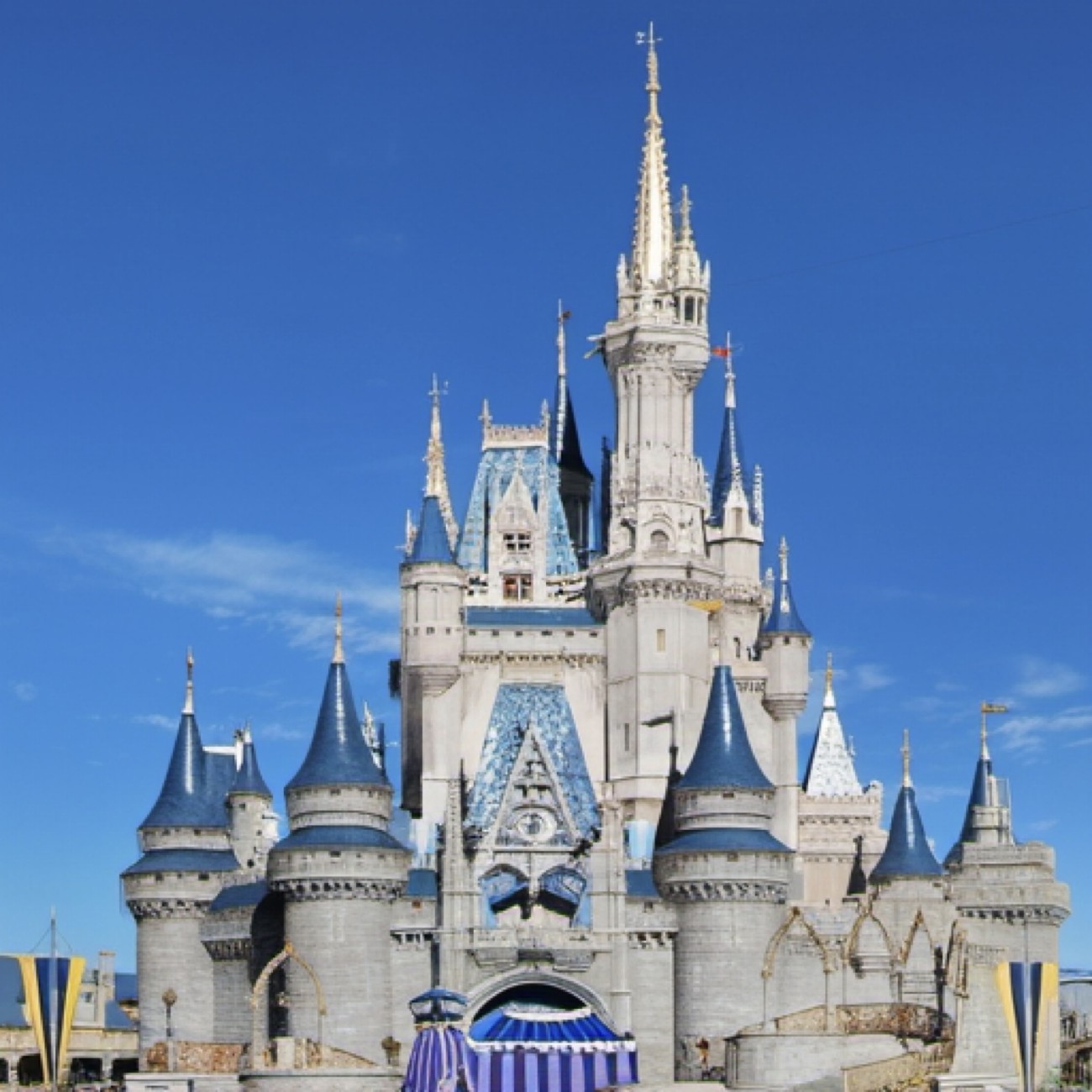} &
    \includegraphics[width=\linewidth]{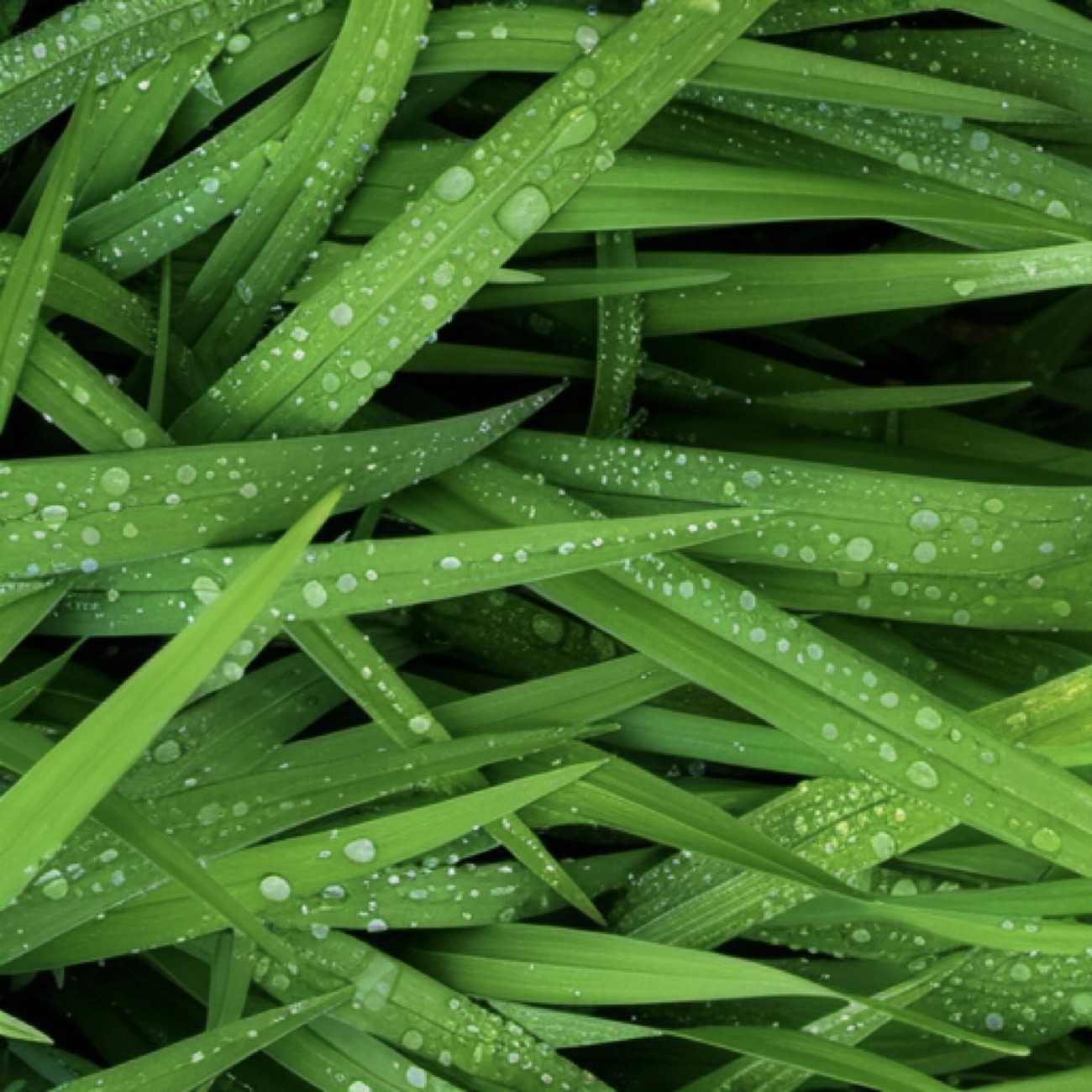} &
    \includegraphics[width=\linewidth]{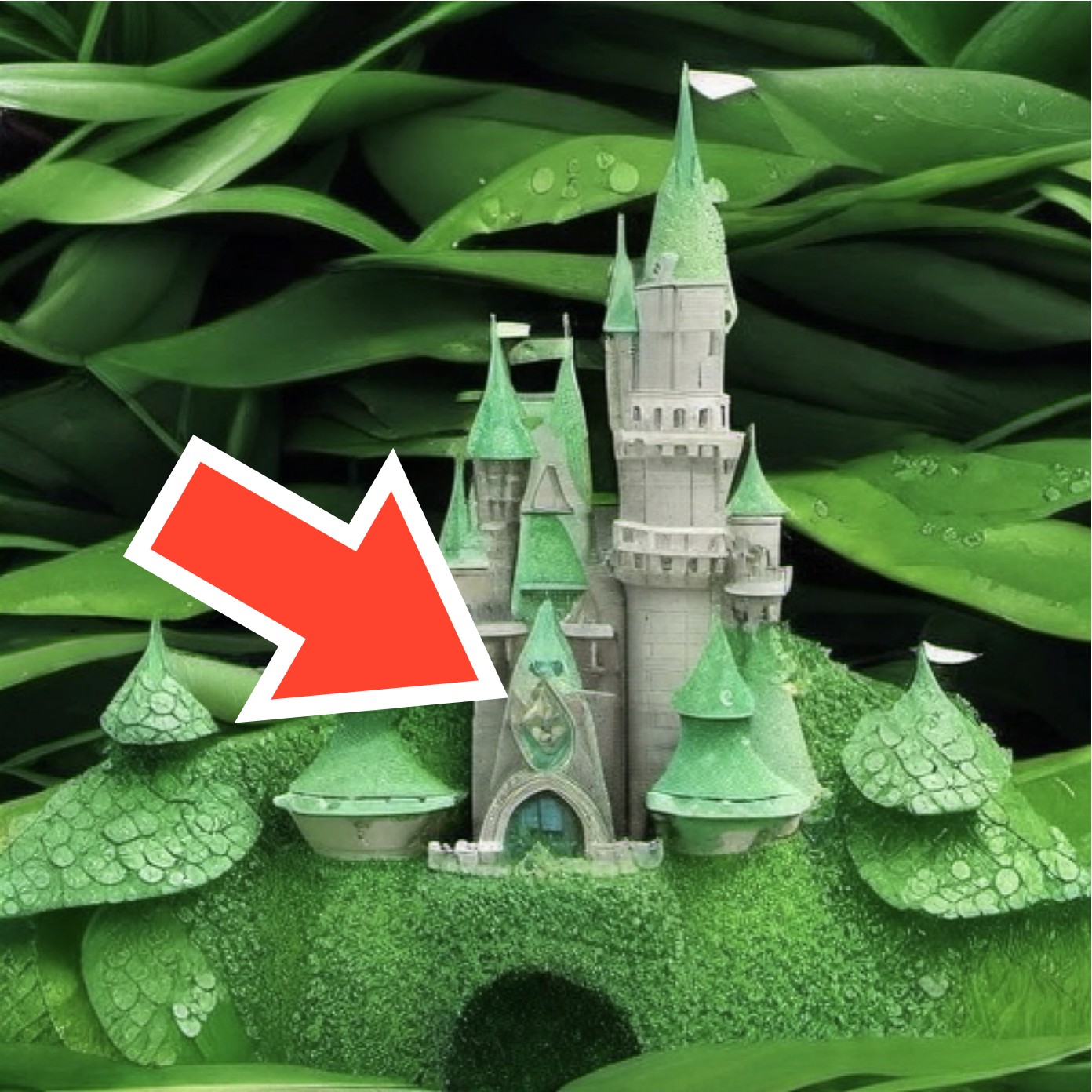} &
    \includegraphics[width=\linewidth]{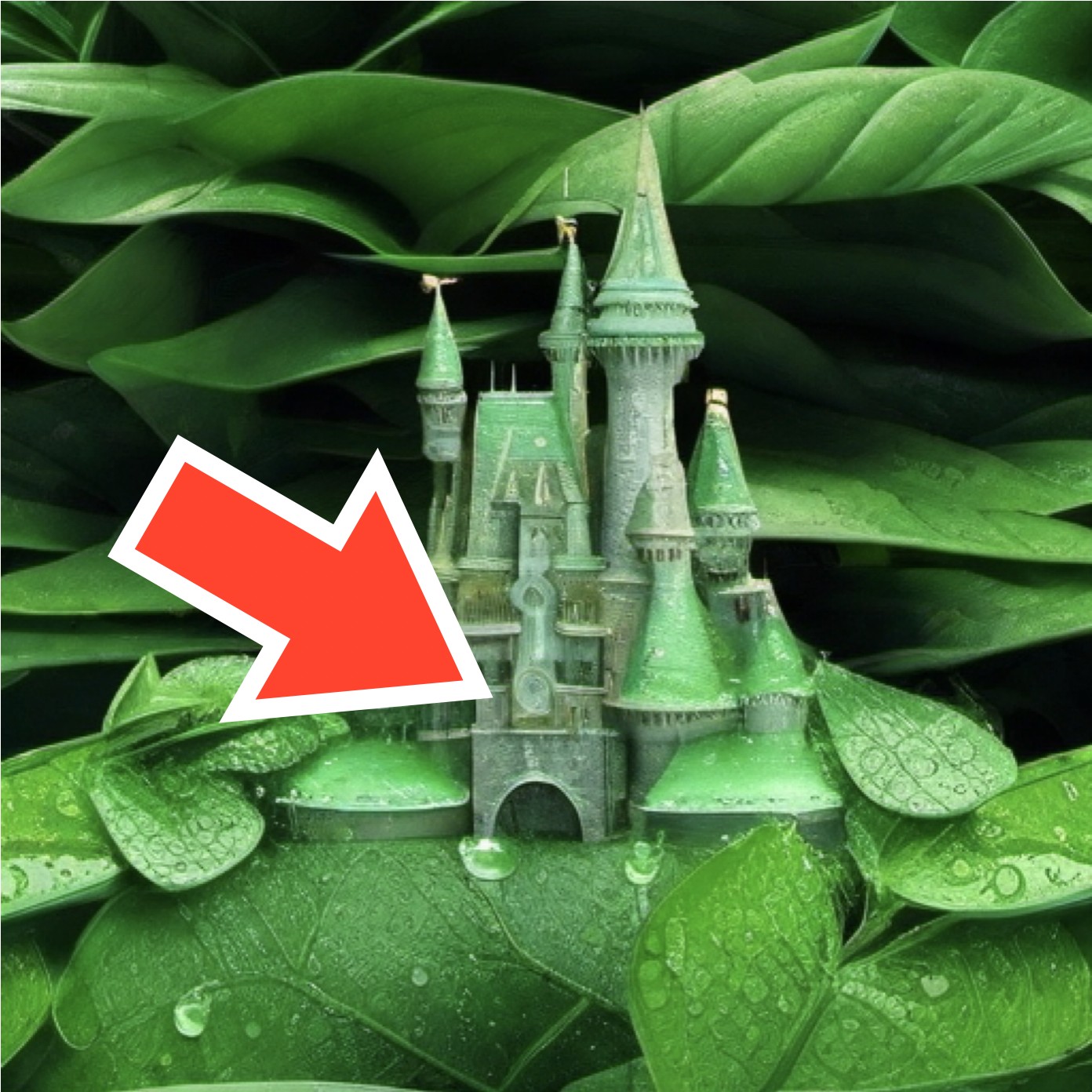} &
    \includegraphics[width=\linewidth]{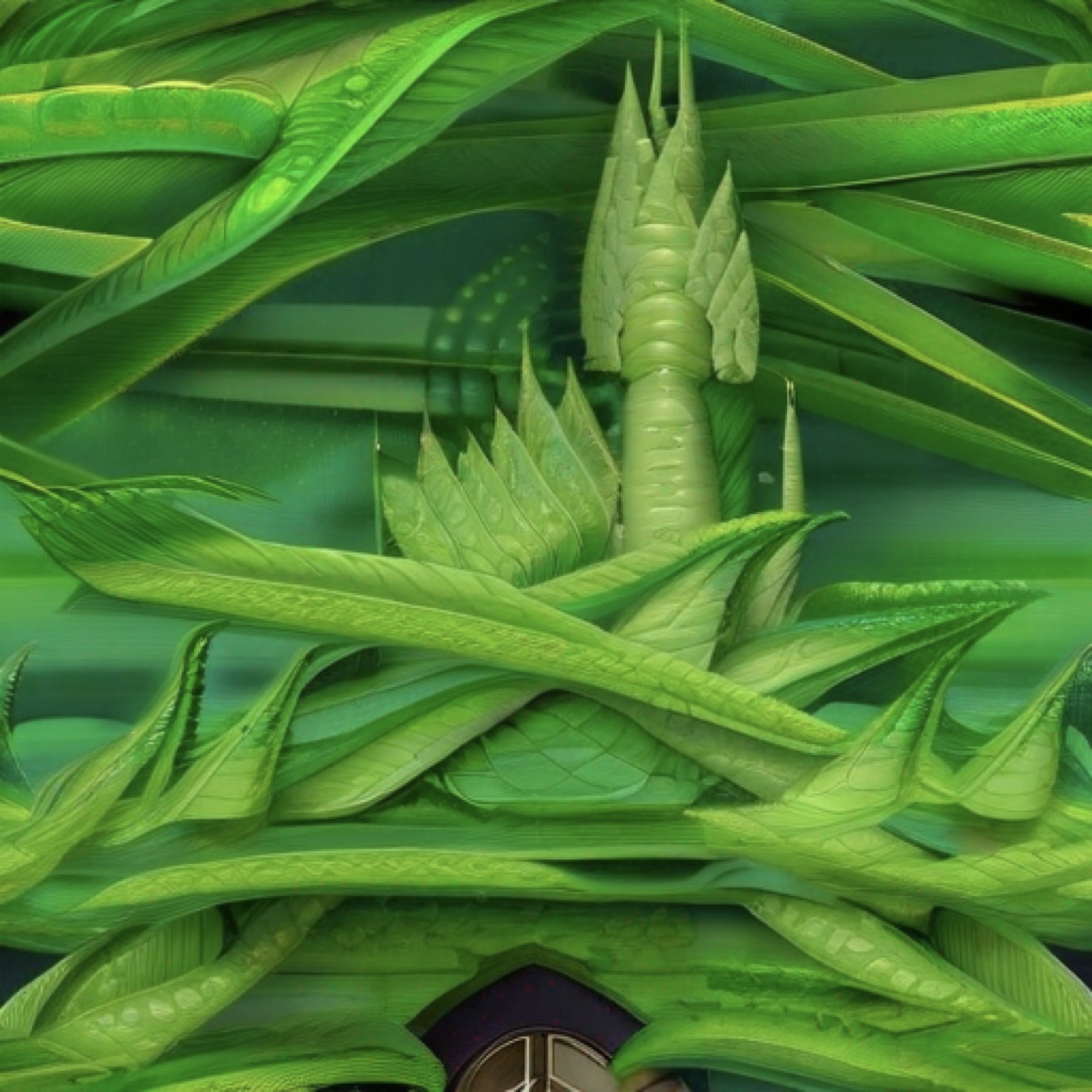} &
    \includegraphics[width=\linewidth]{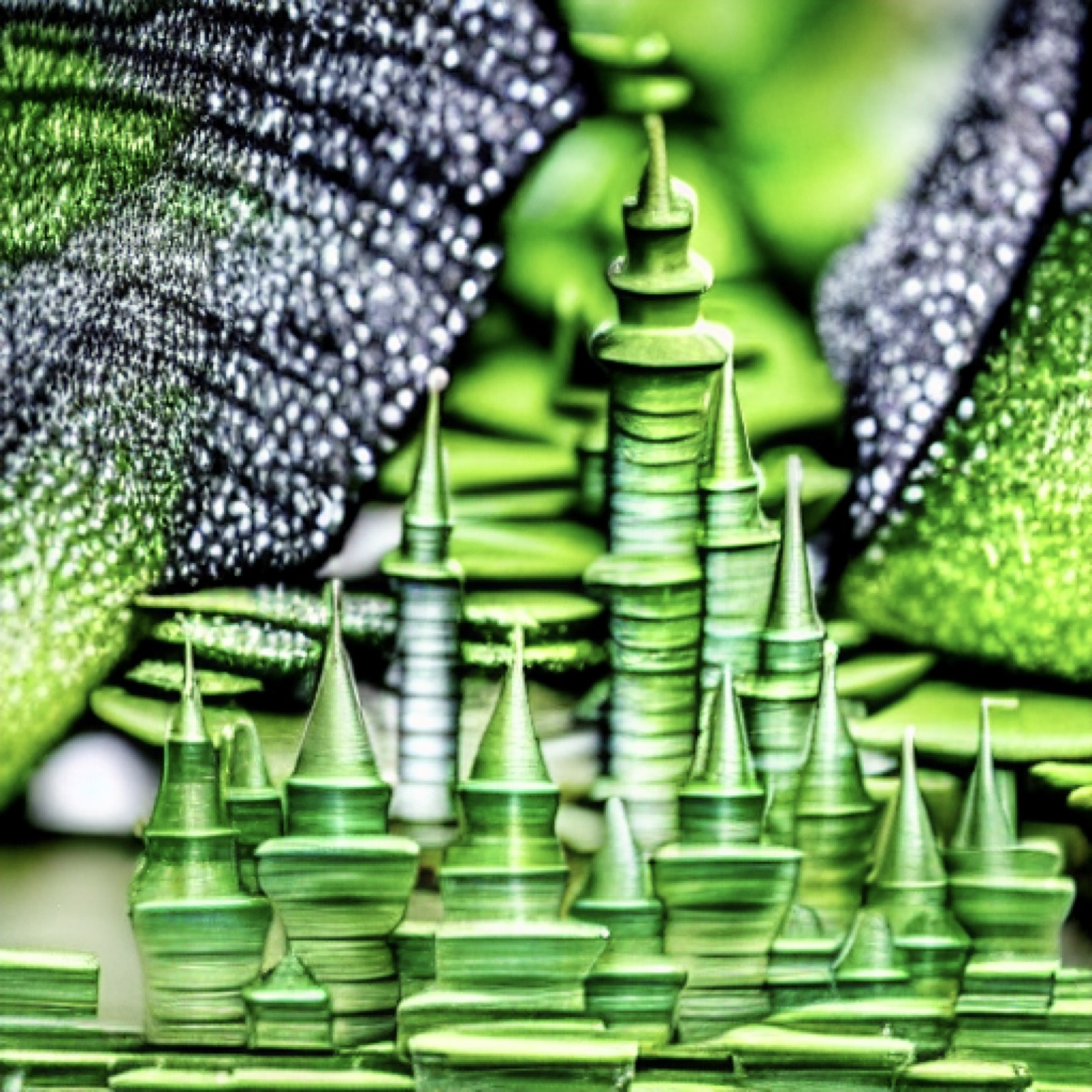} &
    \includegraphics[width=\linewidth]{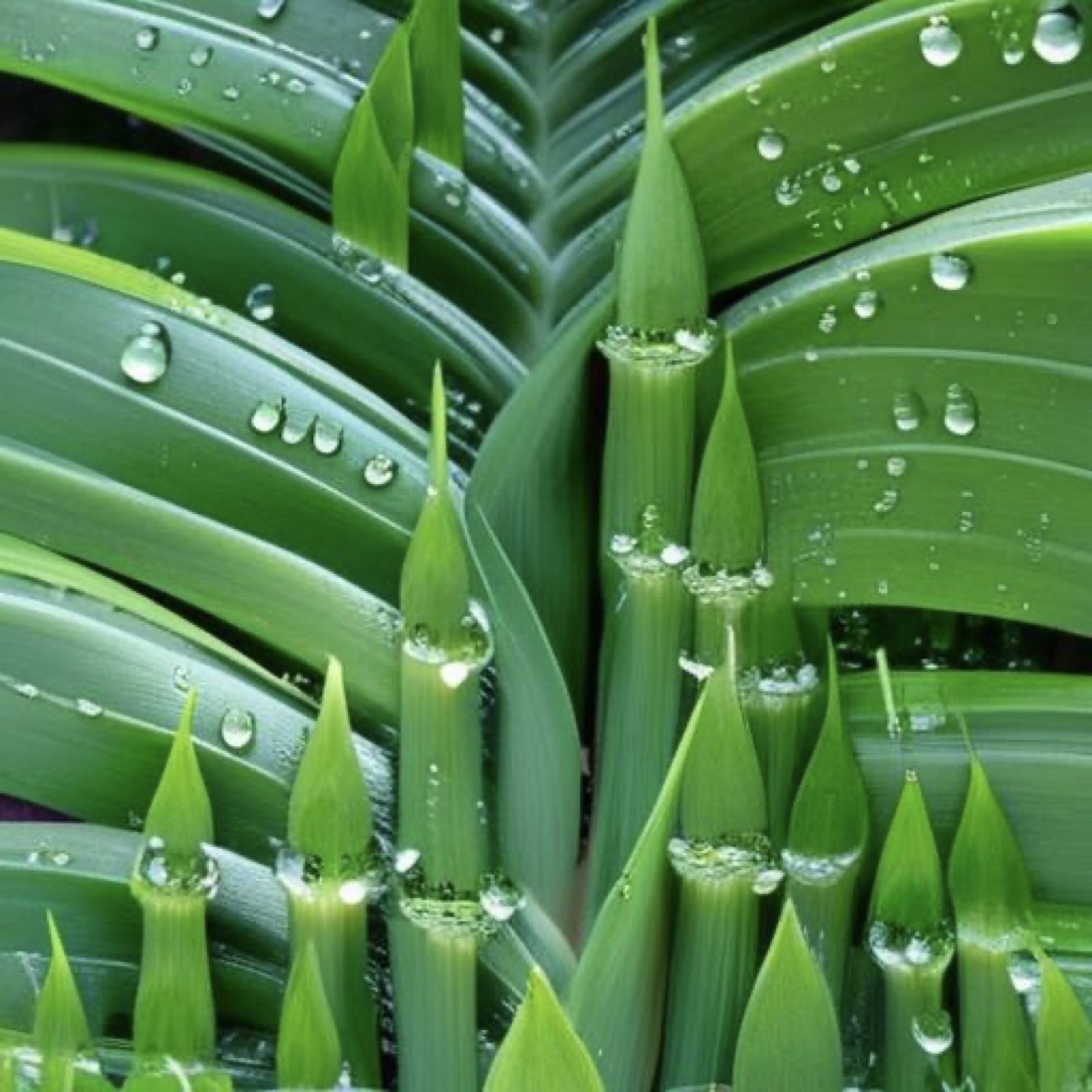} &
    \includegraphics[width=\linewidth]{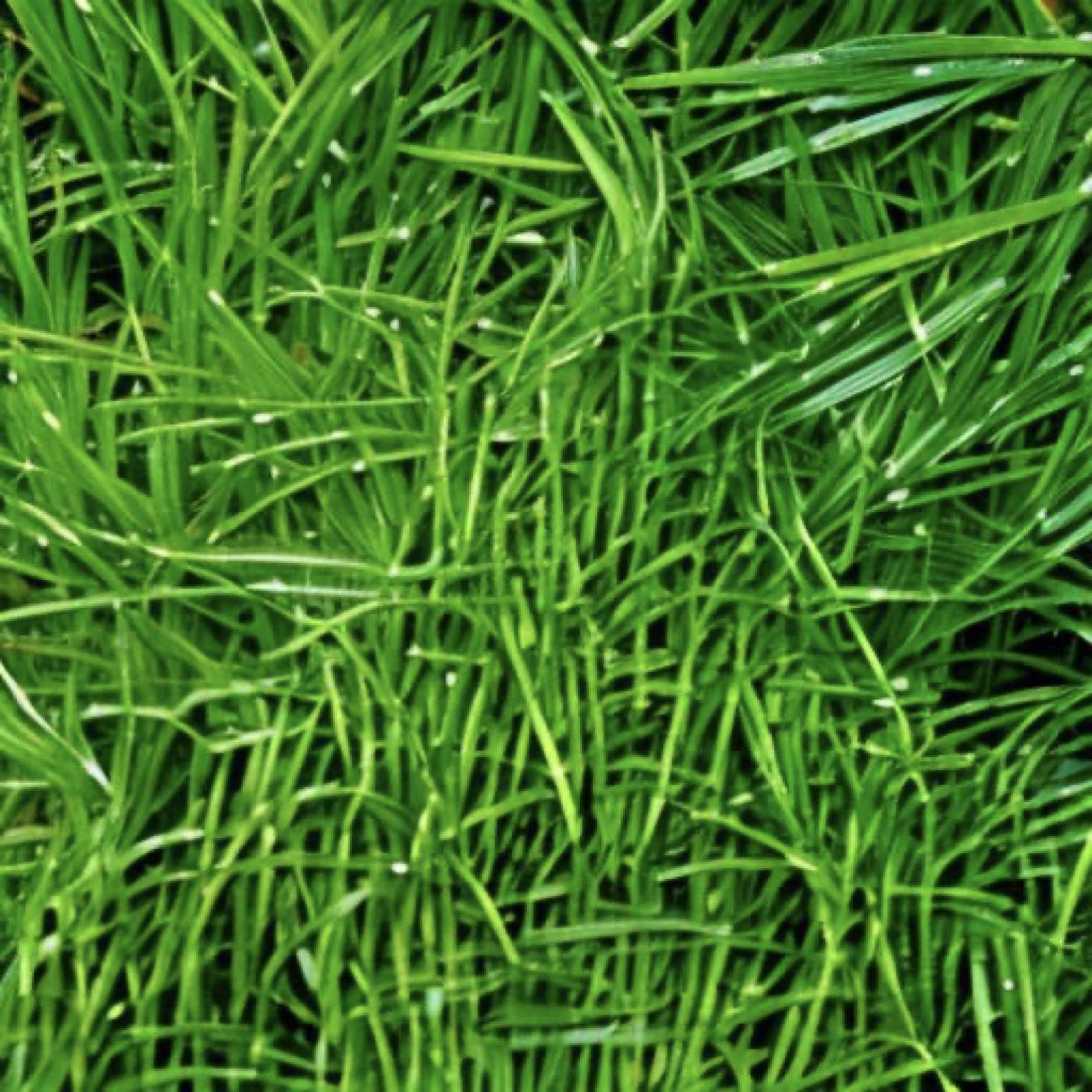}\\
\end{tabular}
}
\vspace{-12pt}
\caption{\textbf{Qualitative results of Style images.} DRF outperforms the baselines with appearance image style to structure image.}
\vspace{-5pt}
\label{fig:style_comparison_qual}
\end{figure*}
\begin{table*}[t]  
\centering
\footnotesize  
\setlength{\tabcolsep}{4pt}      
\renewcommand{\arraystretch}{1.1} 

\scalebox{0.75}{%
\begin{tabular}{l|ccc|ccc|ccc|ccc|c} 
\toprule
\multirow{2}{*}{\textbf{Method}}
& \multicolumn{3}{c|}{\textbf{Mesh \cite{hanocka2020point2mesh}}}
& \multicolumn{3}{c|}{\textbf{Pose}}
& \multicolumn{3}{c|}{\textbf{Point cloud \cite{huang2024surface}}}
& \multicolumn{3}{c|}{\textbf{Canny}}
& \multirow{2}{*}{\textbf{Successive Rate}} 
\\
\cline{2-13}
& Self-Sim $\downarrow$ & CLIP $\uparrow$ & DINO-I $\uparrow$
& Self-Sim $\downarrow$ & CLIP $\uparrow$ & DINO-I $\uparrow$
& Self-Sim $\downarrow$ & CLIP $\uparrow$ & DINO-I $\uparrow$
& Self-Sim $\downarrow$ & CLIP $\uparrow$ & DINO-I $\uparrow$
&  
\\
\midrule
\textcolor{gray}{T2I-Adapter+IP-Adapter}~\cite{mou2024t2i, ye2023ip} 
& \textcolor{gray}{0.2374} & \textcolor{gray}{0.3062} & \textcolor{gray}{0.6627}
& \textcolor{gray}{0.2949} & \textcolor{gray}{0.2865} & \textcolor{gray}{0.5304}
& \textcolor{gray}{0.1749} & \textcolor{gray}{0.2708} & \textcolor{gray}{\textbf{0.4224}}
& \textcolor{gray}{0.0457} & \textcolor{gray}{\textbf{0.3192}} & \textcolor{gray}{0.6467}
& \textcolor{gray}{0.9718}
\\
\textcolor{gray}{ControlNet+IP-Adapter}~\cite{zhang2023adding, ye2023ip}
& \textcolor{gray}{0.2024} & \textcolor{gray}{0.3320} & \textcolor{gray}{0.6068}
& \textcolor{gray}{0.3035} & \textcolor{gray}{0.2904} & \textcolor{gray}{0.6857}
& \textcolor{gray}{0.1328} & \textcolor{gray}{0.2966} & \textcolor{gray}{0.3169}
& \textcolor{gray}{0.0497} & \textcolor{gray}{0.3164} & \textcolor{gray}{0.5875}
& \textcolor{gray}{0.8873}
\\
\textcolor{gray}{Uni-ControlNet}~\cite{qin2023unicontrol}
& \textcolor{gray}{0.2188} & \textcolor{gray}{0.2997} & \textcolor{gray}{0.6232}
& \textcolor{gray}{0.3348} & \textcolor{gray}{0.3077} & \textcolor{gray}{0.6450}
& \textcolor{gray}{0.1302} & \textcolor{gray}{0.2836} & \textcolor{gray}{0.2918}
& \textcolor{gray}{\textbf{0.0421}} & \textcolor{gray}{0.3160} & \textcolor{gray}{0.5530}
& \textcolor{gray}{0.7605}
\\
\midrule
FreeControl~\cite{mo2024freecontrol}
& \textbf{0.1503} & 0.3270 & 0.7288
& 0.2839 & 0.2880 & 0.6162
& 0.1445 & 0.2928 & 0.2552
& 0.0533 & 0.3084 & \textbf{0.7381}
& 0.9152
\\
Ctrl-X~\cite{lin2025ctrl}
& 0.1542 & 0.3464 & 0.7139
& 0.2332 & 0.3429 & 0.7378
& 0.1205 & \textbf{0.3295} & 0.2654
& 0.0493 & 0.3075 & 0.7335
& 0.9577
\\
\rowcolor{gray!30}
\textbf{DRF (Ours)}
& 0.1532 & \textbf{0.3492} & \textbf{0.7342}
& \textbf{0.2294} & \textbf{0.3503} & \textbf{0.7391}
& \textbf{0.1114} & 0.3282 & 0.2923
& 0.0473 & 0.3102 & 0.7135
& \textbf{0.9859}
\\
\bottomrule
\end{tabular}
} 
\vspace{-9pt}
\caption{\textbf{Quantitative results} Compared with baselines. Metrics measured with various structures (Mesh, Pose, Point cloud, Canny). DINO-ViT self-similarity \cite{tumanyan2022splicing} is lower-better ($\downarrow$), CLIP \cite{radford2021learning} and DINO-I \cite{ruiz2023dreambooth} are higher-better ($\uparrow$).}
\vspace{-14pt}
\label{tab:main_quan}
\end{table*}

\noindent\textbf{Baselines.}\quad
We compare our method against recent controllable T2I diffusion models~\cite{lin2025ctrl, ye2023ip, mo2024freecontrol, zhang2023adding, qin2023unicontrol, mou2024t2i} as baselines. For all experiments except Uni-ControlNet and T2I-Adapter, we use the SDXL~\cite{podell2023sdxl} model optimized for 512×512 images to ensure a fair comparison. The implementations of the baselines are conducted by referring to the official source code provided for each respective method.\\
\noindent\textbf{Dataset.}\quad Structure images, sourced from prior studies~\cite{hanocka2020point2mesh, huang2024surface, zhou2016thingi10k}, comprise 30\% ControlNet-supported conditions (\textit{e.g.,} Canny, skeleton), 51\% in-the-wild conditions (\textit{e.g.,} 3D mesh, point cloud), and 19\% natural images. Appearance images are collected from open-source datasets~\cite{agustsson2017ntire,schuhmann2022laion} and generated images. Text prompts for structure, appearance, and generation are defined by combining templates with manual annotations.\\
\noindent\textbf{Evaluation metrics.}\quad  Structure and appearance control are evaluated using widely adopted metrics in T2I diffusion models. Specifically, DINO Self-sim~\cite{tumanyan2022splicing} metric quantifies structure preservation by computing the self-similarity distance between the structure image and the generated output in the DINO-ViT~\cite{caron2021emerging} feature space, while the DINO-I~\cite{ruiz2023dreambooth} metric assesses appearance transfer through cosine similarity of the DINO-ViT tokens. CLIP score~\cite{radford2021learning} measures text-image similarity for prompt-based control generation. 

In addition, categorical fidelity is quantified by a \textit{successive rate}. Each synthesized image is passed through an ImageNet-pretrained ResNet-18~\cite{he2016deep}, whose top-1 prediction is compared with the target class (e.g., “tiger,” ImageNet class 292). A concordant prediction is counted as a success; any discrepancy is deemed a failure. The reported score is the proportion of successes averaged over the evaluation set.
\section{Results}
\label{sec:results}
\subsection{Qualitative results}

We evaluate the proposed method against baselines in three appearance-structure fusion scenarios. (1) As the first to fourth rows of \cref{fig:comparison_qual} demonstrate, our approach reliably aligns the appearance image with diverse structure types (mesh, skeleton, canny, 3D) while preserving its key features. In contrast, baselines often fail by losing appearance details or generating incoherent results. (2) The last two rows in \cref{fig:comparison_qual} show that simply adjusting the prompt for the structure object allows us to inject texture and color from the appearance image into the structure image without sacrificing the structure’s essential objects or attributes. (3) Lastly, when applying an illustrative style of appearance image to the structure image as depicted in \cref{fig:style_comparison_qual}, our method remains robust and versatile, consistently producing high-quality results across different styles.

\begin{figure}[H] 
\footnotesize
\centering 

\newcommand{\imgwidth}{0.23\linewidth} 

\begin{tikzpicture}[x=1cm, y=1cm]
    \node[anchor=south] (FigA1) at (0,0) {
        \includegraphics[width=\imgwidth]{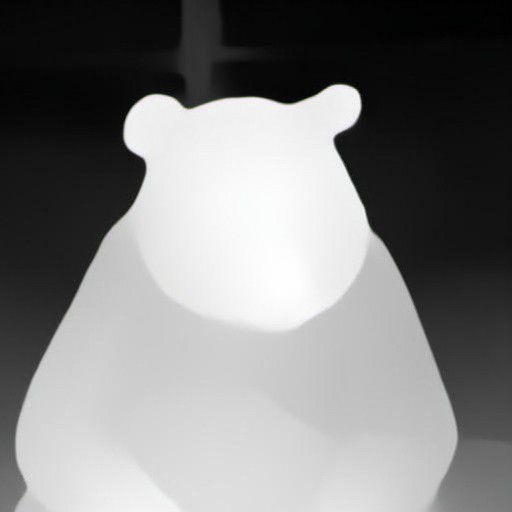}
    };
\end{tikzpicture}\hspace{-1mm}%
\begin{tikzpicture}[x=1cm, y=1cm]
    \node[anchor=south] (FigD1) at (0,0) {
        \includegraphics[width=\imgwidth]{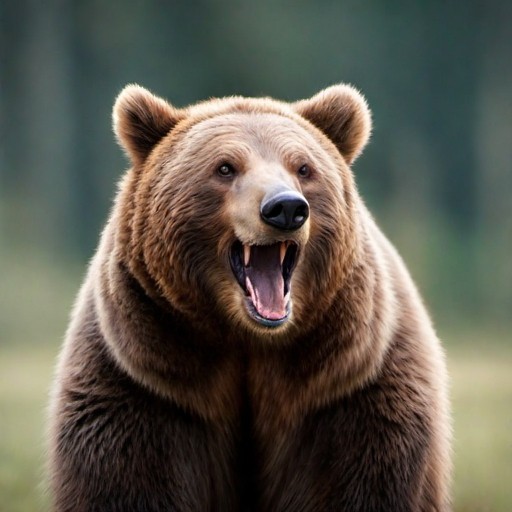}
    };
\end{tikzpicture}\hspace{-1mm}%
\begin{tikzpicture}[x=1cm, y=1cm]
    \node[anchor=south] (FigC1) at (0,0) {
        \includegraphics[width=\imgwidth]{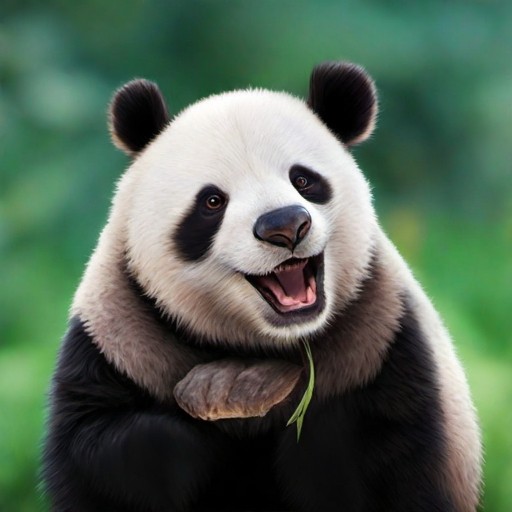}
    };
\end{tikzpicture}\hspace{-1mm}%
\begin{tikzpicture}[x=1cm, y=1cm]
    \node[anchor=south] (FigB1) at (0,0) {
        \includegraphics[width=\imgwidth]{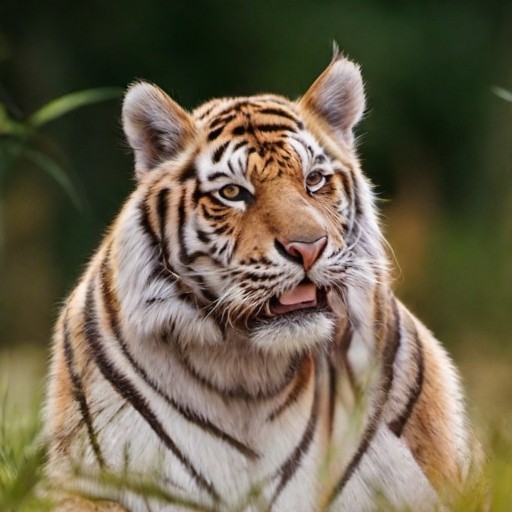}
    };
\end{tikzpicture}
\\[-4pt]
\vspace{0pt}
\setulcolor{Dandelion}
\setul{0.2pt}{1.5pt}
\centering \textit{``A photo of a \ul{bear} /  \ul{panda} /  \ul{tiger} smile looking in front"} 

\begin{tikzpicture}[x=1cm, y=1cm]
    \node[anchor=south] (FigA1) at (0,0) {
        \includegraphics[width=\imgwidth]{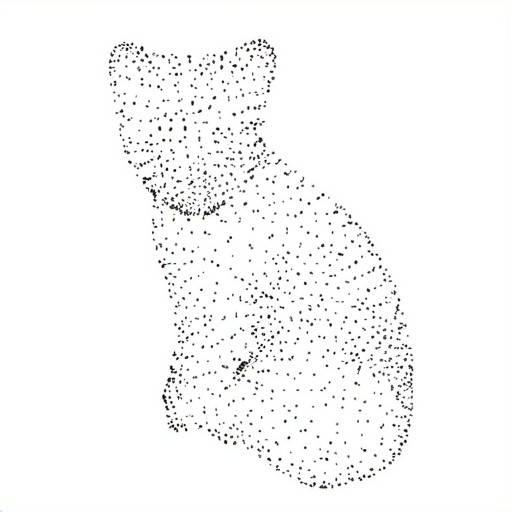}
    };
\end{tikzpicture}\hspace{-1mm}%
\begin{tikzpicture}[x=1cm, y=1cm]
    \node[anchor=south] (FigD1) at (0,0) {
        \includegraphics[width=\imgwidth]{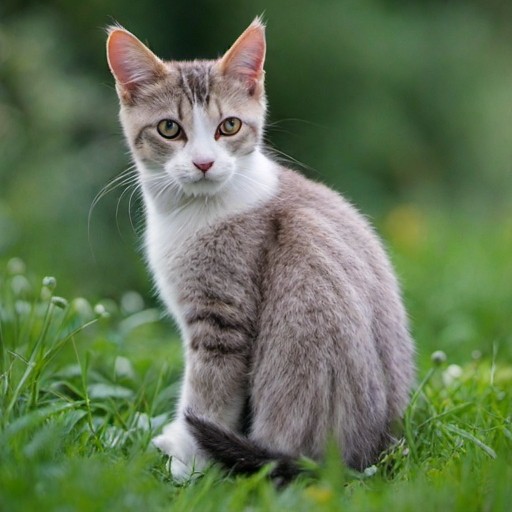}
    };
\end{tikzpicture}\hspace{-1mm}%
\begin{tikzpicture}[x=1cm, y=1cm]
    \node[anchor=south] (FigC1) at (0,0) {
        \includegraphics[width=\imgwidth]{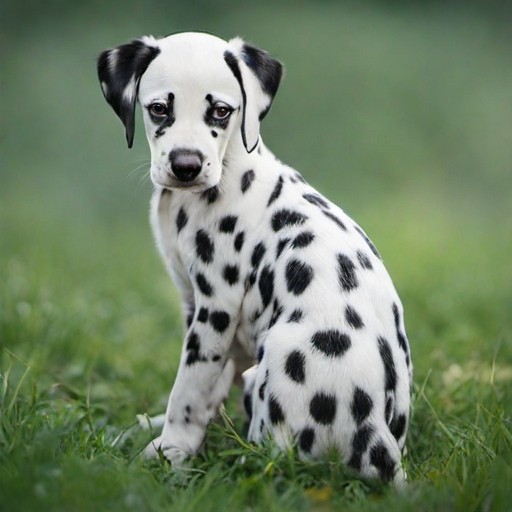}
    };
\end{tikzpicture}\hspace{-1mm}%
\begin{tikzpicture}[x=1cm, y=1cm]
    \node[anchor=south] (FigB1) at (0,0) {
        \includegraphics[width=\imgwidth]{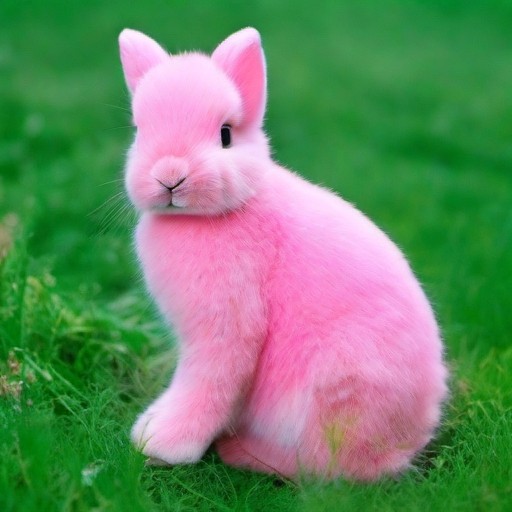}
    };
\end{tikzpicture}
\\[-4pt]
\vspace{0pt}
\setulcolor{Dandelion}
\setul{0.2pt}{1.5pt}
\centering \textit{``A photo of a \ul{cat} / \ul{dalmatian dog} / \ul{pink bunny} sitting on the grass "} 

\vspace{-5pt}
\caption{\textbf{Qualitative results of conditional Text-to-Image.} Our method shows that T2I is possible using images produced by diffusion, even when an appearance image is not given.}
\vspace{-5pt}
\label{fig:t2i}
\end{figure}

We also demonstrate that controllable T2I generation can be achieved solely from a structure image by first producing a text-guided image, then using it as DRF’s appearance reference as shown in \cref{fig:t2i}. 

%
\subsection{Quantitative results}
As reported in \cref{tab:main_quan}, we evaluate Self-Sim~\cite{tumanyan2022splicing}, CLIP score~\cite{radford2021learning}, and DINO-I~\cite{ruiz2023dreambooth} on four structure datasets. While our method often attains the best scores, each metric focuses on a single aspect—similarity to either structure, appearance, or the prompt—so we additionally measure a \textit{successive rate} (procedure in \cref{sec:experiments}) that quantifies how faithfully both images are fused. Notably, our approach achieves the highest successive rate, even on a dataset with a lower DINO-I score, demonstrating robust generation performance.

We also conduct a user study on 20 comparison sets  that include five different structure datasets (real images, sketches, meshes, poses, and 3D) against five baselines. Participants evaluated how well each method \textit{(1) matched the text prompt, (2) preserved structure details, and (3) retained appearance characteristics.} As \cref{tab:user_study} shows, our method outperforms all baselines in every category, indicating it not only aligns with the text prompt but also maintains core features of both the structure and appearance images.
\setul{0.3ex}{0.2ex}
\newcommand{\ylu}[1]{%
  {\setulcolor{WildStrawberry}\ul{#1}}%
}

\vspace{-3pt}
\begin{table*}[!t]
  \centering
  \scriptsize
  \setlength\tabcolsep{3pt} 
  \resizebox{\textwidth}{!}{%
    \begin{tabular}{c *{4}{cccc}}
      \toprule
      Steps
      & \multicolumn{4}{c}{\bfseries DPM-Solver++ (Multi-step)~\cite{lu2025dpm} + DRF}
      & \multicolumn{4}{c}{\bfseries UniPC~\cite{zhao2023unipc} + DRF}
      & \multicolumn{4}{c}{\bfseries DPM-Solver (Single-step)~\cite{lu2022dpm} + DRF}
      & \multicolumn{4}{c}{\bfseries DDIM~\cite{song2020denoising} + DRF (Ours)}
      \\
      \cmidrule(lr){2-5}  \cmidrule(lr){6-9}
      \cmidrule(lr){10-13}\cmidrule(lr){14-17}
      & Clip $\uparrow$ & Self-Sim $\downarrow$ & DINO-I $\uparrow$ & Time (s)
      & Clip $\uparrow$ & Self-Sim $\downarrow$ & DINO-I $\uparrow$ & Time (s)
      & Clip $\uparrow$ & Self-Sim $\downarrow$ & DINO-I $\uparrow$ & Time (s)
      & Clip $\uparrow$ & Self-Sim $\downarrow$ & DINO-I $\uparrow$ & Time (s)
      \\
      \midrule
      5  & 0.3117 & 0.2126 & 0.444 & 8.80  & 0.3076 & 0.2129 & 0.646 & 8.78  & 0.1885 & 0.2622 & 0.544 & 8.64  & 0.3267 & 0.1613 & 0.549 & 7.68  \\
      10 & 0.3256 & \cellcolor{gray!30}\textbf{0.1605} & \cellcolor{gray!30}\textbf{0.826} & \cellcolor{gray!30}\textbf{15.56} & 0.3324 & 0.1898 & 0.765 & 16.06 & 0.3070 & 0.2398 & 0.706 & 16.04 & 0.3394 & 0.1917 & 0.768 & 13.52 \\
      20 & 0.3202 & 0.2449 & 0.843 & 29.81 & 0.3023 & 0.1653 & 0.622 & 30.15 & 0.3449 & 0.1721 & 0.870 & 29.51 & 0.3270 & 0.2061 & 0.854 & 25.33 \\
      30 & 0.3204 & 0.1904 & 0.846 & 43.27 & 0.3074 & 0.1832 & 0.792 & 42.86 & 0.3242 & 0.2207 & 0.770 & 44.15 & 0.3492 & 0.1204 & 0.891 & 35.74 \\
      40 & 0.3290 & 0.2225 & 0.776 & 61.14 & 0.2912 & 0.1667 & 0.762 & 58.68 & 0.3019 & 0.2337 & 0.827 & 57.54 & 0.3249 & 0.1628 & 0.836 & 47.87 \\
      50 & 0.3395 & 0.1864 & 0.851 & 70.45 & 0.2925 & 0.1790 & 0.717 & 72.35 & 0.3474 & 0.1895 & 0.840 & 70.78 & 0.3343 & 0.1533 & 0.863 & 58.50 \\
      \bottomrule
    \end{tabular}
    }
    \vspace{-8pt}
    \caption{\textbf{Scheduler-agnostic performance of DRF.} Quantitative results for DRF combined with four diffusion schedulers~\cite{lu2022dpm,lu2025dpm,song2020denoising,zhao2023unipc}. DRF delivers consistently strong fidelity regardless of the solver, and pairing it with the faster DPM-Solver++ cuts latency by roughly $\times3$ relative to DDIM while preserving perceptual quality.}
  \label{scheduler}
  \vspace{-15pt}
\end{table*}
We measure inference time, preprocessing time, and peak GPU memory usage (\cref{tab:latency}). Although our dual recursive approach involves multiple fixed points and thus a higher runtime than Ctrl-X~\cite{lin2025ctrl}, the overhead can be reduced by applying DRF only at selected iterations. This enables a flexible trade-off between speed and performance. Moreover, because our method requires no training, the slight loss of efficiency can be compensated for by the gain in performance.
\newcommand{\cmark}{\ding{51}} 
\newcommand{\xmark}{\ding{55}} 

\begin{table}[!h]
\centering
\setlength{\tabcolsep}{10pt} 
\renewcommand{\arraystretch}{1} 

\scalebox{0.6}{
\begin{tabular}{l c c c}
\toprule
\multicolumn{4}{c}{\textbf{User Preference Rate (\%)}}\\
\midrule
Method & 
Text ($\uparrow$) & 
\begin{tabular}[c]{@{}c@{}} Structure ($\uparrow$)\end{tabular} &
\begin{tabular}[c]{@{}c@{}}Appearance ($\uparrow$)\end{tabular} 
\\
\midrule
T2I-Adapter + IP-Adapter \cite{mou2024t2i, ye2023ip}
& 7.528
& 8.092
& 11.43
\\
ControlNet + IP Adapter \cite{zhang2023adding, ye2023ip}
& 16.02
& 12.90
& 14.14
\\
Uni-ControlNet \cite{qin2023unicontrol}
& 10.61
& 10.98
& 9.302
\\
FreeControl \cite{mo2024freecontrol}
& 12.93
& 14.83
& 17.63
\\
Ctrl-X \cite{lin2025ctrl}
& 17.37
& 9.441
& 13.75
\\
\rowcolor{gray!30}
{\textbf{DRF (Ours)}}
& {\textbf{35.52}}
& \textbf{43.73}
& \textbf{33.72}
\\
\bottomrule
\end{tabular}
}
\vspace{-8pt}
\caption{\textbf{User study} shows that our method achieved the highest rates over all questions for controllable T2I.}
\vspace{-7.5pt}
\label{tab:user_study}
\end{table}

\begin{table}[!h]
\centering
\setlength{\tabcolsep}{10pt} 
\renewcommand{\arraystretch}{1} 
\scalebox{0.6}{
\begin{tabular}{l c c c}
\toprule
Method & 
Training & 
\begin{tabular}[c]{@{}c@{}}Inference \\ latency (s)\end{tabular} &
\begin{tabular}[c]{@{}c@{}}Peak GPU memory \\ usage (GiB)\end{tabular} 
\\
\midrule
\midrule
\textcolor{gray}{T2I-Adapter + IP-Adapter}~\cite{mou2024t2i, ye2023ip}
& \textcolor{gray}{\cmark}
& \textcolor{gray}{11.07}
& \textcolor{gray}{4.45}
\\
\textcolor{gray}{ControlNet + IP Adapter}~\cite{zhang2023adding, ye2023ip}
& \textcolor{gray}{\cmark} 
& \textcolor{gray}{23.39}
& \textcolor{gray}{13.68}
\\
\textcolor{gray}{Uni-ControlNet}~\cite{qin2023unicontrol}
& \textcolor{gray}{\cmark}
& \textcolor{gray}{18.74}
& \textcolor{gray}{8.73}
\\
FreeControl~\cite{mo2024freecontrol}
& \xmark
& 154.76
& 18.69
\\
Ctrl-X~\cite{lin2025ctrl}
& \xmark
& 15.52
& 13.07
\\
\textbf{Ours-DRF $N=1$}
& \xmark
& 29.69
& 18.71 \\
\rowcolor{gray!30}
\textbf{Ours-DRF $N=3$} & \xmark & 56.87 & 18.71 \\
\bottomrule
\end{tabular}
} 
\vspace{-8pt}
\caption{\textbf{Inference efficiency} DRF, as a training-free method, exhibits variations in inference time depending on the number of iterations $N$.}
\vspace{-14pt}
\label{tab:latency}
\end{table}

\section{Discussions}
\label{sec:discussion}

\subsection{Plug-and-Play enhancement with DRF}
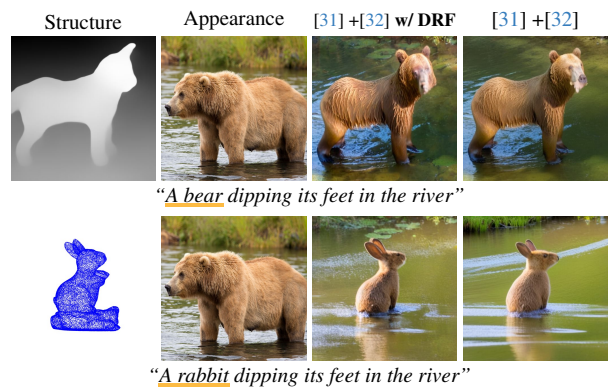
\begin{figure}[H] 
\footnotesize
\centering 

\newcommand{\imgwidth}{0.23\linewidth} 

\begin{tikzpicture}[x=1cm, y=1cm]
    \node[anchor=south] (FigA1) at (0,0) {
        \includegraphics[width=\imgwidth]{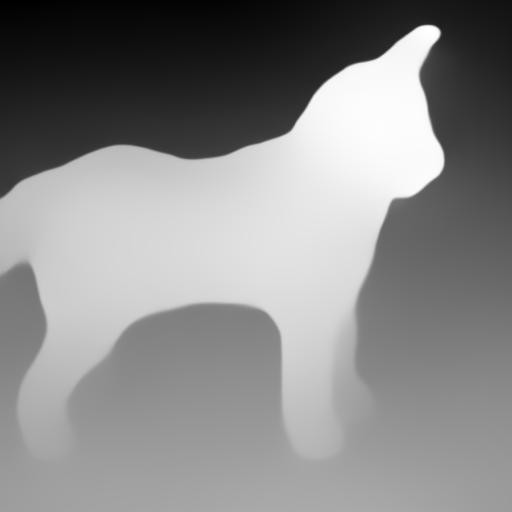}
    };
    \node[anchor=south, yshift=-1mm] at (FigA1.north) {\footnotesize Structure};
\end{tikzpicture}\hspace{-1mm}%
\begin{tikzpicture}[x=1cm, y=1cm]
    \node[anchor=south] (FigD1) at (0,0) {
        \includegraphics[width=\imgwidth]{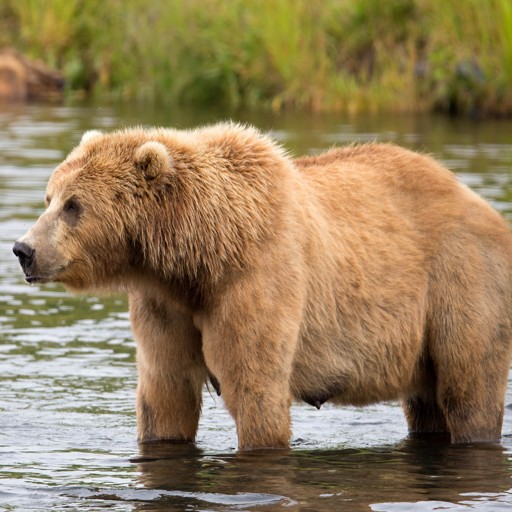}
    };
    \node[anchor=south, yshift=-1mm] at (FigD1.north) {\footnotesize Appearance};
\end{tikzpicture}\hspace{-1mm}%
\begin{tikzpicture}[x=1cm, y=1cm]
    \node[anchor=south] (FigC1) at (0,0) {
        \includegraphics[width=\imgwidth]{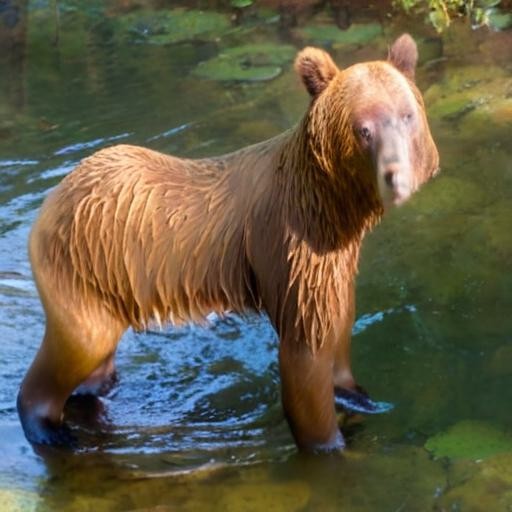}
    };
    \node[anchor=south, yshift=-1mm] at (FigC1.north) {\scriptsize \shortstack{\cite{ye2023ip} +\cite{zhang2023adding} \textbf{w/ DRF}}};
\end{tikzpicture}\hspace{-1mm}%
\begin{tikzpicture}[x=1cm, y=1cm]
    \node[anchor=south] (FigB1) at (0,0) {
        \includegraphics[width=\imgwidth]{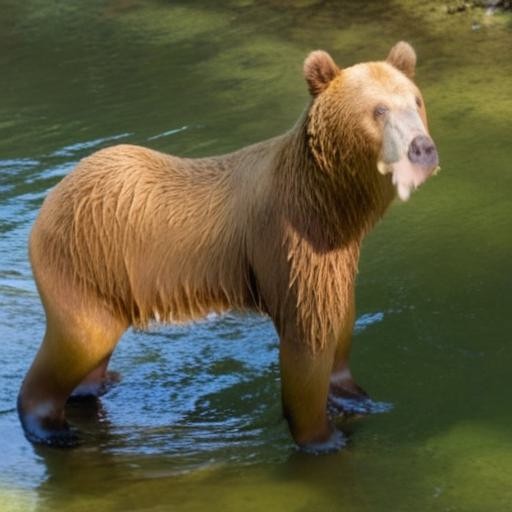}
    };
    \node[anchor=south, yshift=-1mm] at (FigB1.north) 
    {\footnotesize \shortstack{\cite{ye2023ip} +\cite{zhang2023adding}}};
\end{tikzpicture}
\\[-4pt]
\vspace{0pt}
\setulcolor{Dandelion}
\setul{0.2pt}{1.5pt}
\centering \textit{``\ul{A bear} dipping its feet in the river"}

\begin{tikzpicture}[x=1cm, y=1cm]
    \node[anchor=south] (FigA1) at (0,0) {
        \includegraphics[width=\imgwidth]{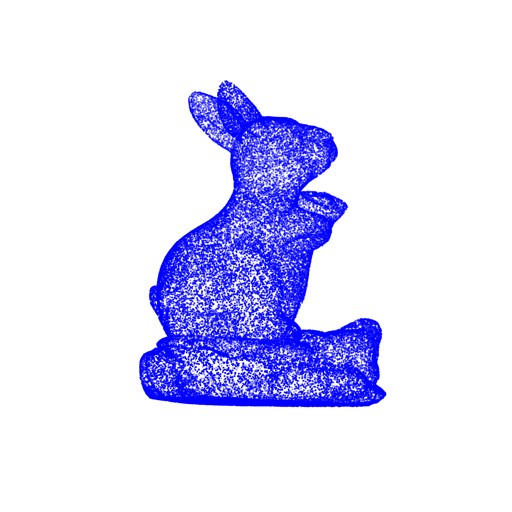}
    };
\end{tikzpicture}\hspace{-1mm}%
\begin{tikzpicture}[x=1cm, y=1cm]
    \node[anchor=south] (FigD1) at (0,0) {
        \includegraphics[width=\imgwidth]{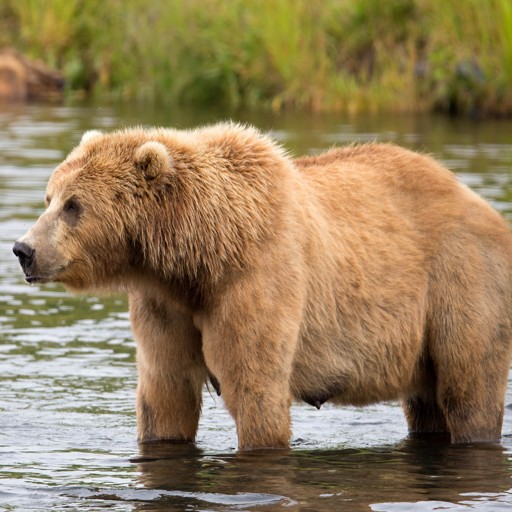}
    };
\end{tikzpicture}\hspace{-1mm}%
\begin{tikzpicture}[x=1cm, y=1cm]
    \node[anchor=south] (FigC1) at (0,0) {
        \includegraphics[width=\imgwidth]{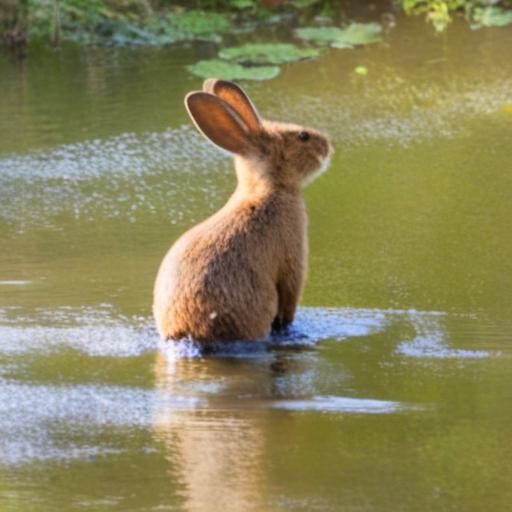}
    };
\end{tikzpicture}\hspace{-1mm}%
\begin{tikzpicture}[x=1cm, y=1cm]
    \node[anchor=south] (FigB1) at (0,0) {
        \includegraphics[width=\imgwidth]{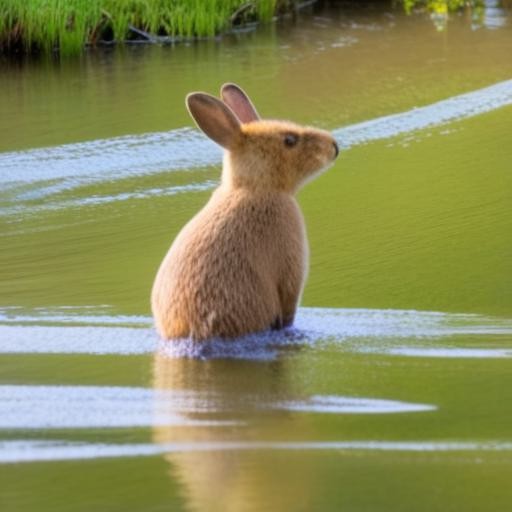}
    };
\end{tikzpicture}
\\[-4pt]
\vspace{0pt}
\setulcolor{Dandelion}
\setul{0.2pt}{1.5pt}
\centering \textit{``\ul{A rabbit} dipping its feet in the river"} 

\vspace{-5pt}
\caption{\textbf{Plug-and-Play of DRF.} Applying DRF to ControlNet + IP-Adapter~\cite{zhang2023adding,ye2023ip} yields better results than using baseline alone.}
\vspace{-9pt}
\label{fig:pnp}
\end{figure}

Our method can be integrated into existing T2I models in a plug-and-play manner, leading to performance improvements. Specifically, we add DRF process to the baseline model~\cite{zhang2023adding,ye2023ip} and observe improved image quality. As shown in \cref{fig:pnp}, incorporating our process leads to enhanced fine-grained details (row 1) while simultaneously producing images that more closely align with both the structure and the specified prompt. 
\vspace{-4pt}

\subsection{Scheduler-Agnostic DRF}
\label{DPMDRF}
Our approach is scheduler-agnostic, extending beyond the default DDIM scheduler ~\cite{song2020denoising} to operate effectively with a broad family of diffusion schedulers. As reported in \cref{scheduler}, coupling our method with the multi-step DPM-Solver++ ~\cite{lu2025dpm} reduces the sampling budget from 40 steps ($\sim$47 sec.) under DDIM to 10 steps ($\sim$15 sec.) while maintaining high perceptual fidelity. Thus, although the baseline DDIM already yields strong image quality, its latency bottleneck can be decisively mitigated by adopting faster solvers without compromising performance.

\subsection{DRF on Stable Diffusion}

To verify the advantages of our method extend beyond its native SDXL~\cite{podell2023sdxl} backbone, we port the plug-in to two architecturally distinct Stable Diffusion~\cite{rombach2022high} variants: (i) $\epsilon$-prediction SD 1.5, which directly regresses additive noise, and (ii) v-prediction SD 2.0, where the network estimates latent velocity. Quantitative results in \cref{tab:stablediffusion} show that our method preserves, and in some cases improves, image quality while delivering a substantial reduction in GPU memory consumption on both backbones. Additionally, based on validation of the schedulers natively embedded within each backbone (\cref{DPMDRF}), our method demonstrates compatibility irrespective of the underlying model architecture.

\begin{table}[htbp]
  \scriptsize
  \centering
  \resizebox{0.96\columnwidth}{!}{%
    \begin{tabular}{lccccc}
      \toprule
      Metrics & Clip $\uparrow$ & Self-Sim $\downarrow$ & DINO-I $\uparrow$ & Time (s) & GPU memory (GiB)\\
      \midrule
      SD 1.5~\cite{rombach2022high} + DRF& 0.3108 & 0.1820 & 0.6176 & 35.17 & \textbf{5.87} \\
      SD 2.0~\cite{rombach2022high} + DRF & 0.2934 & 0.2166 & 0.453 & 30.83 & 6.28 \\
      \textbf{SDXL~\cite{podell2023sdxl} + DRF (Ours)} & 0.3331 & 0.1586 & 0.6957 & 58.23 & 18.98 \\
      \bottomrule
    \end{tabular}%
  }
  \vspace{-8pt}
  \caption{\textbf{Cross-backbone DRF.} Plug-in DRF from SDXL~\cite{podell2023sdxl} to SD 1.5 and SD 2.0~\cite{rombach2022high} preserves fidelity while reducing GPU memory.}
  \vspace{-8pt}
  \label{tab:stablediffusion}
\end{table}

\section{Limitation}
\label{sec:limitation}

\noindent\textbf{Computational cost.}\quad
DRF adopts a dual framework of appearance feedback and generation feedback, inevitably increasing computational overhead and prolonging inference time. Adjusting the number of DRF iterations can mitigate this cost, but higher iteration counts produce sharper images at the expense of additional resources.

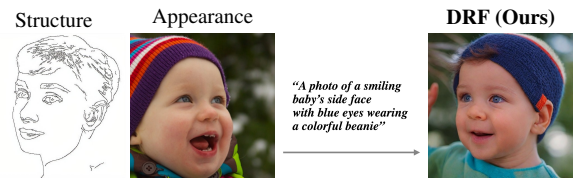
\begin{figure}[H] 
\scriptsize
\centering 

\newcommand{\imgwidth}{0.23\linewidth} 

\begin{tikzpicture}[x=1cm, y=1cm]
    \node[anchor=south] (FigA1) at (0,0) {
        \includegraphics[width=\imgwidth]{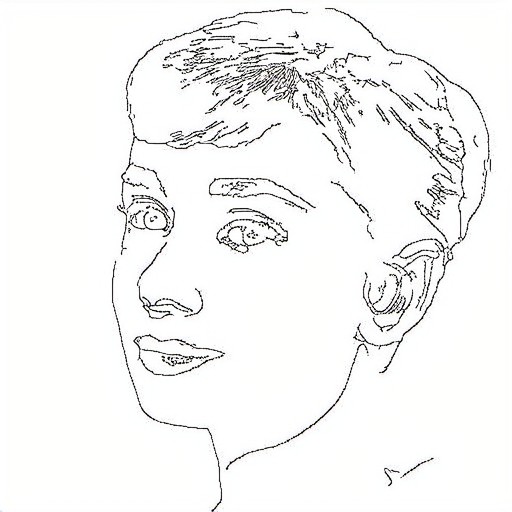}
    };
    \node[anchor=south, yshift=-1mm] at (FigA1.north) {\footnotesize Structure};
\end{tikzpicture}\hspace{-1mm}%
\begin{tikzpicture}[x=1cm, y=1cm]
    \node[anchor=south] (FigD1) at (0,0) {
        \includegraphics[width=\imgwidth]{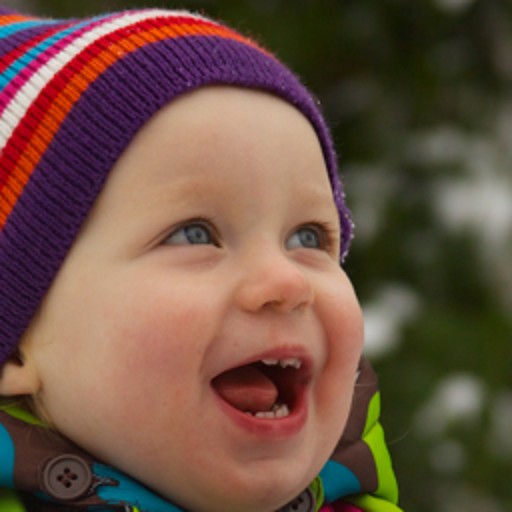}
    };
    \node[anchor=south, yshift=-1mm] at (FigD1.north) {\footnotesize Appearance};
\end{tikzpicture}\hspace{-1mm}%
\begin{tikzpicture}[x=1cm, y=1cm]
    \node[anchor=south] (FigC1) at (0,0) {
        \includegraphics[width=\imgwidth]{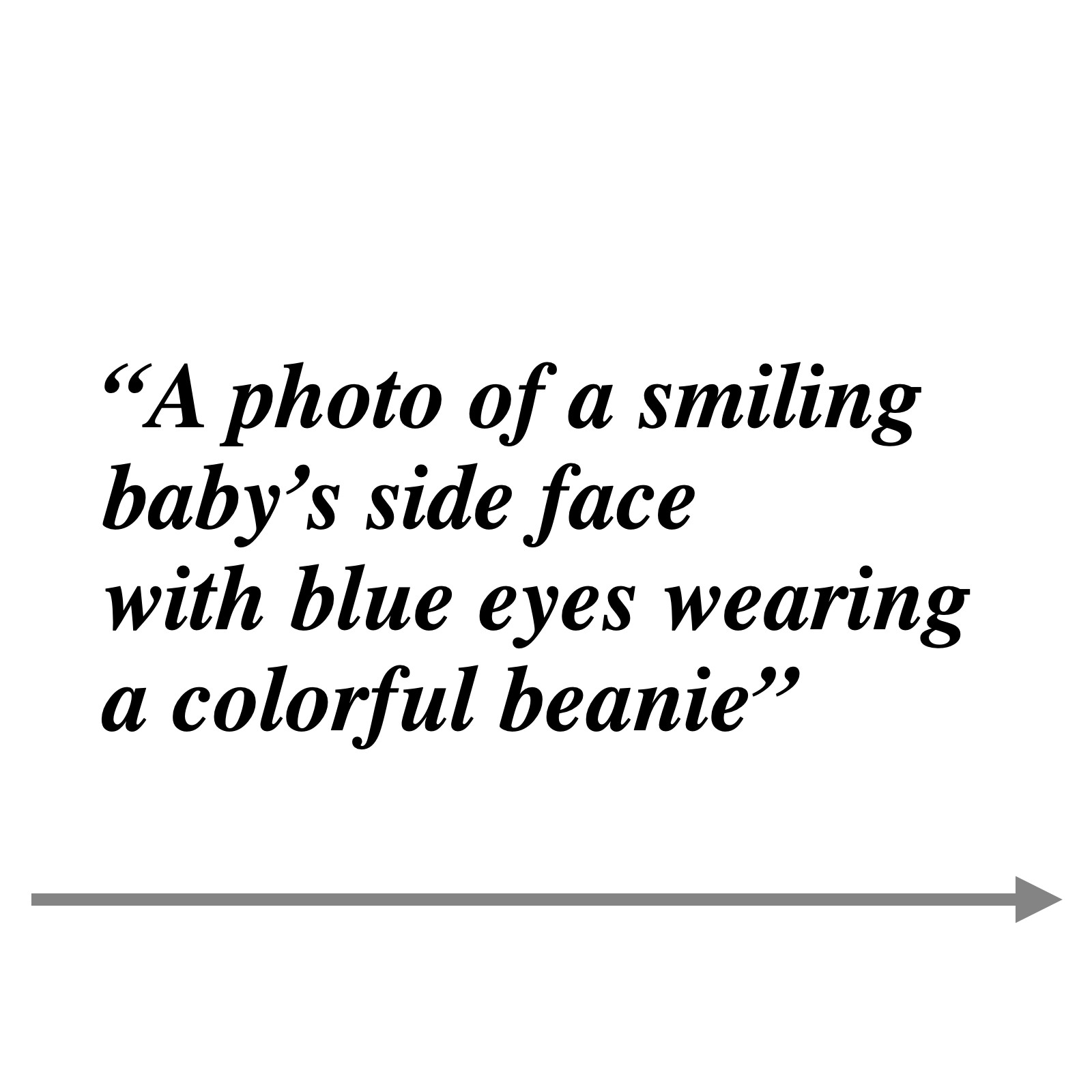}
    };
    \node[anchor=south, yshift=-1mm] at (FigD1.north) {\scriptsize };
\end{tikzpicture}\hspace{-1mm}%
\begin{tikzpicture}[x=1cm, y=1cm]
    \node[anchor=south] (FigB1) at (0,0) {
        \includegraphics[width=\imgwidth]{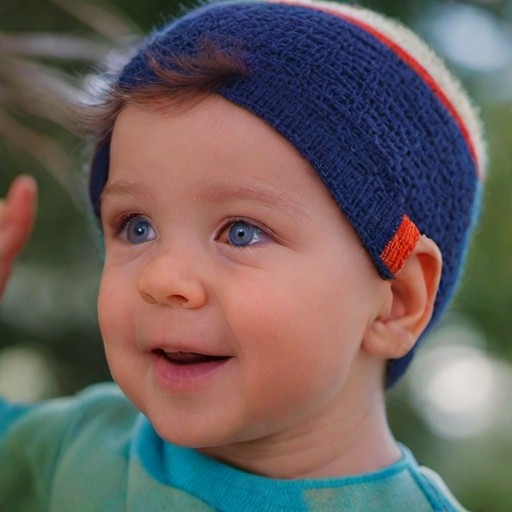}
    };
    \node[anchor=south, yshift=-1mm] at (FigD1.north) {\footnotesize \textbf{DRF (Ours)}};
\end{tikzpicture}

\vspace{-8pt}
\caption{\textbf{Weakness of appearance detail preservation}. }
\vspace{-10pt}
\label{fig:limitation}
\end{figure}

\noindent\textbf{Preservation of detail.}\quad DRF treats appearance and structure images as controllable T2I elements while providing generation guidance through user-defined prompts. While it handles most appearance images effectively. However as shown in \cref{fig:limitation}, it can struggle to retain fine-grained details for data such as the faces of ordinary individuals that the pretrained model did not encounter during training.
\vspace{-10pt}

\section{Conclusion}
\label{sec:conclusion}
DRF is a training-free T2I diffusion framework that provides recursive feedback to each control component for effective control of appearance and structure. This feedback mechanism ensures that the final generation output aligns coherently with both the appearance image and the text-conditioned score, resulting in high-quality generated images. By employing a dual feedback strategy, DRF resolves the longstanding issues in prior T2I methods, the loss of appearance fidelity, and the inability to produce stable image generations across diverse datasets.

Moreover, DRF exhibits robust performance in preserving structure and reflecting the intended appearance, as demonstrated in tasks such as pose transfer and image synthesis on class-invariant datasets. We further show that our approach can be seamlessly integrated into other T2I diffusion models, thereby offering a foundation for broader advancements in diffusion model sampling.
\newpage
\vspace{8pt}
\noindent
\section*{Acknowledgments}
\small{
This work was supported by Institute of Information \& communications Technology Planning \& Evaluation (IITP) grant funded by the Korea government(MSIT) (No.RS-2021-II212068, Artificial Intelligence Innovation Hub) and by the National Research Foundation of Korea(NRF) grant funded by the Korea government(MSIT) (RS-2024-00335741, RS2024-00357197).}

\noindent

{
    \small
    \bibliographystyle{ieeenat_fullname}
    \bibliography{main}

\begin{thebibliography}{34}
\providecommand{\natexlab}[1]{#1}
\providecommand{\url}[1]{\texttt{#1}}
\expandafter\ifx\csname urlstyle\endcsname\relax
  \providecommand{\doi}[1]{doi: #1}\else
  \providecommand{\doi}{doi: \begingroup \urlstyle{rm}\Url}\fi

\bibitem[Agustsson and Timofte(2017)]{agustsson2017ntire}
Eirikur Agustsson and Radu Timofte.
\newblock Ntire 2017 challenge on single image super-resolution: Dataset and study.
\newblock In \emph{Proceedings of the IEEE conference on computer vision and pattern recognition workshops}, pages 126--135, 2017.

\bibitem[Caron et~al.(2021)Caron, Touvron, Misra, J{\'e}gou, Mairal, Bojanowski, and Joulin]{caron2021emerging}
Mathilde Caron, Hugo Touvron, Ishan Misra, Herv{\'e} J{\'e}gou, Julien Mairal, Piotr Bojanowski, and Armand Joulin.
\newblock Emerging properties in self-supervised vision transformers.
\newblock In \emph{Proceedings of the IEEE/CVF international conference on computer vision}, pages 9650--9660, 2021.

\bibitem[Deng et~al.(2019)Deng, Guo, Xue, and Zafeiriou]{deng2019arcface}
Jiankang Deng, Jia Guo, Niannan Xue, and Stefanos Zafeiriou.
\newblock Arcface: Additive angular margin loss for deep face recognition.
\newblock In \emph{Proceedings of the IEEE/CVF conference on computer vision and pattern recognition}, pages 4690--4699, 2019.

\bibitem[Dhariwal and Nichol(2021)]{dhariwal2021diffusion}
Prafulla Dhariwal and Alexander Nichol.
\newblock Diffusion models beat gans on image synthesis.
\newblock \emph{Advances in neural information processing systems}, 34:\penalty0 8780--8794, 2021.

\bibitem[Hanocka et~al.(2020)Hanocka, Metzer, Giryes, and Cohen-Or]{hanocka2020point2mesh}
Rana Hanocka, Gal Metzer, Raja Giryes, and Daniel Cohen-Or.
\newblock Point2mesh: A self-prior for deformable meshes.
\newblock \emph{arXiv preprint arXiv:2005.11084}, 2020.

\bibitem[He et~al.(2016)He, Zhang, Ren, and Sun]{he2016deep}
Kaiming He, Xiangyu Zhang, Shaoqing Ren, and Jian Sun.
\newblock Deep residual learning for image recognition.
\newblock In \emph{Proceedings of the IEEE conference on computer vision and pattern recognition}, pages 770--778, 2016.

\bibitem[Hertz et~al.(2022)Hertz, Mokady, Tenenbaum, Aberman, Pritch, and Cohen-Or]{hertz2022prompt}
Amir Hertz, Ron Mokady, Jay Tenenbaum, Kfir Aberman, Yael Pritch, and Daniel Cohen-Or.
\newblock Prompt-to-prompt image editing with cross attention control.
\newblock \emph{arXiv preprint arXiv:2208.01626}, 2022.

\bibitem[Hertz et~al.(2023)Hertz, Aberman, and Cohen-Or]{hertz2023delta}
Amir Hertz, Kfir Aberman, and Daniel Cohen-Or.
\newblock Delta denoising score.
\newblock In \emph{Proceedings of the IEEE/CVF International Conference on Computer Vision}, pages 2328--2337, 2023.

\bibitem[Ho and Salimans(2022)]{ho2022classifier}
Jonathan Ho and Tim Salimans.
\newblock Classifier-free diffusion guidance.
\newblock \emph{arXiv preprint arXiv:2207.12598}, 2022.

\bibitem[Ho et~al.(2020)Ho, Jain, and Abbeel]{ho2020denoising}
Jonathan Ho, Ajay Jain, and Pieter Abbeel.
\newblock Denoising diffusion probabilistic models.
\newblock \emph{Advances in neural information processing systems}, 33:\penalty0 6840--6851, 2020.

\bibitem[Huang et~al.(2024)Huang, Wen, Wang, Ren, and Jia]{huang2024surface}
Zhangjin Huang, Yuxin Wen, Zihao Wang, Jinjuan Ren, and Kui Jia.
\newblock Surface reconstruction from point clouds: A survey and a benchmark.
\newblock \emph{IEEE transactions on pattern analysis and machine intelligence}, 2024.

\bibitem[Kim et~al.(2025)Kim, Kim, Park, Ahn, Kang, Kim, Jin, and Cha]{kim2025identity}
SeonHwa Kim, Jiwon Kim, Soobin Park, Donghoon Ahn, Jiwon Kang, Seungryong Kim, Kyong~Hwan Jin, and Eunju Cha.
\newblock Identity-preserving distillation sampling by fixed-point iterator.
\newblock \emph{arXiv preprint arXiv:2502.19930}, 2025.

\bibitem[Kirillov et~al.(2023)Kirillov, Mintun, Ravi, Mao, Rolland, Gustafson, Xiao, Whitehead, Berg, Lo, et~al.]{kirillov2023segment}
Alexander Kirillov, Eric Mintun, Nikhila Ravi, Hanzi Mao, Chloe Rolland, Laura Gustafson, Tete Xiao, Spencer Whitehead, Alexander~C Berg, Wan-Yen Lo, et~al.
\newblock Segment anything.
\newblock In \emph{Proceedings of the IEEE/CVF international conference on computer vision}, pages 4015--4026, 2023.

\bibitem[Lin et~al.(2025)Lin, Mo, Klingher, Mu, and Zhou]{lin2025ctrl}
Kuan~Heng Lin, Sicheng Mo, Ben Klingher, Fangzhou Mu, and Bolei Zhou.
\newblock Ctrl-x: Controlling structure and appearance for text-to-image generation without guidance.
\newblock \emph{Advances in Neural Information Processing Systems}, 37:\penalty0 128911--128939, 2025.

\bibitem[Lu et~al.(2022)Lu, Zhou, Bao, Chen, Li, and Zhu]{lu2022dpm}
Cheng Lu, Yuhao Zhou, Fan Bao, Jianfei Chen, Chongxuan Li, and Jun Zhu.
\newblock Dpm-solver: A fast ode solver for diffusion probabilistic model sampling in around 10 steps.
\newblock \emph{Advances in neural information processing systems}, 35:\penalty0 5775--5787, 2022.

\bibitem[Lu et~al.(2025)Lu, Zhou, Bao, Chen, Li, and Zhu]{lu2025dpm}
Cheng Lu, Yuhao Zhou, Fan Bao, Jianfei Chen, Chongxuan Li, and Jun Zhu.
\newblock Dpm-solver++: Fast solver for guided sampling of diffusion probabilistic models.
\newblock \emph{Machine Intelligence Research}, pages 1--22, 2025.

\bibitem[Mo et~al.(2024)Mo, Mu, Lin, Liu, Guan, Li, and Zhou]{mo2024freecontrol}
Sicheng Mo, Fangzhou Mu, Kuan~Heng Lin, Yanli Liu, Bochen Guan, Yin Li, and Bolei Zhou.
\newblock Freecontrol: Training-free spatial control of any text-to-image diffusion model with any condition.
\newblock In \emph{Proceedings of the IEEE/CVF Conference on Computer Vision and Pattern Recognition}, pages 7465--7475, 2024.

\bibitem[Mou et~al.(2024)Mou, Wang, Xie, Wu, Zhang, Qi, and Shan]{mou2024t2i}
Chong Mou, Xintao Wang, Liangbin Xie, Yanze Wu, Jian Zhang, Zhongang Qi, and Ying Shan.
\newblock T2i-adapter: Learning adapters to dig out more controllable ability for text-to-image diffusion models.
\newblock In \emph{Proceedings of the AAAI conference on artificial intelligence}, pages 4296--4304, 2024.

\bibitem[Nam et~al.(2024)Nam, Kwon, Park, and Ye]{nam2024contrastive}
Hyelin Nam, Gihyun Kwon, Geon~Yeong Park, and Jong~Chul Ye.
\newblock Contrastive denoising score for text-guided latent diffusion image editing.
\newblock In \emph{Proceedings of the IEEE/CVF conference on computer vision and pattern recognition}, pages 9192--9201, 2024.

\bibitem[Podell et~al.(2023)Podell, English, Lacey, Blattmann, Dockhorn, M{\"u}ller, Penna, and Rombach]{podell2023sdxl}
Dustin Podell, Zion English, Kyle Lacey, Andreas Blattmann, Tim Dockhorn, Jonas M{\"u}ller, Joe Penna, and Robin Rombach.
\newblock Sdxl: Improving latent diffusion models for high-resolution image synthesis.
\newblock \emph{arXiv preprint arXiv:2307.01952}, 2023.

\bibitem[Poole et~al.(2022)Poole, Jain, Barron, and Mildenhall]{poole2022dreamfusion}
Ben Poole, Ajay Jain, Jonathan~T Barron, and Ben Mildenhall.
\newblock Dreamfusion: Text-to-3d using 2d diffusion.
\newblock \emph{arXiv preprint arXiv:2209.14988}, 2022.

\bibitem[Qin et~al.(2023)Qin, Zhang, Yu, Feng, Yang, Zhou, Wang, Niebles, Xiong, Savarese, et~al.]{qin2023unicontrol}
Can Qin, Shu Zhang, Ning Yu, Yihao Feng, Xinyi Yang, Yingbo Zhou, Huan Wang, Juan~Carlos Niebles, Caiming Xiong, Silvio Savarese, et~al.
\newblock {Unicontrol: A unified diffusion model for controllable visual generation in the wild}.
\newblock \emph{arXiv preprint arXiv:2305.11147}, 2023.

\bibitem[Radford et~al.(2021)Radford, Kim, Hallacy, Ramesh, Goh, Agarwal, Sastry, Askell, Mishkin, Clark, et~al.]{radford2021learning}
Alec Radford, Jong~Wook Kim, Chris Hallacy, Aditya Ramesh, Gabriel Goh, Sandhini Agarwal, Girish Sastry, Amanda Askell, Pamela Mishkin, Jack Clark, et~al.
\newblock Learning transferable visual models from natural language supervision.
\newblock In \emph{International conference on machine learning}, pages 8748--8763. PmLR, 2021.

\bibitem[Rombach et~al.(2022)Rombach, Blattmann, Lorenz, Esser, and Ommer]{rombach2022high}
Robin Rombach, Andreas Blattmann, Dominik Lorenz, Patrick Esser, and Bj{\"o}rn Ommer.
\newblock High-resolution image synthesis with latent diffusion models.
\newblock In \emph{Proceedings of the IEEE/CVF conference on computer vision and pattern recognition}, pages 10684--10695, 2022.

\bibitem[Ruiz et~al.(2023)Ruiz, Li, Jampani, Pritch, Rubinstein, and Aberman]{ruiz2023dreambooth}
Nataniel Ruiz, Yuanzhen Li, Varun Jampani, Yael Pritch, Michael Rubinstein, and Kfir Aberman.
\newblock Dreambooth: Fine tuning text-to-image diffusion models for subject-driven generation.
\newblock In \emph{Proceedings of the IEEE/CVF conference on computer vision and pattern recognition}, pages 22500--22510, 2023.

\bibitem[Schuhmann et~al.(2022)Schuhmann, Beaumont, Vencu, Gordon, Wightman, Cherti, Coombes, Katta, Mullis, Wortsman, et~al.]{schuhmann2022laion}
Christoph Schuhmann, Romain Beaumont, Richard Vencu, Cade Gordon, Ross Wightman, Mehdi Cherti, Theo Coombes, Aarush Katta, Clayton Mullis, Mitchell Wortsman, et~al.
\newblock Laion-5b: An open large-scale dataset for training next generation image-text models.
\newblock \emph{Advances in neural information processing systems}, 35:\penalty0 25278--25294, 2022.

\bibitem[Song et~al.(2020{\natexlab{a}})Song, Meng, and Ermon]{song2020denoising}
Jiaming Song, Chenlin Meng, and Stefano Ermon.
\newblock Denoising diffusion implicit models.
\newblock \emph{arXiv preprint arXiv:2010.02502}, 2020{\natexlab{a}}.

\bibitem[Song et~al.(2020{\natexlab{b}})Song, Sohl-Dickstein, Kingma, Kumar, Ermon, and Poole]{song2020score}
Yang Song, Jascha Sohl-Dickstein, Diederik~P Kingma, Abhishek Kumar, Stefano Ermon, and Ben Poole.
\newblock Score-based generative modeling through stochastic differential equations.
\newblock \emph{arXiv preprint arXiv:2011.13456}, 2020{\natexlab{b}}.

\bibitem[Tumanyan et~al.(2022)Tumanyan, Bar-Tal, Bagon, and Dekel]{tumanyan2022splicing}
Narek Tumanyan, Omer Bar-Tal, Shai Bagon, and Tali Dekel.
\newblock Splicing vit features for semantic appearance transfer.
\newblock In \emph{Proceedings of the IEEE/CVF Conference on Computer Vision and Pattern Recognition}, pages 10748--10757, 2022.

\bibitem[Tumanyan et~al.(2023)Tumanyan, Geyer, Bagon, and Dekel]{tumanyan2023plug}
Narek Tumanyan, Michal Geyer, Shai Bagon, and Tali Dekel.
\newblock Plug-and-play diffusion features for text-driven image-to-image translation.
\newblock In \emph{Proceedings of the IEEE/CVF Conference on Computer Vision and Pattern Recognition}, pages 1921--1930, 2023.

\bibitem[Ye et~al.(2023)Ye, Zhang, Liu, Han, and Yang]{ye2023ip}
Hu Ye, Jun Zhang, Sibo Liu, Xiao Han, and Wei Yang.
\newblock Ip-adapter: Text compatible image prompt adapter for text-to-image diffusion models.
\newblock \emph{arXiv preprint arXiv:2308.06721}, 2023.

\bibitem[Zhang et~al.(2023)Zhang, Rao, and Agrawala]{zhang2023adding}
Lvmin Zhang, Anyi Rao, and Maneesh Agrawala.
\newblock Adding conditional control to text-to-image diffusion models.
\newblock In \emph{Proceedings of the IEEE/CVF international conference on computer vision}, pages 3836--3847, 2023.

\bibitem[Zhao et~al.(2023)Zhao, Bai, Rao, Zhou, and Lu]{zhao2023unipc}
Wenliang Zhao, Lujia Bai, Yongming Rao, Jie Zhou, and Jiwen Lu.
\newblock Unipc: A unified predictor-corrector framework for fast sampling of diffusion models.
\newblock \emph{Advances in Neural Information Processing Systems}, 36:\penalty0 49842--49869, 2023.

\bibitem[Zhou and Jacobson(2016)]{zhou2016thingi10k}
Qingnan Zhou and Alec Jacobson.
\newblock Thingi10k: A dataset of 10,000 3d-printing models.
\newblock \emph{arXiv preprint arXiv:1605.04797}, 2016.

\end{thebibliography}
}

\clearpage
\setcounter{equation}{1}
\setcounter{page}{1}

\setcounter{table}{0}
\renewcommand{\thetable}{S\arabic{table}}

\setcounter{figure}{0}
\renewcommand{\thefigure}{S\arabic{figure}}

\setcounter{section}{0}
\renewcommand{\thesection}{S\arabic{section}}
\renewcommand{\theHsection}{S\arabic{section}}

\twocolumn[{
\renewcommand\twocolumn[1][]{#1}%
\maketitlesupplementary
}]

\normalsize
\section{Introduction}
This supplementary material is intended to support the main paper. We provide comprehensive ablation studies that substantiate the chosen hyper‑parameters and methodological decisions underpinning the Dual Recursive Feedback framework. Moreover, we report additional experiments with complementary evaluation metrics that rigorously quantify the contribution of each feedback mechanisms.

\section{Ablation studies on DRF}
\begin{figure}[H] 
\footnotesize
\centering 

\newcommand{\imgwidth}{0.18\linewidth} 

\begin{tikzpicture}[x=1cm, y=1cm]
    \node[anchor=south] (FigA1) at (0,0) {
        \includegraphics[width=\imgwidth]{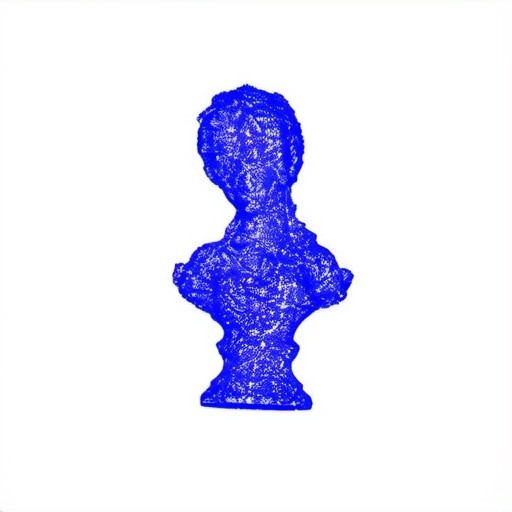}
    };
    \node[anchor=south, yshift=-1mm] at (FigA1.north) {\footnotesize Structure};
\end{tikzpicture}\hspace{-1mm}%
\begin{tikzpicture}[x=1cm, y=1cm]
    \node[anchor=south] (FigD1) at (0,0) {
        \includegraphics[width=\imgwidth]{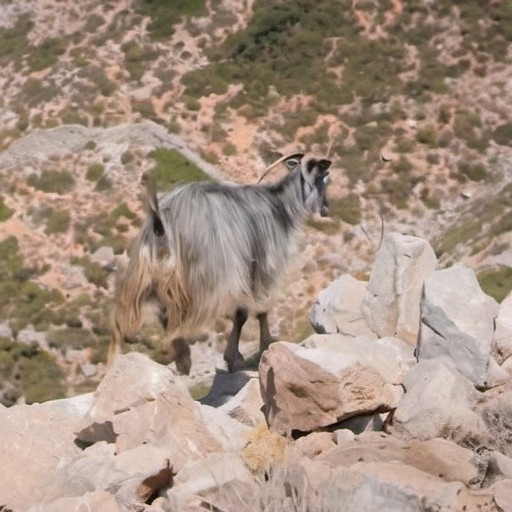}
    };
    \node[anchor=south, yshift=-1mm] at (FigD1.north) {\footnotesize Appearance};
\end{tikzpicture}\hspace{-1mm}%
\begin{tikzpicture}[x=1cm, y=1cm]
    \node[anchor=south] (FigC1) at (0,0) {
        \includegraphics[width=\imgwidth]{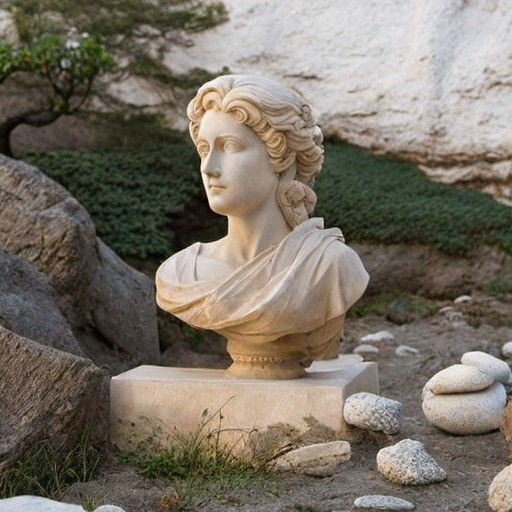}
    };
    \node[anchor=south, yshift=-1mm] at (FigC1.north) {\footnotesize 10 steps};
\end{tikzpicture}\hspace{-1mm}%
\begin{tikzpicture}[x=1cm, y=1cm]
    \node[anchor=south] (FigB1) at (0,0) {
        \includegraphics[width=\imgwidth]{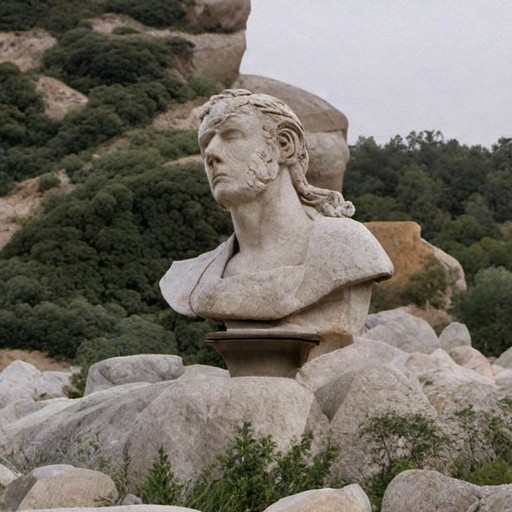}
    };
    \node[anchor=south, yshift=-1mm] at (FigB1.north) {\footnotesize \textbf{20 steps}};
\end{tikzpicture}\hspace{-1mm}%
\begin{tikzpicture}[x=1cm, y=1cm]
    \node[anchor=south] (FigE1) at (0,0) { 
        \includegraphics[width=\imgwidth]{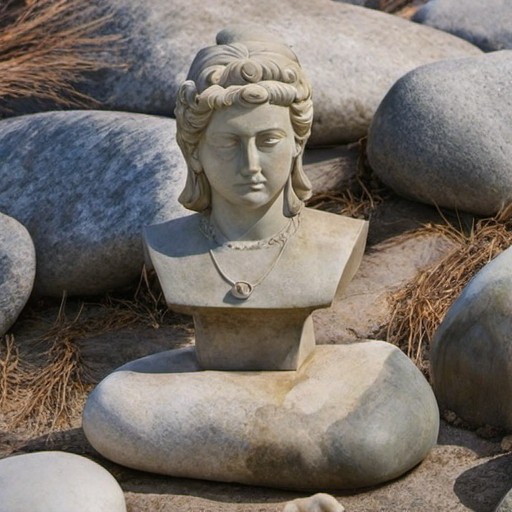}
    };
    \node[anchor=south, yshift=-1mm] at (FigE1.north) {\footnotesize 50 steps}; 
\end{tikzpicture}
\\[-4pt]
\vspace{0pt}
\setulcolor{Dandelion}
\setul{0.2pt}{1.5pt}
\centering \textit{``A photo of \ul{a statue} on stones"} 
\vspace{-5pt}
\caption{\textbf{Ablation study on the number of steps.} DRF is applied during intermediate 20 steps after the first five steps.}
\vspace{-10pt}
\label{fig:ablation_steps}
\end{figure}

\phantomsection
\label{sec.sup.ablation}
\noindent\textbf{Steps for DRF.}\quad
To balance both efficiency and generated image fidelity, we apply DRF during a subset of the total inference steps (50 steps). We compare the scenarios of applying DRF for 10 steps, 20 steps (ours), and all 50 steps, with results summarized in \cref{fig:ablation_steps}. 
\begin{itemize}
    \item 10 steps: While inference time (36.09s) is reduced, certain structural details (e.g., the statue’s head orientation) are not sufficiently captured.
    \item All 50 steps: It takes 135.07s but does not yield substantially better outcomes than the 20 steps setting.
    \item \textbf{20 steps (Ours)}: Although slightly longer (56.87s), it preserved fine-grained structure effectively.
\end{itemize}
\vspace{-6pt}

\begin{figure}[htbp]
  \centering
  \includegraphics[width=0.8\linewidth]{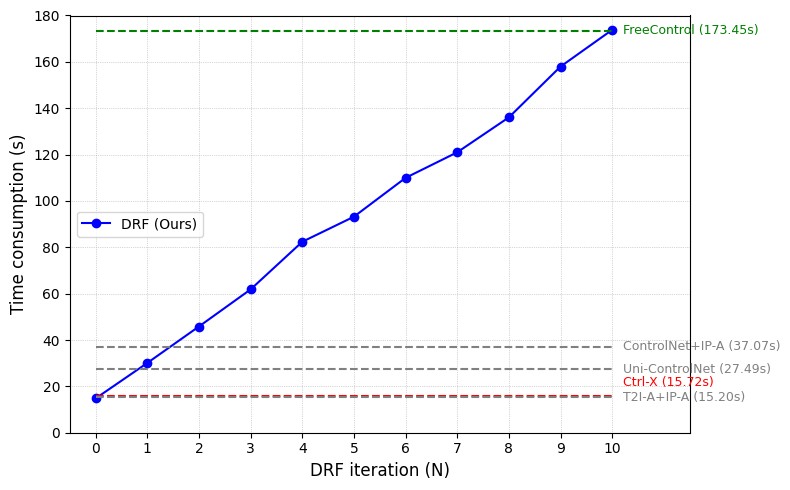} 
  \caption{\textbf{DRF iteration cost.} Inference time rises roughly linearly with the number of DRF passes (blue), staying below FreeControl and matching faster baselines at low iteration counts.}
  \vspace{-15pt}
  \label{fig:time_graph}
\end{figure}

\noindent\textbf{Number of DRF iterations.}\quad
We investigate how many times DRF should be invoked at each timestep within its recurrence loop to maximize its impact. As illustrated in \cref{fig:ablation_drf_iter}, increasing the number of DRF applications at each timestep more robustly transforms the appearance into the intended structural form. This confirms that elevating the iteration count can further strengthen the alignment between the structure and the generated image. Furthermore, we analyze the cost of these gains by measuring time as a function of the DRF iteration count, as summarized in \cref{fig:time_graph}. Inference latency grows almost linearly with additional feedback passes, allowing practitioners to trade fidelity for speed by selecting an iteration budget that fits their deployment constraints.

\usetikzlibrary{calc, positioning}

\begin{figure}[ht]
\centering
\footnotesize

\newcommand{\imgsize}{1.7cm}

\begin{tikzpicture}[>=latex]

\node[anchor=south] (headerFree) at (4.0,1.2) {\textbf{\textcolor{black}{DRF iteration}}};

\draw[very thick, gray, opacity=0.5] (1.3,1.2) -- (6.65,1.2);

\node (structure) at (0.3,0)
{\includegraphics[width=\imgsize]{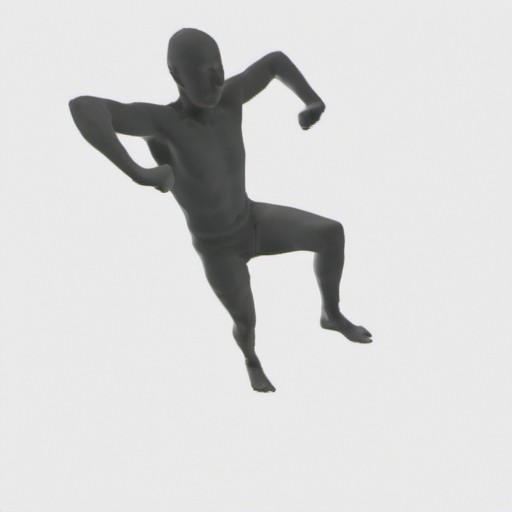}};
\node[above=-3.5pt of structure] {\textbf{Structure}};

\node (drf0) at (2.2,0)
{\includegraphics[width=\imgsize]{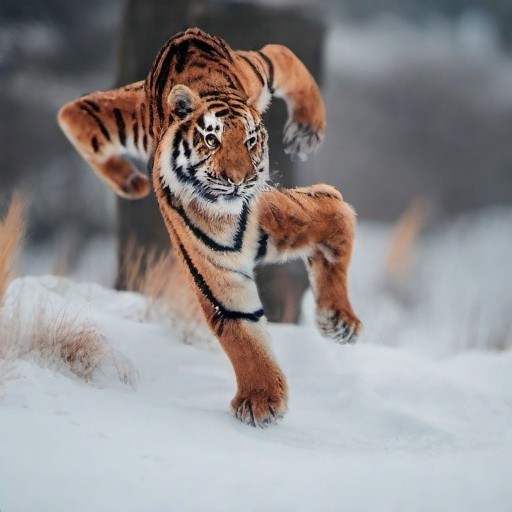}};
\node[above=-3.5pt of drf0] {\textbf{1}};

\node (drf1) at (4.0,0)
{\includegraphics[width=\imgsize]{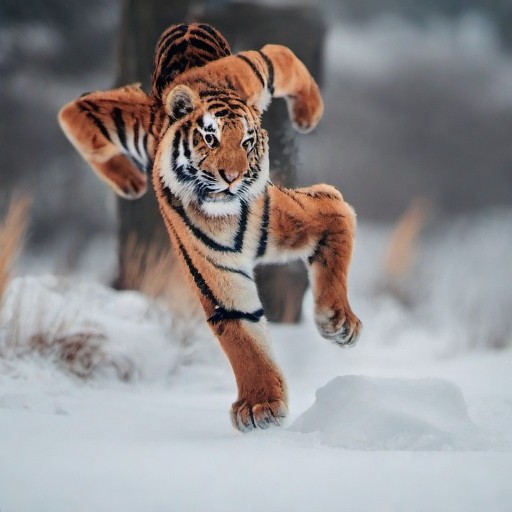}};
\node[above=-3.5pt of drf1] {\textbf{2}};

\node (drf2) at (5.8,0)
{\includegraphics[width=\imgsize]{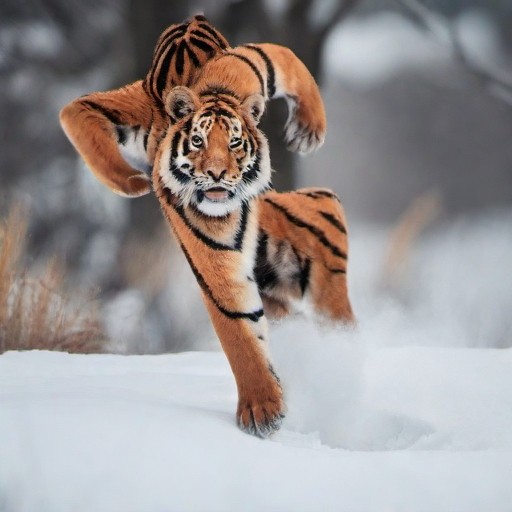}};
\node[above=-3.5pt of drf2] {\textbf{3}};
\vspace{-5pt}
\node (appearance) at (0.3,-2.1)
{\includegraphics[width=\imgsize]{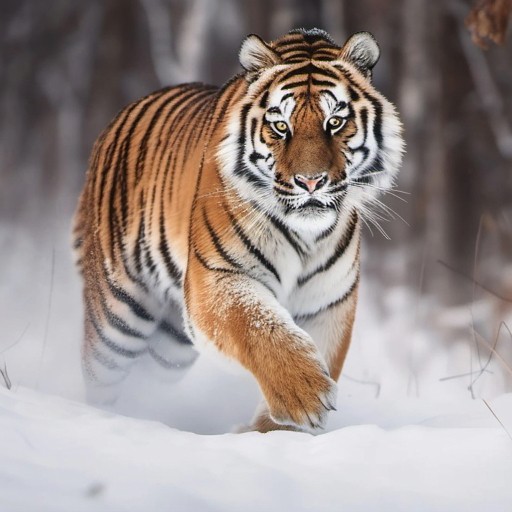}};
\node[above=-3.5pt of appearance] {\textbf{Appearance}};

\node (drf3) at (2.2,-2.1)
{\includegraphics[width=\imgsize]{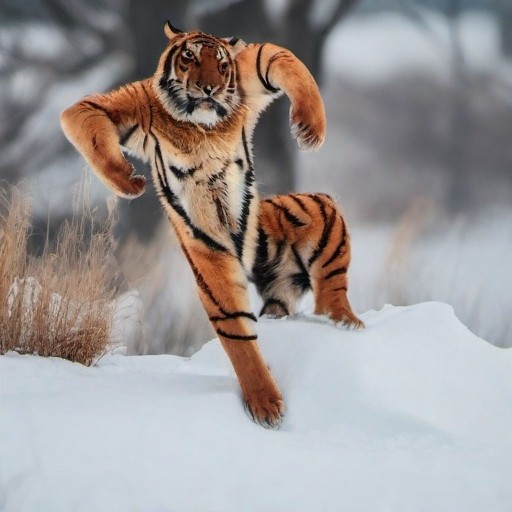}};
\node[above=-3.5pt of drf3] {\textbf{4}};

\node (drf4) at (4.0,-2.1)
{\includegraphics[width=\imgsize]{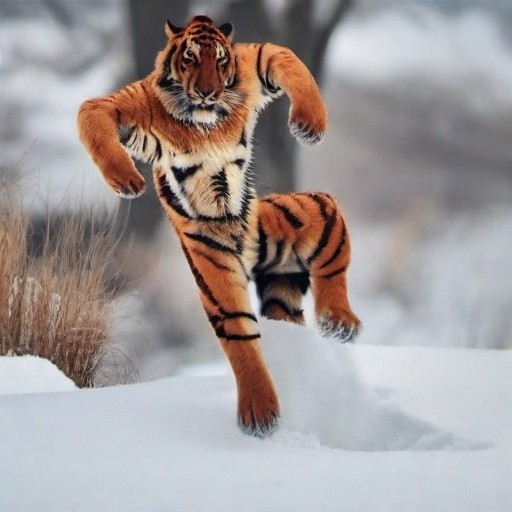}};
\node[above=-3.5pt of drf4] {\textbf{5}};

\node (drf5) at (5.8,-2.1)
{\includegraphics[width=\imgsize]{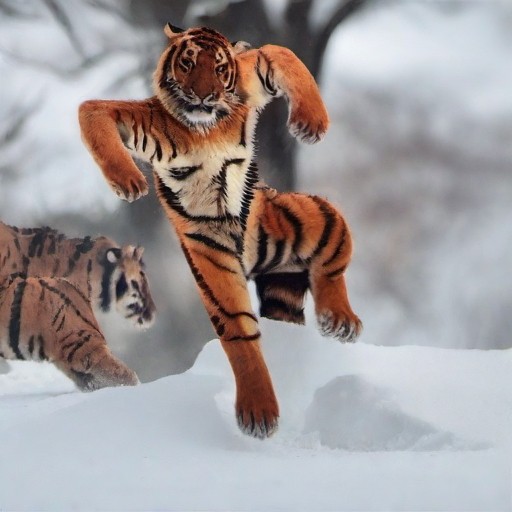}};
\node[above=-3.5pt of drf5] {\textbf{6}};

\draw[dotted, thick]
 (1.262, 0.88) -- (1.262, -2.95);

\end{tikzpicture}
\vspace{-8pt}
\caption{\textbf{Ablation study on the number of iterations for DRF.} The fusion of appearance and structure image is formed by iteration number of DRF.}
\label{fig:ablation_drf_iter}
\end{figure}

\noindent\textbf{Hyper-parameters of DRF.}\quad
We performed a parameter sweep over $\lambda$, $\rho$, and k to validate our chosen settings. Across all three metrics, CLIP, Self-Sim, and DINO-I, each hyper-parameter exhibits a clear, monotonic sweet spot shown in \cref{fig:time_graph}).
Increasing the update weight $\lambda$ mildly affects CLIP and Self-Sim but raises DINO-I, peaking at $\lambda{=}1.0$.
For the feedback balance $\rho$, smaller values favor appearance consistency; $\rho{=}0.001$ delivers the lowest Self-Sim and the highest DINO-I without degrading CLIP.
Finally, the recursion depth $K$ provides diminishing returns: quality improves up to $K{=}5$ and plateaus thereafter, while further iterations incur extra latency. Accordingly, we adopt $\lambda,\rho,K=(1.0,,0.001,,5)$ in all subsequent experiments.

\setlength{\abovecaptionskip}{0pt}
\setlength{\belowcaptionskip}{-4pt}
\vspace{-3pt}
\begin{figure}[h]
    \centering
    \begin{subfigure}[t]{0.32\linewidth}
        \includegraphics[width=\linewidth]{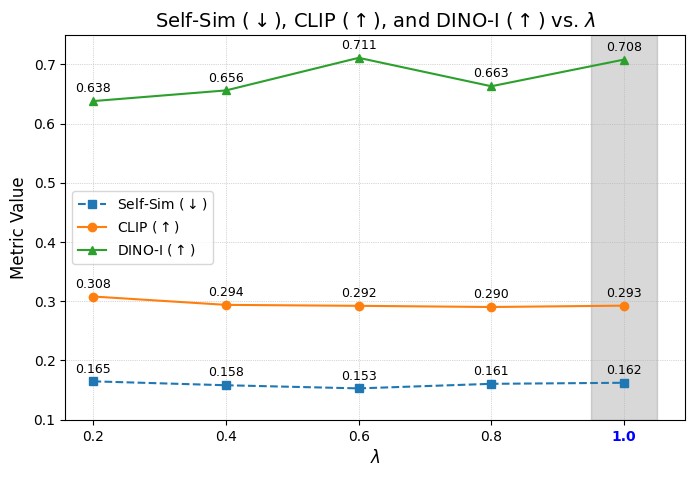}
    \end{subfigure}
    \hspace{-5pt}
    \begin{subfigure}[t]{0.32\linewidth}
        \includegraphics[width=\linewidth]{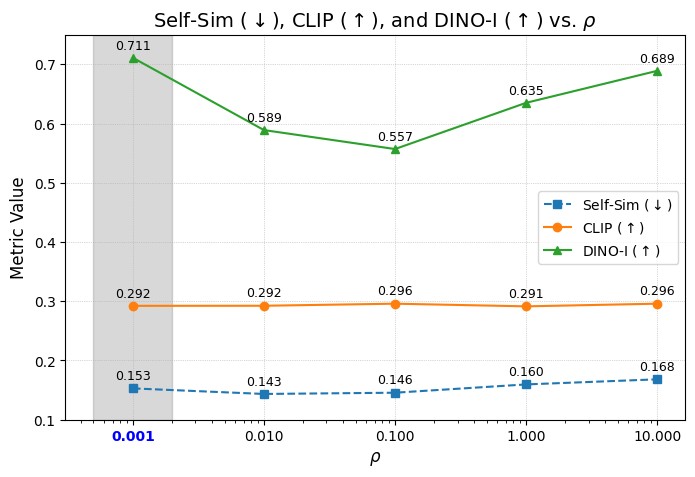}
    \end{subfigure}
    \hspace{-5pt}
    \begin{subfigure}[t]{0.32\linewidth}
        \includegraphics[width=\linewidth]{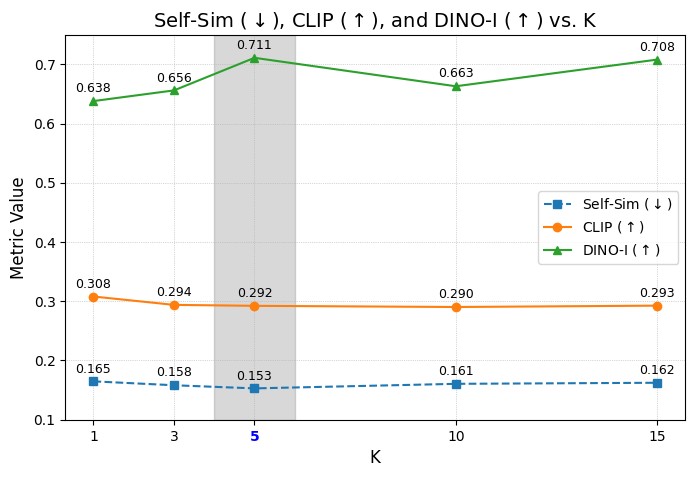}
    \end{subfigure}
    \caption{\textbf{Hyper-parameters of DRF.}}
\end{figure}

\begin{figure}[H] 
\footnotesize
\centering 

\newcommand{\imgwidth}{0.18\linewidth} 

\begin{tikzpicture}[x=1cm, y=1cm]
    \node[anchor=south] (FigA1) at (0,0) {
        \includegraphics[width=\imgwidth]{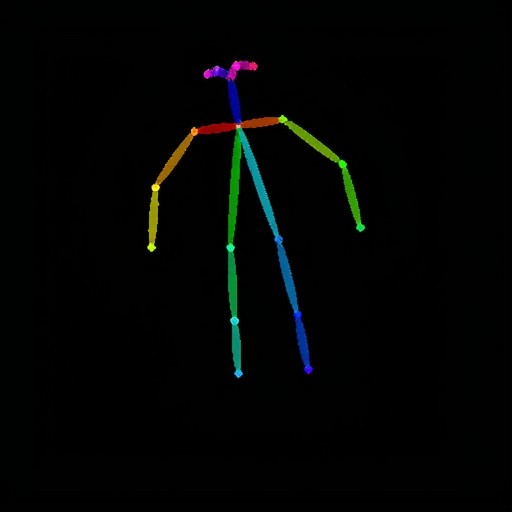}
    };
    \node[anchor=south, yshift=-1mm] at (FigA1.north) {\scriptsize Structure};
\end{tikzpicture}\hspace{-1mm}%
\begin{tikzpicture}[x=1cm, y=1cm]
    \node[anchor=south] (FigD1) at (0,0) {
        \includegraphics[width=\imgwidth]{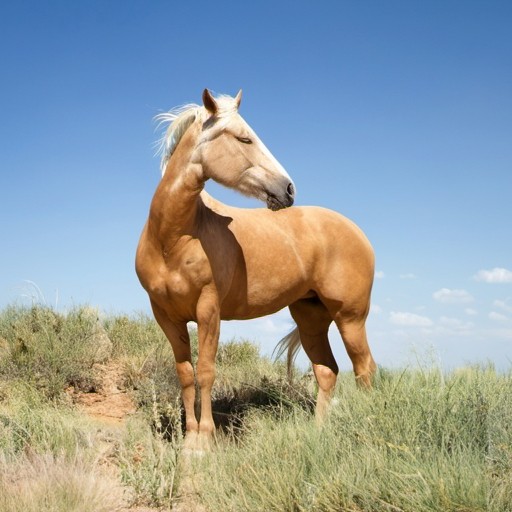}
    };
    \node[anchor=south, yshift=-1mm] at (FigD1.north) {\scriptsize Appearance};
\end{tikzpicture}\hspace{-1mm}%
\begin{tikzpicture}[x=1cm, y=1cm]
    \node[anchor=south] (FigC1) at (0,0) {
        \includegraphics[width=\imgwidth]{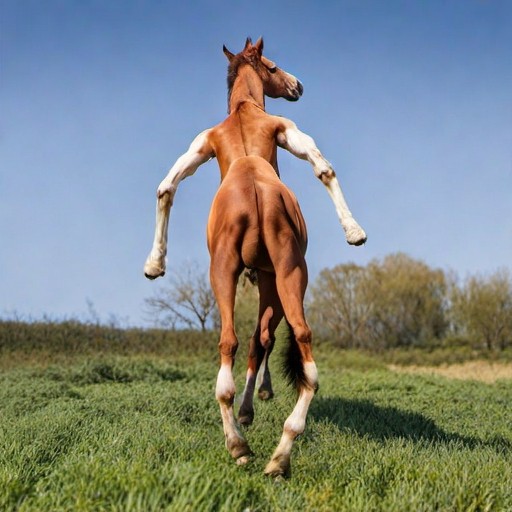}
    };
    \node[anchor=south, yshift=-1mm] at (FigC1.north) {\scriptsize \textbf{Exponential}};
\end{tikzpicture}\hspace{-1mm}%
\begin{tikzpicture}[x=1cm, y=1cm]
    \node[anchor=south] (FigB1) at (0,0) {
        \includegraphics[width=\imgwidth]{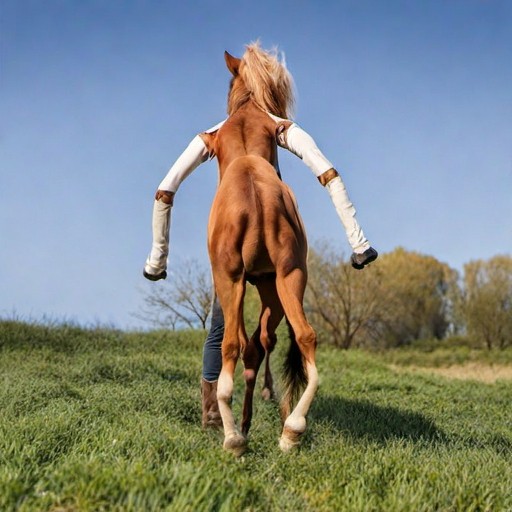}
    };
    \node[anchor=south, yshift=-1mm] at (FigB1.north) {\scriptsize Cosine};
\end{tikzpicture}\hspace{-1mm}%
\begin{tikzpicture}[x=1cm, y=1cm]
    \node[anchor=south] (FigE1) at (0,0) { 
        \includegraphics[width=\imgwidth]{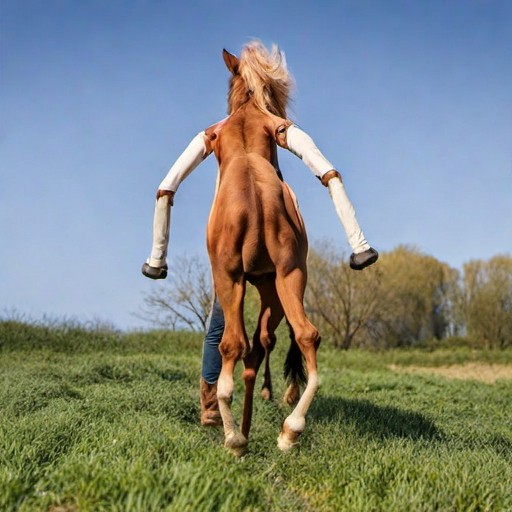}
    };
    \node[anchor=south, yshift=-1mm] at (FigE1.north) {\scriptsize Linear}; 
\end{tikzpicture}

\vspace{0pt}
\caption{\textbf{Comparison of weight method of DRF.} Exponential weight schedule determines the optimized weight for Generation feedback.}
\vspace{-5pt}
\label{fig:weight}
\end{figure}

\noindent\textbf{Weight for DRF loss.}\quad
Building on the theoretical analysis in \cref{sec:experiments}, we empirically verify that amplifying the generation feedback term as the recursive feedback iteration index $i$ grows is crucial for harmonising appearance and structure features. Specifically, the exponential schedule of \cref{eq:weight} progressively shifts the optimisation focus from low-level appearance injection in early iterations to high-level structural refinement in later ones. As shown in \cref{fig:weight}, this strategy consistently yields sharper details and more faithful pose alignment than linear or uniform weighting, confirming that a larger weight on generation feedback at higher recursion depths is the most effective way to fuse appearance and structural constraints in the final image.

\section{Additional Experiments}
To more rigorously confirm that DRF preserves both appearance and structure, we add two complementary metrics. ArcFace~\cite{deng2019arcface} similarity quantifies identity retention by measuring the cosine distance between face embeddings of the generated image and the appearance reference, while SAM~\cite{kirillov2023segment}-IoU assesses pose fidelity by comparing structure-aware mask segmentations of the structure reference with those of each synthesis. As reported in \cref{tab:additional_experiments}, DRF achieves the highest ArcFace similarity, demonstrating superior identity preservation alongside accurate pose alignment. While DRF matches Ctrl-X in IoU, it delivers superior overall image quality.
\noindent 
\begin{table}[ht]
\vspace{-3pt}
\centering
\scriptsize 
\renewcommand{\arraystretch}{0.5}
\setlength{\tabcolsep}{1.9pt}
\begin{tabular}{r@{\hspace{3pt}} c@{\hspace{1pt}} c@{\hspace{1pt}} c@{\hspace{1pt}} c@{\hspace{1pt}} c@{\hspace{1pt}} c }
\toprule
\textbf{} &
\begin{tabular}{@{}c@{}}
     \tiny \textbf{DRF}\\
     \tiny \textbf{(Ours)}
\end{tabular}  & \tiny Ctrl-X & \tiny FreeControl & 
\begin{tabular}{@{}c@{}}
     \tiny Uni-\\
     \tiny ControlNet
\end{tabular} & 
\begin{tabular}{@{}c@{}}
     \tiny ControlNet +\\
     \tiny IP-Adapter
\end{tabular} & 
\begin{tabular}{@{}c@{}}
     \tiny T2I-Adapter +\\
     \tiny IP-Adapter
\end{tabular} \\
\midrule
\textbf{ArcFace $\downarrow$} & \cellcolor{gray!30}\textbf{0.6221} & 0.6497 & 0.7043 & 0.6961 & 0.7089 & 0.6469 \\
\midrule
\vspace{1pt}
\textbf{IoU $\uparrow$ \;} & \textbf{0.8048} & \cellcolor{gray!30}0.8205 & 0.7984 & 0.6983 & 0.7498 & 0.4767 \\
\includegraphics[width=0.07\textwidth]{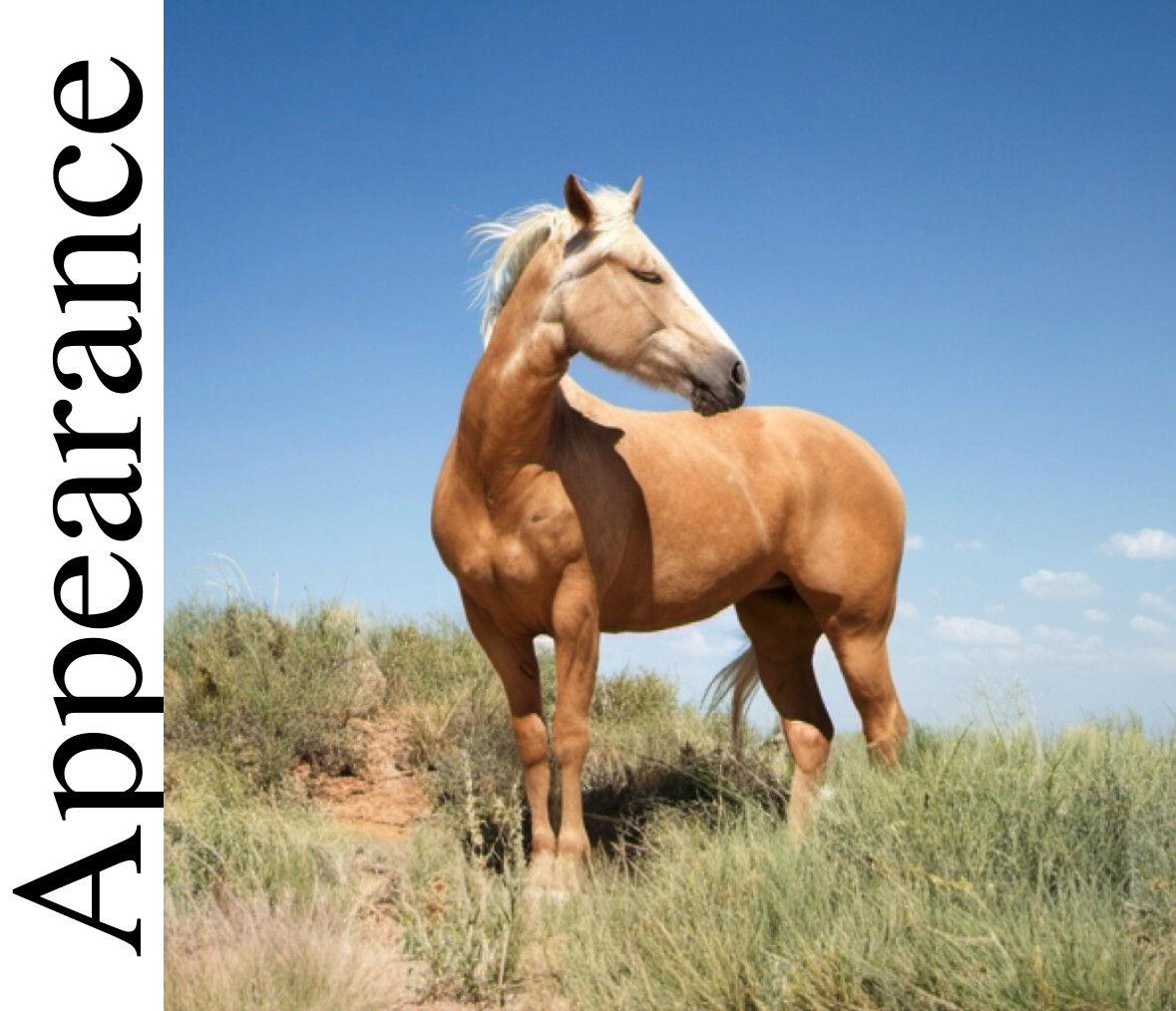} & 
\includegraphics[width=0.061\textwidth]{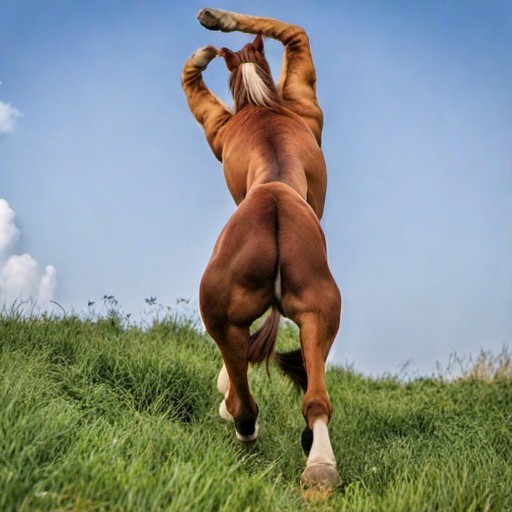} &
\includegraphics[width=0.061\textwidth]{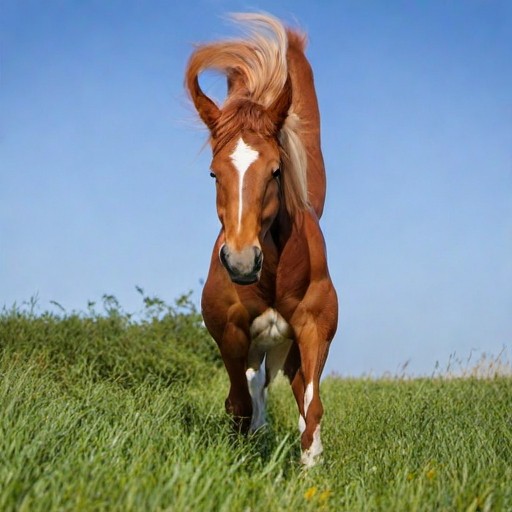} &
\includegraphics[width=0.061\textwidth]{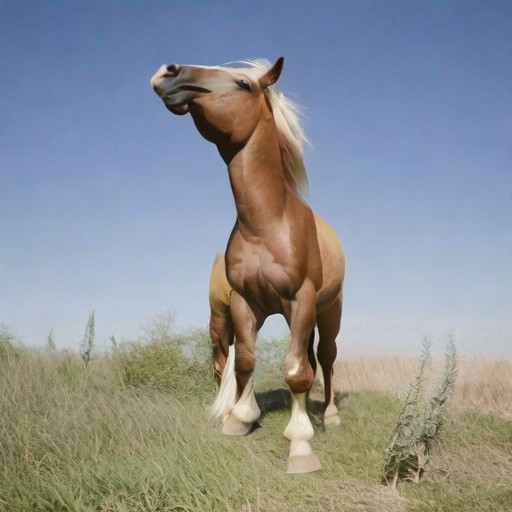} &
\includegraphics[width=0.061\textwidth]{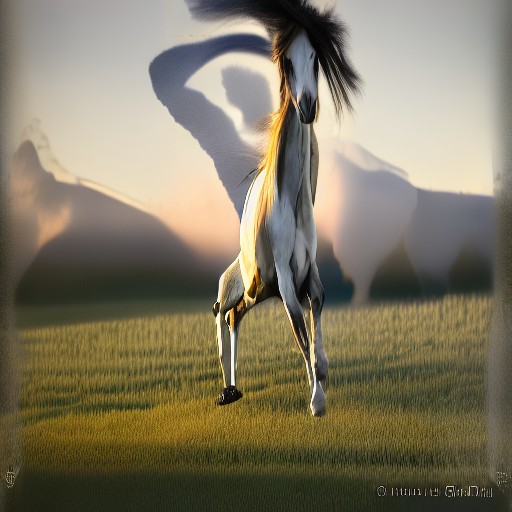} &
\includegraphics[width=0.061\textwidth]{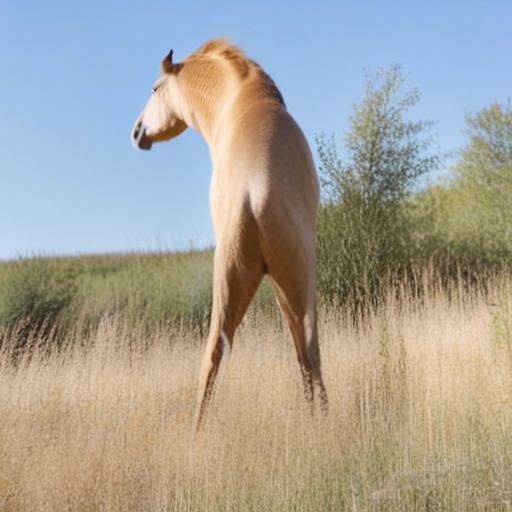} &
\includegraphics[width=0.061\textwidth]{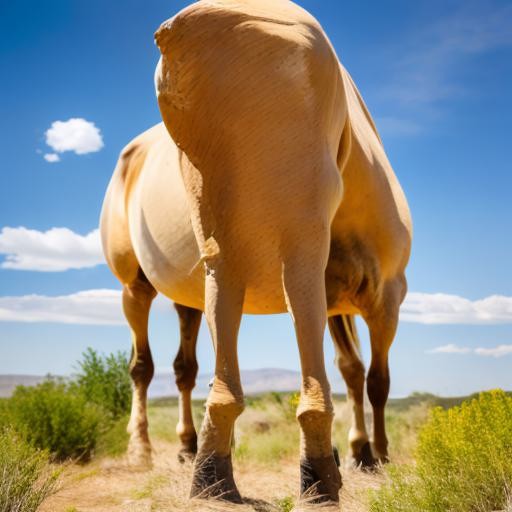} \\
\vspace{-1.5pt}
\includegraphics[width=0.07\textwidth]{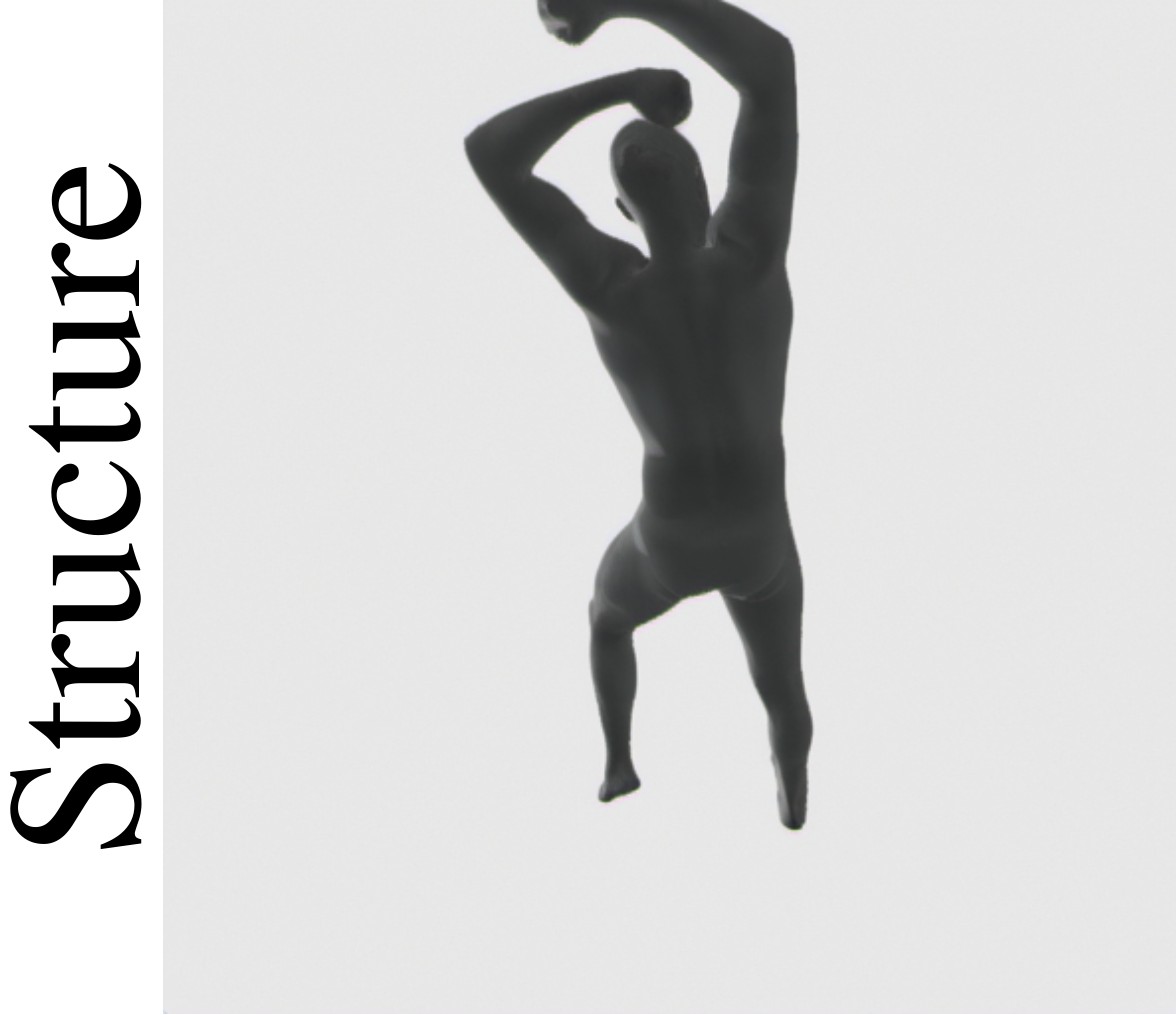} &
\includegraphics[width=0.061\textwidth]{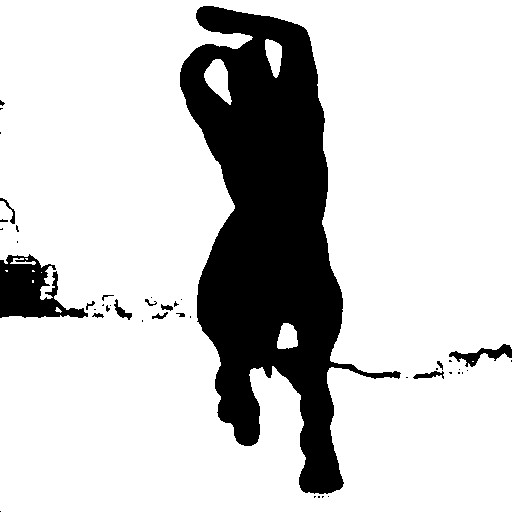} &
\includegraphics[width=0.061\textwidth]{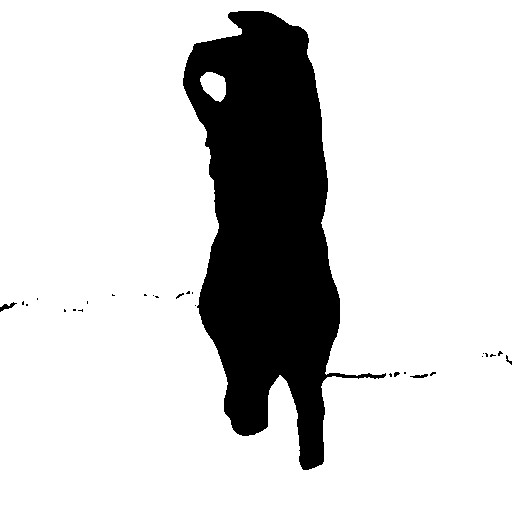} &
\includegraphics[width=0.061\textwidth]{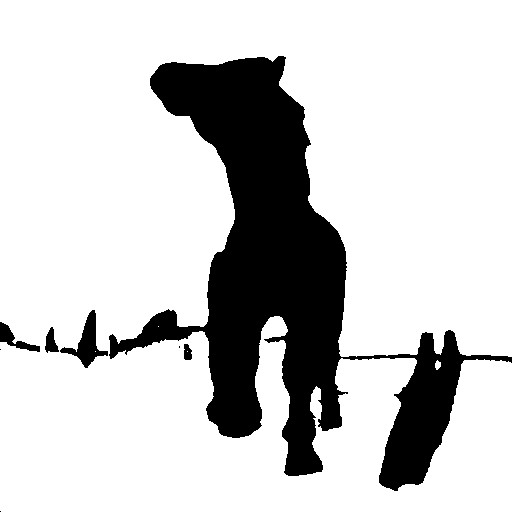} &
\includegraphics[width=0.061\textwidth]{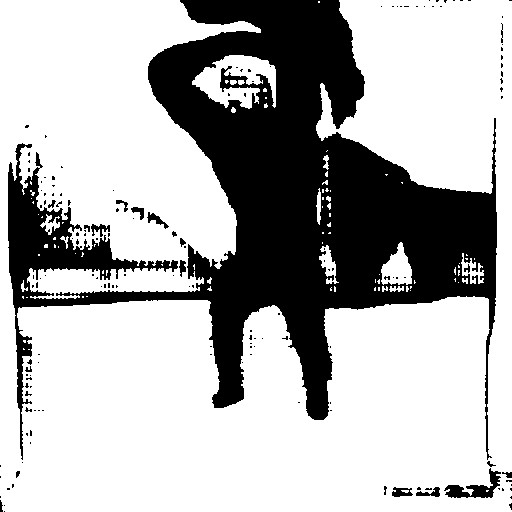} &
\includegraphics[width=0.061\textwidth]{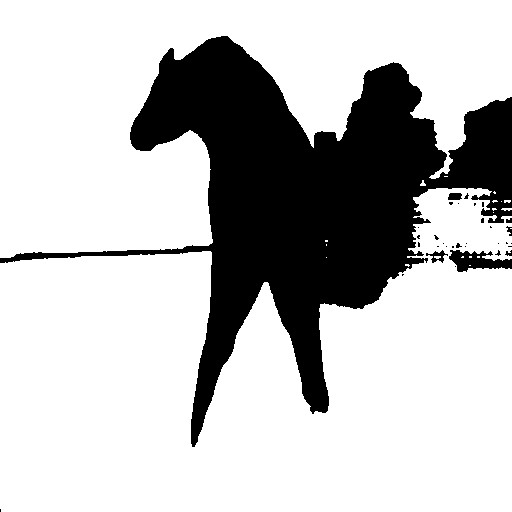} &
\includegraphics[width=0.061\textwidth]{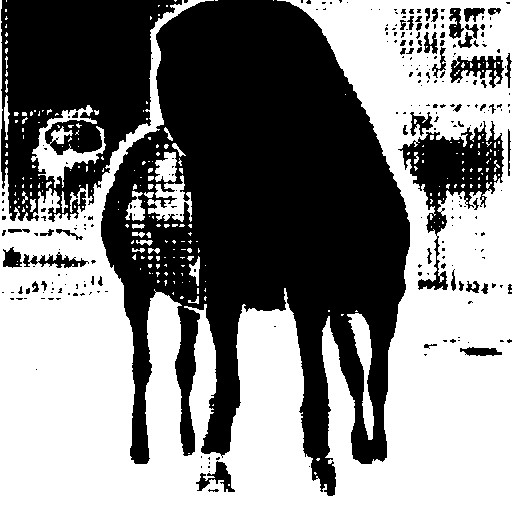} \\
\bottomrule
\end{tabular}
\vspace{3pt}
\caption{\textbf{Additional experiments of DRF.} DRF attains the lowest ArcFace~\cite{deng2019arcface} ($\downarrow$) and strong SAM~\cite{kirillov2023segment} ($\uparrow$), visibly fusing appearance and structure more faithfully than baseline models.}
\label{tab:additional_experiments}
\end{table}

\end{document}